\documentclass{article}

\PassOptionsToPackage{numbers, compress}{natbib}
\usepackage[final]{neurips_2024}

\usepackage{changes}
\definechangesauthor[name={Pablo Morales}, color=orange]{pma}

\usepackage[utf8]{inputenc} % allow utf-8 input
\usepackage[T1]{fontenc}    % use 8-bit T1 fonts
\usepackage{hyperref}       % hyperlinks
\usepackage{url}            % simple URL typesetting
\usepackage{booktabs}       % professional-quality tables
\usepackage{amsfonts}       % blackboard math symbols
\usepackage{nicefrac}       % compact symbols for 1/2, etc.
\usepackage{microtype}      % microtypography
\usepackage{xcolor}         % colors
\usepackage{graphicx}
\usepackage{subcaption}
\usepackage{adjustbox}
\usepackage{tabularray}
\usepackage{multirow}
\usepackage{makecell}
\usepackage{rotating}

\usepackage{xcolor}
\colorlet{review}{black}
\definecolor{improve}{rgb}{0.01, 0.75, 0.24}

\newcommand{\smoothop}[1]{\texttt{Sm}\left( #1 \right)}
\newcommand{\smoothopp}{\texttt{Sm}}

\newcommand*{\abmil}{ABMIL\@}
\newcommand*{\clam}{CLAM\@}
\newcommand*{\dsmil}{DSMIL\@}
\newcommand*{\pathgcn}{PathGCN\@}
\newcommand*{\deepgraphsurv}{DeepGraphSurv\@}

\newcommand*{\transmil}{TransMIL\@}
\newcommand*{\dftdmil}{DFTD-MIL\@}
\newcommand*{\gtp}{GTP\@}
\newcommand*{\setmil}{SETMIL\@}
\newcommand*{\iibmil}{IIBMIL\@}

\newcommand*{\camil}{CAMIL\@}

\newcommand{\transformer}{T\@}
\newcommand{\smoothtransformer}{\smoothopp T\@}
\newcommand{\smoothattpool}{\smoothopp AP\@}
\newcommand{\smoothattpoollate}{\smoothopp AP-late\@}
\newcommand{\smoothattpoolmid}{\smoothopp AP-mid\@}
\newcommand{\smoothattpoolearly}{\smoothopp AP-early\@}
\newcommand{\smoothtransformerattpool}{\smoothopp TAP\@}
\newcommand{\attpool}{AP\@}

\newcommand*{\softmax}{\operatorname{Softmax}}
\newcommand*{\trace}{\operatorname{Trace}}
\newcommand*{\diag}[1]{\operatorname{Diag} \left( #1 \right)}

\newcommand{\direnergy}[1]{\cE_{D}\left( #1 \right)}
\newcommand{\direnergyy}{\cE_{D}}

\newcommand{\energy}[1]{\cE\left( #1 \right)}

\usepackage{bm}
\usepackage{amssymb}
\usepackage{amsmath}

\DeclareMathOperator*{\argmin}{arg\,min}

\def\Rbb{{\mathbb{R}}}
\def\Nbb{{\mathbb{N}}}

\def\cE{{\mathcal E}}

\def\bzero{{\mathbf 0}}

\def\bb{{\mathbf b}}

\def\bff{{\mathbf f}}

\def\bh{{\mathbf h}}

\def\bt{{\mathbf t}}
\def\bu{{\mathbf u}}
\def\bv{{\mathbf v}}
\def\bw{{\mathbf w}}
\def\bx{{\mathbf x}}
\def\by{{\mathbf y}}
\def\bz{{\mathbf z}}

\def\bA{{\mathbf A}}
\def\bB{{\mathbf B}}

\def\bD{{\mathbf D}}

\def\bF{{\mathbf F}}
\def\bG{{\mathbf G}}
\def\bH{{\mathbf H}}
\def\bI{{\mathbf I}}

\def\bL{{\mathbf L}}

\def\bU{{\mathbf U}}

\def\bW{{\mathbf W}}
\def\bX{{\mathbf X}}
\def\bY{{\mathbf Y}}

\title{\smoothopp: enhanced localization in Multiple Instance Learning for medical imaging classification}
\vspace{-1mm}
\author{%
    Francisco M.~Castro-Macías\\
    CITIC-UGR\\
    Dept. of Comp. Science and A. I. \\
    University of Granada\\
    \And
    Pablo Morales-Álvarez\\
    Dept. of Statistics and Operations Research \\
    CITIC-UGR\\
    University of Granada\\
    \And
    Yunan Wu\\
    Dept. of Elect. and Comp. Engineering\\
    Northwestern University
    \And
    Rafael Molina\\
    Dept. of Comp. Science and A. I.\\
    University of Granada\\
    \And
    Aggelos K. Katsaggelos\\
    Dept. of Elect. and Comp. Engineering\\
    Northwestern University
}

\begin{document}

\maketitle

\vspace{-1mm}
\begin{abstract}
Multiple Instance Learning (MIL) is widely used in medical imaging classification to reduce the labeling effort. 
While only bag labels are available for training, one typically seeks predictions at both bag and instance levels (classification and localization tasks, respectively). 
Early MIL methods treated the instances in a bag independently. 
Recent methods account for global and local dependencies among instances. 
Although they have yielded excellent results in classification, their performance in terms of localization is comparatively limited. 
We argue that these models have been designed to target the classification task, while implications at the instance level have not been deeply investigated. 
Motivated by a simple observation -- that neighboring instances are likely to have the same label -- we propose a novel, principled, and flexible mechanism to model local dependencies. It can be used alone or combined with any mechanism to model global dependencies (e.g., transformers).
A thorough empirical validation shows that our module leads to state-of-the-art performance in localization while being competitive or superior in classification. \textcolor{review}{Our code is at \url{https://github.com/Franblueee/SmMIL}.}
\end{abstract}

\vspace{-1mm}
\section{Introduction}
\label{section:introduction}

% Annotating is costly -> weakly supervised (MIL)
% Over the last decades, different medical imaging techniques have been used for the diagnosis and treatment of various diseases \citep{brody2013medical}.
Over the last decades, medical imaging classification has benefited from advances in deep learning \citep{song_analysis_2024,zhou2021review}. 
However, the performance of these methods drops when the number of labeled samples is low, which is common in real-world medical scenarios \citep{aggarwal2021diagnostic}. 
To overcome this, Multiple Instance Learning (MIL) has emerged as a popular weakly supervised approach \citep{fourkioti2023camil,carbonneau2018multiple,dietterich1997solving}. 

% MIL and medical imaging
In MIL, instances are arranged in bags. 
At train time, a label is available for the entire bag, while the instance labels remain unknown. 
The goal is to train a method that, given a test bag, can predict both at bag and instance levels (classification and localization tasks, respectively). 
This paradigm is well suited to the medical imaging domain \citep{quellec2017multiple}.
In cancer detection from Whole Slide Images (WSIs), the WSI represents the bag, and the patches are the instances.
In intracranial hemorrhage detection from Computerized Tomographic (CT) scans, the full scan represents the bag, and the slices at different heights are the instances.
In these scenarios, making accurate predictions of instance labels is extremely important from a clinical viewpoint, as it translates into pinpointing the location of the lesion \citep{campanella2019clinical}. 
% Mention classification and localization as bag-level and instance-level results

% MIL first approach: ABMIL. This has been enhanced in different directions: DFTMIL, DSMIL... other approaches that do not use global dependencies. Then, global dependencies were modeled. Recently, also local dependencies.
Most successful approaches in MIL build on the attention-based pooling \citep{ilse2018attention}, a permutation-invariant operator that assigns an attention value to each instance independently. 
%These values are widely used to detect positive instances (those containing the lesion). 
This method has been extended in different ways
%to compute the attention values
while maintaining the permutation-invariant property \citep{li2021dual,lu2021data,zhang2022dtfd}.
The aforementioned works pose a problem: the dependencies between the instances, which are important when making a diagnosis, are ignored.
To account for this, TransMIL \citep{shao2021transmil} proposed to model global dependencies using a transformer encoder. 
% In short, the goal is to find dependencies between instances through the self-attention mechanism (possibly long-range dependencies).
The idea is to use the self-attention mechanism to introduce interactions between each pair of instances. 
Based on it, other transformer-based approaches have emerged, also focusing on global dependencies \citep{cersovsky2023towards, li2021dt, xiong2023diagnose}.
%More recently, some works have also incorporated local dependencies, which are based on the spatial relationships among the instances \citep{}. 
More recently, several works have also incorporated natural local interactions, which are those between neighboring instances \citep{fourkioti2023camil,zhao2022setmil,zheng2022graph}.
% In parallel, some works incorporated natural local interactions, which are those between neighboring instances \citep{li2018graph,chen2021whole}.
% Recently, both types of interactions have been combined \citep{fourkioti2023camil,zhao2022setmil,zheng2022graph}. 

% Alternative to the following two paragraphs
Although these methods accounting for dependencies have resulted in excellent performance at the bag level, the evaluation at the instance level has received less attention and the results are not comparatively good so far, see the very recent \citep{fourkioti2023camil}. 
In this work, we argue that recent MIL methods have been designed with the classification task in mind, and we propose a new model that focuses on both the classification and localization tasks.
Specifically, we propose a novel and theoretically grounded mechanism to introduce local dependencies, hereafter referred to as \emph{the smooth operator} \smoothopp. 
This is a flexible module that can be used alone on top of classical MIL approaches, or in combination with transformers to also account for global dependencies.
%, or can be used alone on top of classical MIL approaches that do not account for global dependencies. 
In both cases, we show that the proposed operator achieves state-of-the-art localization results while being competitive in classification. 
We compare against a total amount of eight methods, including very recent ones \citep{fourkioti2023camil,zhao2022setmil}. We utilize three different datasets of different nature and size, covering two different medical imaging problems (cancer detection in WSI images and hemorrhage detection in CT scans).

Our main contributions are:
(i) we provide a unified view of current deep MIL approaches;
(ii) we propose a principled mechanism to introduce local interactions, which is a modular component that can be combined or not with global interactions; 
and (iii) we evaluate up to eight state-of-the-art MIL methods on three real-world MIL datasets in both classification and localization tasks, showing that the proposed method stands out in localization while being competitive or superior in classification.

\vspace{-1mm}
\section{Related work}\label{sec:related_work}
% \vspace*{-6mm}

In this work, we tackle the localization task in deep MIL methods using existing concepts and techniques from deep MIL and Graph Neural Networks (GNNs) theory. 
%In the following, we review the related literature in these two fields.

%\subsection{Deep Multiple Instance Learning.}

\textbf{Deep Multiple Instance Learning.}
As explained by \citet{song2023artificial}, deep MIL methods can be divided into two broad categories, namely instance-based or embedding-based, depending on the level at which a specific \textit{aggregation} operator is applied. 
In this paper, we focus on the embedding-based category, and in particular attention-based ones. 
% We briefly review them in the following.

% The first category, which is typically referred to as instance-based, involves aggregating predictions at the instance level. 
% The methods in this category initially generate predictions for the label of each instance in the bag, and then aggregate them to obtain the final bag label prediction \citep{}. 
% The second category is usually designated as embedding-based due to the aggregation occurring at the embedding level. The methods in this group initially obtain an embedding for each instance. Then, these embeddings are aggregated into a bag embedding, which is used to predict the final bag label. 
\citet{ilse2018attention} proposed attention-based pooling to weigh each instance in the bag. 
To improve it, different modifications were proposed, including the use of clustering layers \citep{lu2021data}, grouping the instances in pseudo-bags \citep{zhang2022dtfd}, and using similarity measures to critical instances to compute the attention values \citep{li2021dual}. However, these methods ignore the existing dependencies between the instances in a bag. 
To address this, \citet{shao2021transmil} proposed to use a transformer-based architecture and the PPEG position encoding module. 
This has been extended with different transformer variations, including the deformable transformer architecture \citep{li2021dt}, hierarchical attention \citep{xiong2023diagnose}, and regional sampling \citep{cersovsky2023towards}. 
Recently, these methods have been improved to include spatial information in different ways, including the use of a Graph Convolutional Network (GCN) before the transformer \citep{zheng2022graph}, a neighbor-constrained attention module \citep{fourkioti2023camil}, and a spatial-encoding transformer architecture \citep{zhao2022setmil}. 

In the studies mentioned above, the objective is to obtain increasingly better bag-level results, while the evaluation at the instance level is usually performed qualitatively. 
In contrast, our work addresses both the instance localization task and the bag classification task, as both are of great importance for making a diagnosis. 
% We employ a principled approach that leverages both global and local instance interactions to improve both bag and instance-level results. 
Moreover, our work is not limited to WSI classification; it is also valid for other medical imaging modalities.

% \subsection{Graph Neural Networks.}
\textbf{Graph Neural Networks.}
Our motivation --- that neighboring instances are likely to have the same label --- is a well-established assumption within the machine learning community, often referred to as the cluster assumption \citep{chapelle2002cluster,seeger2000learning}.
Since leveraged in 1984 by \citet{ripley1981spatial} in the context of spatial statistics, it has been extensively used in spectral clustering \citep{ng2001spectral}, semi-supervised learning on graphs \citep{belkin2002semi}, and recently in GNNs \citep{kipf2016semi}. Our work builds upon seminal works in these areas. 

The proposed smooth operator is derived considering a Dirichlet energy minimization problem, similar to the work by \citet{zhou2005regularization} and \citet{zhou2003learning}. 
This approach has been employed in recent years to obtain new GNN models, including the $p$-Laplacian layer \citep{fu2022p}, and PPNP layer \citep{gasteiger2018predict}. 
Moreover, the Dirichlet energy has been studied in the context of GNNs to analyze the over-smoothing phenomenon \citep{cai2020note,li2018deeper}. 
In this regard, our bound on the decrease of the Dirichlet energy is analogous to the result derived by \citet{li2018deeper} to study over-smoothing for GCNs.
Our result, however, holds for the proposed mechanism, of which the graph convolutional layer is a special type. 

\vspace{-1mm}
\section{Background: A unified view of deep MIL approaches}
\label{section:background}

% Outline: MIL notation + unif view of deep MIL
We first describe the binary MIL problem tackled in this paper. 
Then, we provide a unified view of the most popular deep MIL methods. As explained in \autoref{sec:related_work}, we focus on embedding-based approaches. 

% MIL notation
In MIL, the training set consists of pairs of the form $\left( \bX, Y\right)$, where $\bX \in \Rbb^{N \times P}$ is a bag of instances and $Y \in \left\{ 0,1 \right\}$ is the bag label.
We write $\bX = \left[ \bx_1, \ldots, \bx_{N}\right]^T \in \Rbb^{N \times P}$, where $\bx_n \in \Rbb^P$ are the instances. 
Each instance $\bx_n$ is associated to a label $y_n \in \{0,1\}$, \emph{not available during training}. 
It is assumed that $Y = \max \left\{ y_1, \ldots, y_{N}\right\}$, i.e., a bag $\bX$ is considered positive if and only if there is at least one positive instance in the bag. 

\begin{figure}
    \centering
    \begin{subfigure}{0.24\textwidth}
        \centering
        \includegraphics[trim={0cm 0cm 0cm 0cm},clip,height=65pt]{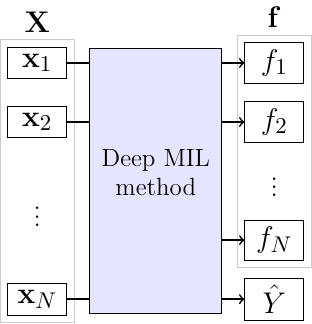}
        \caption{General deep MIL model.}
        \label{fig:att_mil_model-general}
    \end{subfigure}
    \hfill
    \begin{subfigure}{0.24\textwidth}
        \centering
        \includegraphics[trim={0cm 0cm 0cm 0cm},clip,height=65pt]{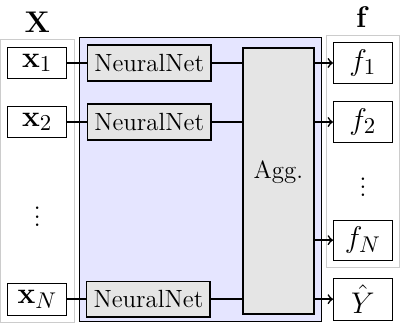}
        % \caption{}
        \caption{Instances are treated independently.}
        \label{fig:att_mil_model-independent}
    \end{subfigure}    
    \hfill
    \begin{subfigure}{0.24\textwidth}
        \centering
        \includegraphics[trim={0cm 0cm 0cm 0cm},clip,height=65pt]{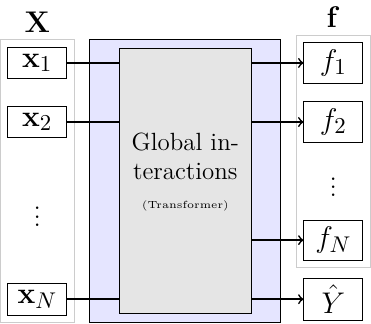}
        % \caption{}
        \caption{Only global interactions.}
        \label{fig:att_mil_model-global}
    \end{subfigure}  
    \hfill
    \begin{subfigure}{0.24\textwidth}
        \centering
        \includegraphics[trim={0cm 0cm 0cm 0cm},clip,height=65pt]{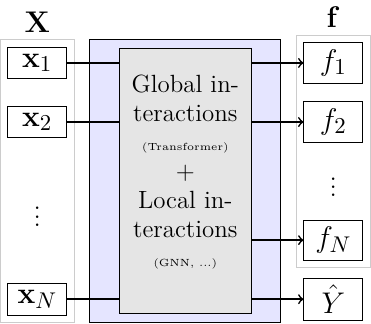}
        % \caption{}
        \caption{Global and local interactions.}
        \label{fig:att_mil_model-global_local}
    \end{subfigure}  
    \caption{(a) Unified view of deep MIL models. Depending on how instances interact with each other in (a), we devise three different families of methods: (b), (c), (d).
    % {\color{red} Update figure with last version}
    }
    \label{fig:att_mil_model}
    \vspace{-4mm}
\end{figure}

% general deep MIL outputs and overview of three families
Given a previously unseen bag (e.g., a medical image), the goal at test time is to: i) predict the bag label (classification task) and ii) obtain predictions or estimates for the instance labels (localization task). 
In general, deep MIL models output a bag-level prediction $\hat{Y}$, as well as instance-level scalars $f_n$ that are used for instance-level prediction. 
This general process is depicted in \autoref{fig:att_mil_model-general}.
In many approaches, these $f_n$ are the so-called \emph{attention values} (e.g., \abmil\ \citep{ilse2018attention}, \transmil\  \citep{shao2021transmil}, \camil\ \citep{fourkioti2023camil}), but they can be obtained in different ways (e.g., through GraphCAM in \gtp\ \citep{zheng2022graph}).
Within the general process in \autoref{fig:att_mil_model-general}, deep MIL models can be categorized into three families, depending on how instances interact with each other, see \autoref{fig:att_mil_model-independent}, \autoref{fig:att_mil_model-global}, and \autoref{fig:att_mil_model-global_local}. 

% Fam1: explain the idea (no interactions). Present ABMIL as an example. 
In the first family, shown in \autoref{fig:att_mil_model-independent}, the instances are encoded \emph{independently} and then aggregated.
The well-known ABMIL \citep{ilse2018attention} fits in this paradigm.
Subsequent works introduce slight modifications to \abmil, while still encoding each instance \emph{independently} \citep{li2021dual,lu2021data,zhang2022dtfd}.
% Let us set the notation for ABMIL, since we will rely on it to introduce our model.
\abmil, on which we will rely to introduce our model, is depicted in \autoref{fig:abmil}.
First, a bag of embeddings $\bH = \left[ \bh_1, \ldots, \bh_N \right] \in \Rbb^{N \times D}$ is obtained by applying a neural network independently to each instance. 
Then, the attention-based pooling computes the attention values $\bff$ and the bag embedding $\bz$ according to
% \begin{equation}
%     \bF = \tanh \left( \bH \bW^\top \right), \quad \bff = \bF \bw,
% \end{equation}
% \begin{align}
%     \bff &= \bF \bw; \quad \bF = \tanh \left( \bH \bW^\top \right), \label{eq:ABMIL_att} \\ 
%     \bz &=  \bH^\top \softmax\left( \bff \right), \label{eq:ABMIL_agg}
% \end{align}
\begin{gather}
	\bF = \tanh \left( \bH \bW^\top \right), \quad \bff = \bF \bw,   \label{eq:attpool-f}\\
    \bz = \operatorname{AttPool}\left( \bH \right) = \bH^\top \softmax\left( \bff \right), \label{eq:attpool-z}
\end{gather}
where $\bW\!\in\!\Rbb^{L \times D}$, $\bw\!\in\! \Rbb^{L}$ are trainable weights. Last, $\hat Y$ is obtained by applying a linear classifier on $\bz$.
% where $\bW \in \Rbb^{L \times D}$ and $\bw \in \Rbb^{L}$ are trainable weights. \autoref{eq:ABMIL_att} and \autoref{eq:ABMIL_agg} define the so-called attention pooling ($\bH\mapsto \bz=\operatorname{AttPool}(\bH)$). Finally, $\hat T$ is obtained with a linear classifier on $\bz$.

% Fam2: explain the idea (global inter.). Mention TransMIL as an example. 
The second family accounts for \emph{global} interactions between instances, possibly long-range ones, see \autoref{fig:att_mil_model-global}.
These works treat instances as tokens that interact through the self-attention mechanism.
This way, global interactions between instances are learned. 
One of the most popular approaches in this family is TransMIL \citep{shao2021transmil}, which was later extended in different directions \citep{cersovsky2023towards,li2021dt}.
% Fam3: explain the idea (global+local inter.). Mention CAMIL,GTP as examples. 
The third family complements the previous one with \emph{local} interactions defined by a fixed neighborhood, see \autoref{fig:att_mil_model-global_local}.
They differ in how local interactions are represented, e.g., as a graph in CAMIL \citep{fourkioti2023camil} and GTP \citep{zheng2022graph}, or using a position-encoded feature map in \setmil\ \citep{zhao2022setmil}.  
% They differ in how locality is defined, e.g. contiguous instances in CAMIL \citep{fourkioti2023camil} and GTP \citep{zheng2022graph}, or a wider neighborhood in SetMIL \citep{zhao2022setmil}.
% Also, they differ in how these local interactions are encoded into the model. 

% Gap in current methods: localization. Local dependencies are not modeled in a principled way (examples). 
In most of these works, the localization task is assessed qualitatively, e.g., by visually comparing the attention maps. 
This contrasts with the classification task, which is always evaluated quantitatively.
As evidenced by \citet{fourkioti2023camil}, this has translated into comparatively poor performance in terms of localization.
We notice that current models have been designed to target the classification task, and they excel at that. 
However, their model design is not as careful about the instance-level implications.
For example, \camil\ \citep{fourkioti2023camil} does not leverage any local information to obtain the instance-level attention values. 
% \textcolor{review}{
%     Indeed, one deduces that the normalized attention values are obtained from the previous tile representations, which have not undergone any local interaction.
% }
Indeed, from their Eq. (8) one deduces that the $a_i$ values are obtained from the tile representations $\bt_i$, which have not undergone any local interaction. 
Observe that local interactions take place in Eq. (4) and Eq. (5) in their paper, but these only affect the bag-level predictions, not the instance-level ones.  
Similarly, \gtp\ \citep{zheng2022graph} introduces local interactions through an off-the-shelf graph convolutional layer, the effect of which is not investigated at the instance level. 
In the following section, we propose a principled approach to account for meaningful local interactions based on the Dirichlet energy. 
The idea is motivated by a natural property often observed in the instance-level labels of medical images: the similarity between neighboring instances. 

\vspace{-1mm}
\section{Method: Introducing smoothness in the attention values}
\label{section:method}

\begin{figure}
    \begin{subfigure}[b]{1.0\textwidth}
        \centering        
        \includegraphics[trim={0.5cm 0.5cm 0.5cm 0.5cm},clip,height=60pt]{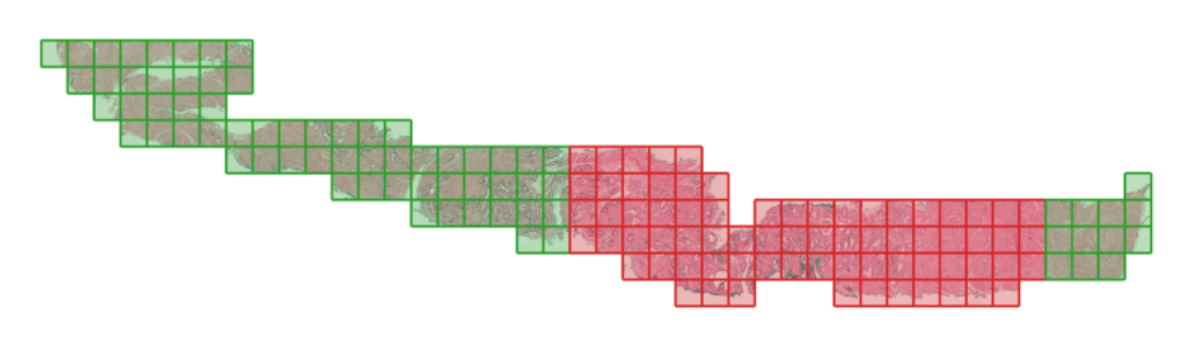}
        \caption{Labelled patches in a WSI.}
    \end{subfigure}
    \begin{subfigure}[b]{1.0\textwidth}
        \centering
        \includegraphics[trim={0cm 0cm 0cm 0cm},clip,width=0.9\textwidth]{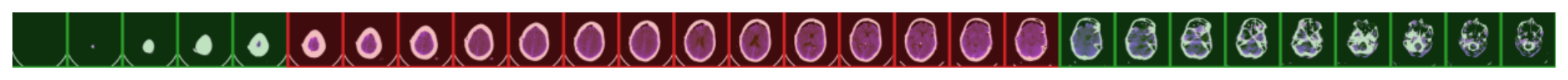}
        \caption{Labelled slices in a CT scan.}
    \end{subfigure}
    \caption{
    WSIs are divided into patches. CT scans are provided as slices. They often show spatial dependencies: in a WSI, a patch is usually surrounded by patches with the same label, while in a CT scan, a slice is usually surrounded by slices with the same label. The red color indicates malignant/hemorrhage patches/slices.
    }
    \label{fig:instance_correlations}
    \vspace{-5mm}
\end{figure}

% Basic idea (smoothing attention) + Outline 
In medical imaging, instance labels are \textit{a priori} expected to exhibit local dependencies with their neighboring instances: an instance is likely to be surrounded by instances with the same label, see \autoref{fig:instance_correlations}.
Recall that attention values are commonly used as a proxy to estimate these labels, so they should inherit this property. 
Based on these observations, our intuitive idea is to favor a \textit{smoothness} property on the attention values. 
To this end, Sec. \ref{subsection:method-modelling_smoothness} formalizes the notion of smoothness through the Dirichlet energy. 
Sec. \ref{subsection:method-smooth_operator} presents the proposed smoothing operator \smoothopp, which encourages smoothness as well as fidelity to the original signal. 
Sec. \ref{subsection:method-proposed_method} proposes how to leverage \smoothopp\ in the context of MIL, both in combination with global interactions (via transformers), and without them. 
We will build on top of the well-known and simple \abmil\ to isolate the effect of \smoothopp\ and avoid over-sophisticated models.

\subsection{Modelling the smoothness}
\label{subsection:method-modelling_smoothness}

We represent each bag as a graph, where the nodes are the instances and the edges represent the spatial connectivity between instances. 
Formally, we suppose that each bag $\bX \in \Rbb^{N \times D}$ has been assigned an adjacency matrix $\bA = \left[ A_{ij} \right] \in \Rbb^{N \times N}$, defined by $A_{ij} > 0$ if instances $\bx_i$ and $\bx_j$ are neighbors, and $A_{ij} = 0$ otherwise. We assume that the adjacency matrix is symmetric, i.e. $A_{ij} = A_{ji}$.

The \textit{Dirichlet energy} is a well-known functional that measures the variability of a function defined on a graph \citep{zhou2005regularization,zhou2003learning}.
In our case, we think of this function as the attention values $\bff \in \Rbb^N$, recall \autoref{fig:att_mil_model-general}. 
% Our goal is to show that the Dirichlet energy of $\bff$ is controlled by that of the (possibly multivariate) representations obtained in previous layers. 
As we shall see below, it will be necessary to define the Dirichlet energy for multivariate graph functions.  
Given a multivariate graph function $\bU = \left[ \bu_1, \ldots, \bu_N \right]^\top \in \Rbb^{N \times D}$ defined on the bag graph, the Dirichlet energy of $\bU$ is given by
\begin{equation}\label{eq:DE_def}\textstyle 
    \direnergy{\bU} = \frac{1}{2}\sum_{i=1}^N \sum_{j=1}^N A_{ij} \left\| \bu_i - \bu_j \right\|_2^2 = \trace\left(\bU^\top \bL \bU\right),
\end{equation}
where $\left\| \cdot \right\|_2$ denotes the Euclidean norm,
$\bL$ is the graph Laplacian matrix $\bL = \bD - \bA$,
$\bD \in \Rbb^{N \times N}$ is the degree matrix, $\bD = \diag{D_1, \ldots, D_N }$, $D_n = \sum_i A_{ni}$.
When $D=1$ we obtain the definition for univariate graph functions, such as the attention values $\bff$.

\textbf{Bounding $\direnergyy$ on the attention values.}
In most deep MIL approaches, the attention values $\bff$ are obtained by applying a neural network to instance-level features. 
One example is \abmil\ \citep{ilse2018attention}, which uses a two-layer perceptron defined by \autoref{eq:attpool-f}. 
Noting that $\tanh$ is a Lipschitz function with Lipschitz constant equal to $1$, we arrive at the following chain of inequalities 
\begin{equation}\label{eq:ineq_chain_2}
    \direnergy{\bff} \leq \left\| \bw \right\|_2^2 \direnergy{\bF} \leq \left\| \bw \right\|_2^2 \left\| \bW \right\|_2^2 \direnergy{\bH},
\end{equation}
where $\left\| \cdot \right \|_2$ denotes the spectral norm. 
A more general result holds in the general case of an arbitrary multi-layer perceptron, see \autoref{appendix:section:proofs} for a proof. 
The above chain of inequalities tells us that if we want $\cE_{D} \left( \bff \right)$ to be low, \emph{we can act on $\bff$ itself or on previous layers (e.g., on $\bF$ or on $\bH$)}, constraining the norm of the trainable weights to remain constant. 
This constraint can be achieved using spectral normalization \citep{miyato2018spectral}, and we study its influence in \autoref{appendix:subsection:ablation}.
In the next subsection, we propose an operator that can be used on any of these levels ($\bff$, $\bF$, $\bH$) to reduce the Dirichlet energy of its output.

\subsection{The smooth operator}
\label{subsection:method-smooth_operator}

Our goal now turns into finding an operator $\smoothopp\ \colon \Rbb^{N \times D} \to \Rbb^{N \times D}$ that, given a bag graph multivariate function $\bU \in \Rbb^{N \times D}$, returns another bag graph multivariate function $\smoothop{\bU} \in \Rbb^{N \times D}$ such that its Dirichlet energy is lower without losing the information present in the original $\bU$. Following seminal works \citep{zhou2005regularization,zhou2003learning}, we cast this as an optimization problem, 
\begin{gather}
    \smoothop{\bU} = \textstyle \argmin_{\bG} \energy{\bG}, \\
    \energy{\bG} = \textstyle \alpha \direnergy{\bG} + \left( 1 - \alpha \right) \left\| \bU - \bG \right\|^2_{F}, \label{eq:energy}
\end{gather}
where $\alpha \in \left[ 0,1 \right)$ accounts for the trade off between both terms, and $\left\| \cdot \right \|_F$ denotes the Frobenius norm. The first term in the above equation penalizes functions with too much variability, while the second term penalizes functions that differ too much from the original $\bU$. 
Note that this can be interpreted as a maximum a posteriori formulation, where the first term corresponds to the prior distribution and the second to the observation model, see \citep{ripley1981spatial}. The objective function $\cE$ is strictly convex, and therefore admits a unique solution, given by
\begin{equation}\label{eq:smoothop_exact}
    \smoothop{\bU} = \left( \bI + \gamma \bL \right)^{-1} \bU,
\end{equation}
where $\gamma = \alpha / (1-\alpha)$. 
% As noted by \citet{gasteiger2018predict}, this result is connected to the PageRank algorithm \citep{ma2008bringing}.
Unfortunately, the expression in Eq. \eqref{eq:smoothop_exact}, although elegant, incurs prohibitive computational and memory costs, especially when the number of instances in the bag is large (which is the case of WSIs). 
Instead, we can take an iterative approach, defining $\smoothop{\bU} = \bG( T )$, with
\begin{gather}
    \bG( 0 ) = \bU; \quad \bG( t ) = \alpha \left( \bI - \bL \right) \bG( t-1 ) + \left( 1 - \alpha \right) \bU, \quad t \in \left\{ 1, \ldots, T\right\}.\label{eq:smoothop_approx}
\end{gather}
As demonstrated by \citet{zhou2003learning}, the sequence $\left\{ \bG( t ) \right\}$ converges to the optimal solution in \autoref{eq:smoothop_exact}. As studied by \citet{gasteiger2018predict}, it is enough to use a small number of iterations $T$ to approximate the exact solution. Therefore, in this work, we will adopt the iterative approach described by \autoref{eq:smoothop_approx}.
Based on previous work \citep{gasteiger2018predict}, we will use $T=10$, and $\alpha$ will be set as a trainable parameter initialized at $\alpha=0.5$. 
See \autoref{appendix:subsection:ablation} for a study on the effects of these hyperparameters \textcolor{review}{and \autoref{fig:attmaps-camelyon-alpha-appendix} for a visual comparison of the effect that $\alpha$ has on the attention maps.}

%\textcolor{red}{In the experimental section, we will show that $\alpha$ keeps almost constant through training}.

\textbf{Theoretical guarantees via the normalized Laplacian.} We present a result that informs us about the rate at which the Dirichlet energy decreases when applying $\smoothopp{}$. Let us define $\lambda_{\gamma}^* = \max \left\{ \left( 1 + \gamma \lambda_n \right)^{-2} \colon \lambda_n \in \Lambda \setminus \left\{ 0 \right\}\right\}$, where $ \Lambda = \left\{ \lambda_1, \ldots, \lambda_N \right\}$ are the eigenvalues of the bag graph Laplacian matrix. Then, we have the following inequality,
\begin{equation}\label{eq:smooth_ineq}
    \direnergy{\smoothop{\bU}} \leq \lambda_{\gamma}^* \direnergy{\bU}.
\end{equation}
The proof is inspired by \citet{cai2020note}, see \autoref{appendix:section:proofs}. If $\lambda_{\gamma}^* < 1$, then the smooth operator effectively decreases the Dirichlet energy. If we replace the Laplacian matrix by the normalized Laplacian matrix%\footnote{Using the same notation as before, the normalized Laplacian matrix is defined as $\tilde{\bL} = \bD^{-1/2} \bL \bD^{-1/2} $. }, 
, $\tilde{\bL} = \bD^{-1/2} \bL \bD^{-1/2} $,
it is known that its eigenvalues lie in the interval $\left[ 0, 2 \right)$, and then $\lambda_{\gamma}^* < 1$ holds. This motivates the use of the normalized Laplacian in our experiments. 

The smooth operator \smoothopp\ only introduces one parameter to be estimated, $\alpha$. Also, it is differentiable with respect to its input. 
Therefore, it can be integrated into simple attention-based MIL models, such as ABMIL, to account for local dependencies.  
%Next section elaborates on how to leverage \smoothopp\ in MIL. 

\subsection{The proposed model}
\label{subsection:method-proposed_method}

\begin{figure}
    \centering
    \centering
    \begin{subfigure}[b]{0.3\textwidth}
        \centering
        \includegraphics[trim={0cm 0cm 0cm 0cm},clip,width=1.0\textwidth]{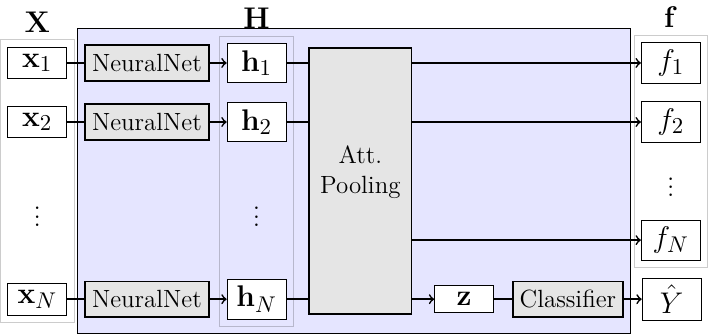}
        \caption{\abmil.}
        \label{fig:abmil}
    \end{subfigure}
    \hfill
    \begin{subfigure}[b]{0.3\textwidth}
        \centering
        \includegraphics[trim={0cm 0cm 0cm 0cm},clip,width=1.0\textwidth]{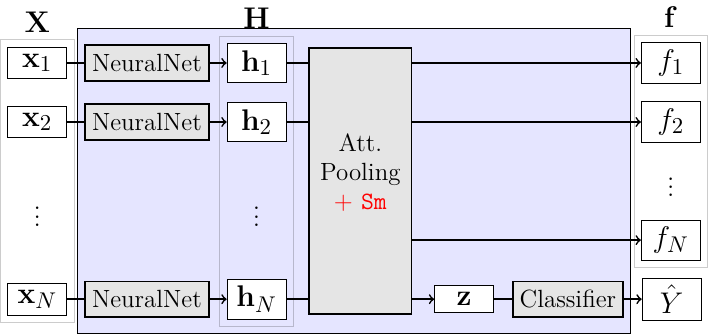}
        \caption{\smoothattpool.}
        \label{fig:smoothattpool}
    \end{subfigure}
    \hfill
    \begin{subfigure}[b]{0.3\textwidth}
        \centering
        \includegraphics[trim={0cm 0cm 0cm 0cm},clip,width=1.0\textwidth]{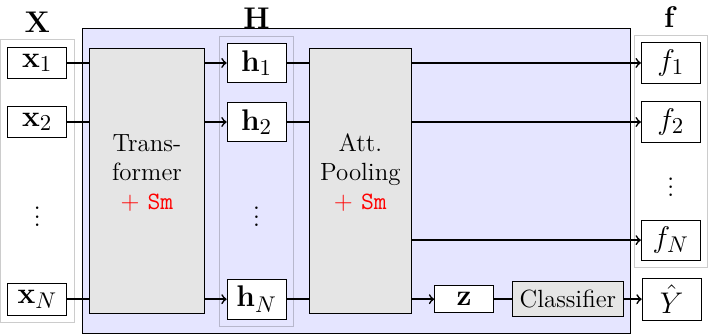}
        \caption{\smoothtransformerattpool.}
        \label{fig:smoothattpool_smoothtransformer}
    \end{subfigure} 
    \caption{Smooth Attention Multiple Instance Learning. (a) The well-known model in \citep{ilse2018attention}, which we build upon. (b): only local interactions are considered by applying the proposed smooth operator \smoothopp\ in the aggregation part. (c): both global and local interactions are considered by applying \smoothopp\ both in the transformer and in the aggregation parts.}
    \label{fig:smooth_att_mil_model}
    \vspace{-4mm}
\end{figure}

Here we propose how to leverage the operator \smoothopp\ in the context of MIL.
We build on top of the well-known \abmil.
First, we introduce \smoothattpool, which integrates \abmil\ with \smoothopp\ and only accounts for local interactions.  
Second, we introduce \smoothtransformerattpool, which equips \smoothattpool\ with a transformer encoder to account for global dependencies. 
The proposed models are depicted in \autoref{fig:smoothattpool} and \autoref{fig:smoothattpool_smoothtransformer}. 
The details about the architecture we have used can be found in \autoref{appendix:subsection:configuration}. 

\textbf{\smoothattpool: Smooth Attention Pooling.}
This is represented in \autoref{fig:smoothattpool}. 
First, the bag of embeddings $\bH$ is obtained as in ABMIL \citep{ilse2018attention}, i.e. treating the instances independently. 
Then, the operator \smoothopp\ is integrated within the attention pooling.
Based on \autoref{eq:ineq_chain_2}, this can be done on the attention values themselves or on previous representations. 
We consider three different variants: \smoothattpoollate, \smoothattpoolmid, \smoothattpoolearly. They act, respectively, on $\bff$ (the attention values themselves), on $\bF$ (i.e. before entering the last layer), and on $\bH$ (i.e. before entering the attention-based pooling). Formally,
\begin{align}
    \text{late:}\quad \bff & = \smoothop{ \tanh \left( \bH \bW^\top \right) \bw}, \label{eq:smoothattpool_late}\\
    \text{mid:}\quad \bff & = \tanh \left( \smoothop{\bH \bW^\top} \right) \bw, \label{eq:smoothattpool_mid}\\
    \text{early:}\quad \bz & = \operatorname{AttPool}\left( \smoothop{\bH} \right) \label{eq:smoothattpool_early},
\end{align}
%  
% see \autoref{fig:attpooling_variants} for a graphical representation.
While \smoothattpoollate\ and \smoothattpoolmid\ act on the computation of the attention values, \smoothattpoolearly\ acts on the embedding that is passed to the attention-based pooling, see \autoref{fig:attpooling_variants} in \autoref{appendix:section:tables_figures}.
We use  \smoothattpoolearly{} by default. \autoref{subsection:experiments-ablation} shows that results do not differ much among configurations.

\textbf{\smoothtransformerattpool: Smooth Transformer Attention Pooling.} 
This is represented in \autoref{fig:smoothattpool_smoothtransformer}.
The only difference with \smoothattpool\ is that the neural network acting independently on the instance embeddings is replaced by a transformer encoder to account for global dependencies.
%We note that the bag of embeddings $\bH$ in \autoref{eq:ineq_chain_2} can be computed by a transformer encoder, so we propose to also apply \smoothopp\ to its output, 
Based on the idea that smoothness can be imposed at previous locations, recall \autoref{eq:ineq_chain_2}, we propose to also apply \smoothopp\ to the transformer output:
\begin{equation}\label{eq:SmTAP}
    \bH = \smoothop{\softmax\left( q\left( \bX \right) k\left( \bX \right)^\top \right) v\left( \bX \right)},
\end{equation}
where $q$, $k$, and $v$ are the standard queries, keys, and values in the dot product self-attention \citep{bishop2024deep}. 
Notice that \smoothtransformerattpool\ uses \smoothopp\ in two places: the first after the transformer encoder and the second in the aggregator. 
Naturally, one could think of other variants that use \smoothopp\ in only one place or the other. 
In \autoref{subsection:experiments-ablation} we ablate these different configurations, leading to similar results.

%{\color{red} Somewhere, we may want to give more details about our particular models (how many layers of NNs/transformers before the H, size of them etc.). Probably, this should be accompanied by a more detailed figure. All this could be included in the appendix.}

\vspace{-1mm}
\section{Experiments}
\label{section:experiments}

We validate the proposed $\smoothopp{}$ in three medical MIL datasets: RSNA \citep{flanders2020construction}, PANDA \citep{bulten2022artificial}, and CAMELYON16 \citep{bejnordi2017diagnostic}. 
We evaluate the effectiveness of our approach by a quantitative and qualitative analysis. 
All experiments have been conducted under fair and reproducible conditions. 
Details on the datasets and experimental setup can be found in \autoref{appendix:section:exp_details}. 
The code is available at \url{https://github.com/Franblueee/SmMIL}. 

We compare our approaches with state-of-the-art deep MIL methods.
We consider two groups of methods, depending on the presence/absence of a transformer block to model global dependencies. 
In the first group, we include those models that do not use this block: the proposed \smoothattpool, \abmil ~\citep{ilse2018attention}, \clam ~\citep{lu2021data}, \dsmil ~\citep{li2021dual}, and \dftdmil ~\citep{zhang2022dtfd}.
The second group consists of models that do use the transformer encoder: the proposed \smoothtransformerattpool, \transmil ~\citep{shao2021transmil}, \setmil ~\citep{zhao2022setmil}, \gtp ~\citep{zheng2022graph}, and \camil ~\citep{fourkioti2023camil}. 
These groups ensure a fair comparison in terms of model capabilities and complexity. 
\textcolor{review}{
    In \autoref{appendix:subsection:ablation} we report the results of three more methods: \deepgraphsurv ~\citep{li2018graph}, \pathgcn ~\citep{chen2021whole}, and \iibmil ~\citep{ren2023iib}. Note that the performance obtained by these methods does not affect the conclusions we will obtain in this section. 
}
% Note that we have adapted \deepgraphsurv\ and \pathgcn\ for MIL since they were proposed for a regression task (survival prediction).
% \textcolor{review}{
%     In the first group, we include those models that do not use this block: the proposed \smoothattpool, \abmil ~\citep{ilse2018attention}, \deepgraphsurv ~\citep{li2018graph}, \clam ~\citep{lu2021data}, \dsmil ~\citep{li2021dual}, \pathgcn ~\citep{chen2021whole}, and \dftdmil ~\citep{zhang2022dtfd}.
%     Note that we have adapted \deepgraphsurv\ and \pathgcn\ for MIL since they were proposed for a regression task (survival prediction).
%     The second group consists of models that do use the transformer encoder: the proposed \smoothtransformerattpool, \transmil ~\citep{shao2021transmil}, \setmil ~\citep{zhao2022setmil}, \gtp ~\citep{zheng2022graph}, \iibmil ~\citep{ren2023iib}, and \camil ~\citep{fourkioti2023camil}. 
% }

In \autoref{subsection:experiments-localization} we consider the localization task. 
In \autoref{subsection:experiments-classification} we turn to the classification task. 
\autoref{subsection:experiments-ablation} shows an ablation study on how different uses of the smooth operator affect the proposed model.

% {
% \color{blue}
% Key messages (both tasks):
% \begin{itemize}
%     \item Including smoothness improves result at both instance and bag level.
%     \item Our models are the only ones that consistently show good performance at bag and instance level at the same time. 
% \end{itemize}
% }

% T means Transformer. ST means Smooth Transformer. SAP means Smooth Attention Pooling.

\subsection{Localization: instance level results}
\label{subsection:experiments-localization}
% \vspace{-1mm}

% {
% \color{blue}
% Key messages (localization):
% \begin{itemize}
%     \item Smoothness improves localization capacity. Our models show better numbers than the rest. 
%     \item In RSNA and PANDA we see similar behavior. DFTD-MIL and Transformer models show similar performance, while DSMIL stays behind. 
%     \item In CAMELYON16 this changes: Transformer based models fail at identifying positive instances (very low numbers). DSMIL show acceptable results.  We attribute this to the use of features obtained with SSL.
%     \item \textcolor{red}{When no Transformer is used, localization capacity increases dramatically in CAMELYON. We attribute this to the use of features obtained with SSL.}
% \end{itemize}
% }

% - Description of the task: how instance predictions are obtained. 
In this subsection, we analyze the ability of each model to predict the label of the instances inside a bag. 
As explained in \autoref{section:background}, deep MIL models assign a scalar value $f_n$ to each instance $\bx_n$, see \autoref{fig:att_mil_model-general}. 
Although these can be obtained in different ways, for simplicity we will refer to them as \emph{attention values}. 
Thus, we compare the attention values with the ground truth instance labels, which are available for the test set only for evaluation purposes.

\textbf{Quantitative analysis.}
We analyze the performance of each method using the area under the ROC curve (AUROC) and the F1 score. 
Note that a critical hyperparameter for the latter is the threshold used on $f_n$ to determine the label of each instance. 
To ensure a fair comparison, we compute the optimal threshold for each method using the validation set.
% and report the corresponding F1 score. 
As a general summary, we also report the average rank achieved by each model across metrics and datasets. 
%It is computed by ranking each model for each dataset and each score, and then computing the average. 

The results are shown in \autoref{tab:localization-results}. 
We find that using \smoothopp\ provides the best performance overall, placing as the best or second-best within each group.
Only in RSNA the proposed \smoothattpool\ is outperformed by \abmil.
We attribute this to the fact that the bag graphs in CT scans are not as complex as in WSIs, and therefore the local interactions are not as meaningful. 
Note that the performance gain is particularly significant on CAMELYON16, where the bags have a larger number of instances, the graphs are much denser and the imbalance between positive and negative instances is more severe. 
% As it was already observed in \citep{fourkioti2023camil}, in CAMELYON16 Transformer-based models fall short of simpler models, all of which are outperformed by the proposed SAP. 
Notably, \smoothtransformerattpool\ significantly outperforms \setmil, \gtp, and \camil, which also model local dependencies. 
Contrary to our method, their design
is focused on bag-level performance and it does not translate into meaningful instance-level properties.
% While these methods also combine neighbor information, their design
% is focused on bag-level performance and it does not translate into meaningful instance-level properties.
%does not result in the attention maps having any meaningful property. 
% In contrast, our method enforces smooth attention maps in a principled manner.

\begin{table}%[h]
\caption{
Localization results (mean and standard deviation from five independent runs).
The best is in bold and the second-best is underlined.
$(\downarrow)$/$(\uparrow)$ means lower/higher is better.
The proposed operator improves the localization results in all three datasets and both with and without global interactions. It ranks first in eight out of twelve dataset-score pairs. 
}
\label{tab:localization-results}
\centering
\begin{adjustbox}{width=\textwidth}
\begin{tabular}{@{}ccccccccc@{}}
\toprule
& & \multicolumn{2}{c}{\textbf{RSNA}} & \multicolumn{2}{c}{\textbf{PANDA}} & \multicolumn{2}{c}{\textbf{CAMELYON16}} \\ \midrule
& & AUROC $(\uparrow)$ & F1 $(\uparrow)$ & AUROC $(\uparrow)$ & F1 $(\uparrow)$ & AUROC $(\uparrow)$ & F1 $(\uparrow)$ & Rank $(\downarrow)$ \\ \midrule
\multirow{5}{*}{\makecell{Without\\global\\interactions}} & \smoothattpool & $\underline{0.798}_{0.033}$ & $\underline{0.477}_{0.014}$ & $\mathbf{0.799}_{0.005}$ & $\underline{0.635}_{0.006}$ & $\mathbf{0.960}_{0.007}$ & $\mathbf{0.840}_{0.053}$ & $\mathbf{1.500}_{0.548}$ \\
& \abmil & $\mathbf{0.806}_{0.012}$ & $\mathbf{0.486}_{0.033}$ & $0.768_{0.002}$ & $0.602_{0.004}$ & $0.819_{0.074}$ & $0.766_{0.060}$ & $\underline{2.500}_{1.225}$ \\
& \clam & $0.523_{0.069}$ & $0.076_{0.154}$ & $0.727_{0.046}$ & $0.568_{0.038}$ & $0.849_{0.044}$ & $\underline{0.821}_{0.046}$ & $4.167_{1.329}$ \\
& \dsmil & $0.554_{0.004}$ & $0.180_{0.000}$ & $0.765_{0.008}$ & $0.598_{0.006}$ & $0.760_{0.070}$ & $0.654_{0.183}$ & $4.333_{0.516}$ \\
& \dftdmil & $0.747_{0.070}$ & $0.453_{0.194}$ & $\underline{0.795}_{0.004}$ & $\mathbf{0.637}_{0.006}$ & $\underline{0.884}_{0.002}$ & $0.742_{0.040}$ & $2.500_{1.049}$ \\
% \multirow{7}{*}{\makecell{Without\\global\\interactions}} & \smoothattpool & $\underline{0.798}_{0.033}$ & $\underline{0.477}_{0.014}$ & $\mathbf{0.799}_{0.005}$ & $\underline{0.635}_{0.006}$ & $\mathbf{0.961}_{0.007}$ & $\mathbf{0.839}_{0.053}$ & $\mathbf{1.500}_{0.548}$ \\
% & \abmil & $\mathbf{0.806}_{0.012}$ & $\mathbf{0.486}_{0.033}$ & $0.768_{0.002}$ & $0.602_{0.004}$ & $0.816_{0.055}$ & $0.767_{0.039}$ & $\underline{2.833}_{1.602}$ \\
% & \deepgraphsurv & $0.681_{0.054}$ & $0.293_{0.168}$ & $0.720_{0.011}$ & $0.581_{0.026}$ & $\underline{0.959}_{0.033}$ & $0.771_{0.070}$ & $4.333_{1.506}$ \\
% & \clam & $0.523_{0.069}$ & $0.076_{0.154}$ & $0.727_{0.046}$ & $0.568_{0.038}$ & $0.849_{0.044}$ & $\underline{0.821}_{0.046}$ & $5.167_{1.941}$ \\
% & \dsmil & $0.554_{0.004}$ & $0.180_{0.000}$ & $0.765_{0.008}$ & $0.598_{0.006}$ & $0.760_{0.078}$ & $0.654_{0.203}$ & $5.333_{1.033}$ \\
% & \pathgcn & $0.711_{0.049}$ & $0.447_{0.014}$ & $0.664_{0.019}$ & $0.526_{0.019}$ & $0.443_{0.138}$ & $0.077_{0.114}$ & $6.000_{1.549}$ \\
% & \dftdmil & $0.747_{0.070}$ & $0.453_{0.194}$ & $\underline{0.795}_{0.004}$ & $\mathbf{0.637}_{0.006}$ & $0.884_{0.002}$ & $0.742_{0.040}$ & $2.833_{1.329}$ \\
\midrule
\multirow{5}{*}{{\makecell{With\\global\\interactions}}} & \smoothtransformerattpool& $\mathbf{0.767}_{0.046}$ & $\mathbf{0.474}_{0.023}$ & $\mathbf{0.790}_{0.007}$ & $0.622_{0.01}$ & $\mathbf{0.789}_{0.008}$ & $\mathbf{0.600}_{0.067}$ & $\mathbf{1.500}_{1.225}$ \\
& \transmil & $0.732_{0.013}$ & $\underline{0.471}_{0.014}$ & $0.751_{0.011}$ & $\mathbf{0.636}_{0.008}$ & $\underline{0.781}_{0.024}$ & $0.127_{0.078}$ & $3.083_{1.429}$ \\
& \setmil & $0.726_{0.025}$ & $0.438_{0.027}$ & $0.774_{0.007}$ & $0.631_{0.010}$ & $0.615_{0.231}$ & $0.134_{0.267}$ & $3.667_{0.816}$ \\
& \gtp & $0.736_{0.017}$ & $0.425_{0.018}$ & $0.768_{0.022}$ & $\underline{0.636}_{0.011}$ & $0.442_{0.091}$ & $0.037_{0.036}$ & $3.917_{1.429}$ \\
& \camil & $\underline{0.760}_{0.036}$ & $0.456_{0.013}$ & $\underline{0.785}_{0.011}$ & $0.621_{0.013}$ & $0.742_{0.028}$ & $\underline{0.479}_{0.175}$ & $\underline{2.833}_{1.169}$ \\
% \multirow{6}{*}{{\makecell{With\\global\\interactions}}} & \smoothtransformerattpool & $\mathbf{0.767}_{0.046}$ & $\mathbf{0.474}_{0.023}$ & $\mathbf{0.790}_{0.007}$ & $0.622_{0.010}$ & $\underline{0.789}_{0.008}$ & $\mathbf{0.600}_{0.067}$ & $\mathbf{1.833}_{1.602}$ \\
% & \transmil & $0.732_{0.013}$ & $\underline{0.471}_{0.014}$ & $0.751_{0.011}$ & $\underline{0.636}_{0.008}$ & $0.781_{0.024}$ & $0.127_{0.078}$ & $3.500_{1.378}$ \\
% & \setmil & $0.726_{0.025}$ & $0.438_{0.027}$ & $0.774_{0.007}$ & $0.631_{0.010}$ & $0.615_{0.231}$ & $0.134_{0.267}$ & $4.167_{0.753}$ \\
% & \gtp & $0.736_{0.017}$ & $0.425_{0.018}$ & $0.768_{0.022}$ & $0.636_{0.011}$ & $0.442_{0.091}$ & $0.037_{0.036}$ & $4.500_{1.378}$ \\
% & \iibmil & $0.675_{0.017}$ & $0.420_{0.016}$ & $0.740_{0.020}$ & $\mathbf{0.645}_{0.007}$ & $\mathbf{0.873}_{0.138}$ & $0.352_{0.100}$ & $3.833_{2.483}$ \\
% & \camil & $\underline{0.760}_{0.036}$ & $0.456_{0.013}$ & $\underline{0.785}_{0.011}$ & $0.621_{0.013}$ & $0.742_{0.028}$ & $\underline{0.479}_{0.175}$ & $\underline{3.167}_{1.602}$ \\
\bottomrule
\end{tabular}
\end{adjustbox}
\vspace{-4mm}
\end{table}

% Further insights: attention histograms
\textbf{Attention histograms.}
We examine the attention histograms produced by each model on the CAMELYON16 dataset. 
The corresponding figures for RSNA and PANDA can be found in \autoref{appendix:section:tables_figures}.
In \autoref{fig:attval_histograms-camelyon}, we represent the frequency with which attention values are assigned to positive and negative instances, separately. 
An ideal instance classifier would place all the positive instances on the right and all the negative instances on the left. 
This illustrates why \smoothtransformerattpool\ and \smoothattpool\ achieve such a good performance: they concentrate the negative instances to the left of the histogram while succeeding in grouping a large part of the positive instances to the right. 
\transmil\ and \gtp\ assign low attention values to both positive and negative instances. 
\camil\ is able to identify positive instances, but negative instances are assigned from intermediate to high attention values. 
\clam\ and \dsmil\ assign low attention values to negative instances, but the distribution of the positive instances resembles a uniform and a normal distribution, respectively. 

\begin{figure}%[h]
    \centering
    \begin{adjustbox}{width=\textwidth}
    \begin{tabular}{ccccc}
        \includegraphics[trim={0cm 0cm 0cm 0cm},clip,width=0.18\textwidth]
        {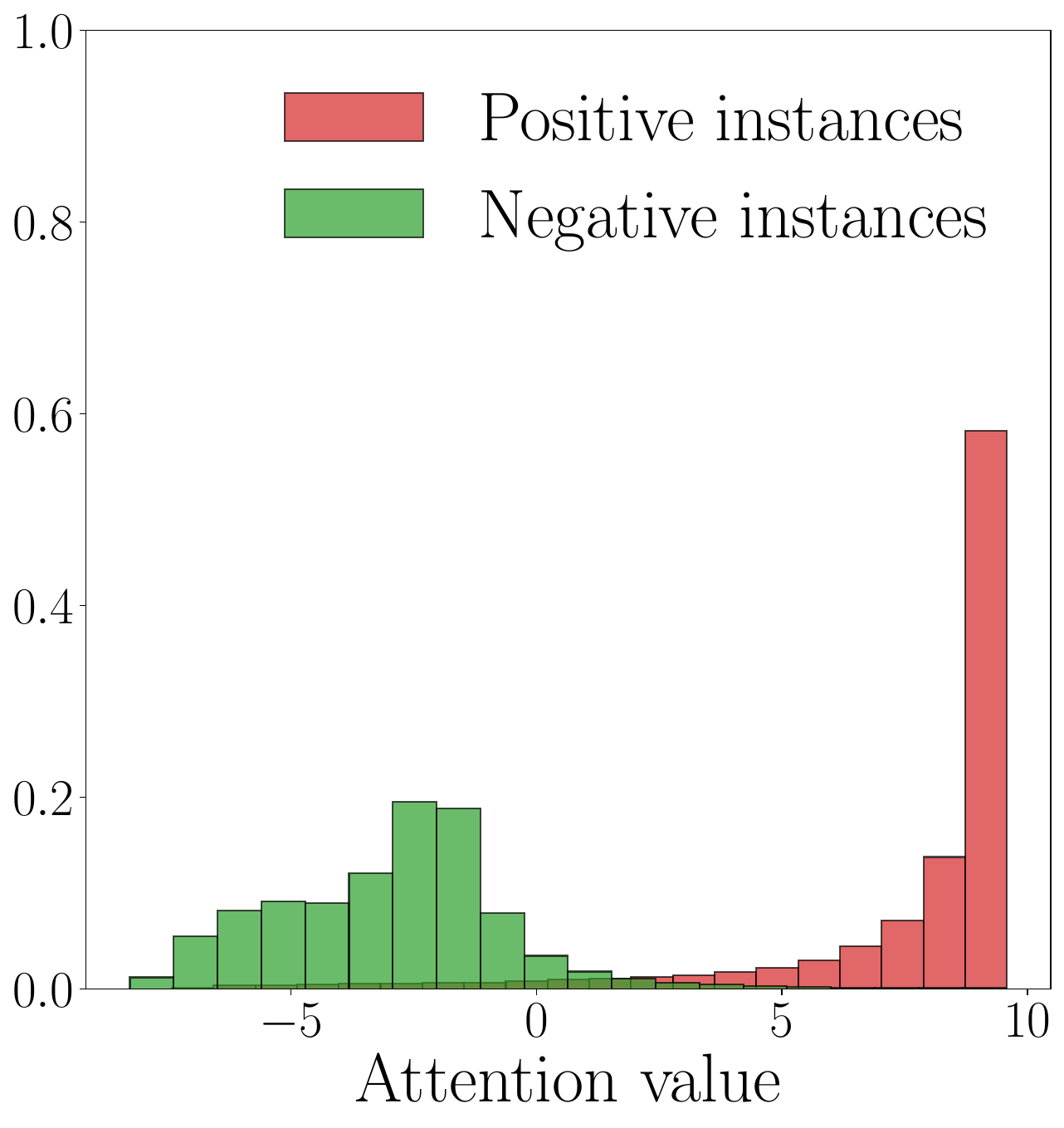}
        & 
        \includegraphics[trim={0cm 0cm 0cm 0cm},clip,width=0.18\textwidth]
        {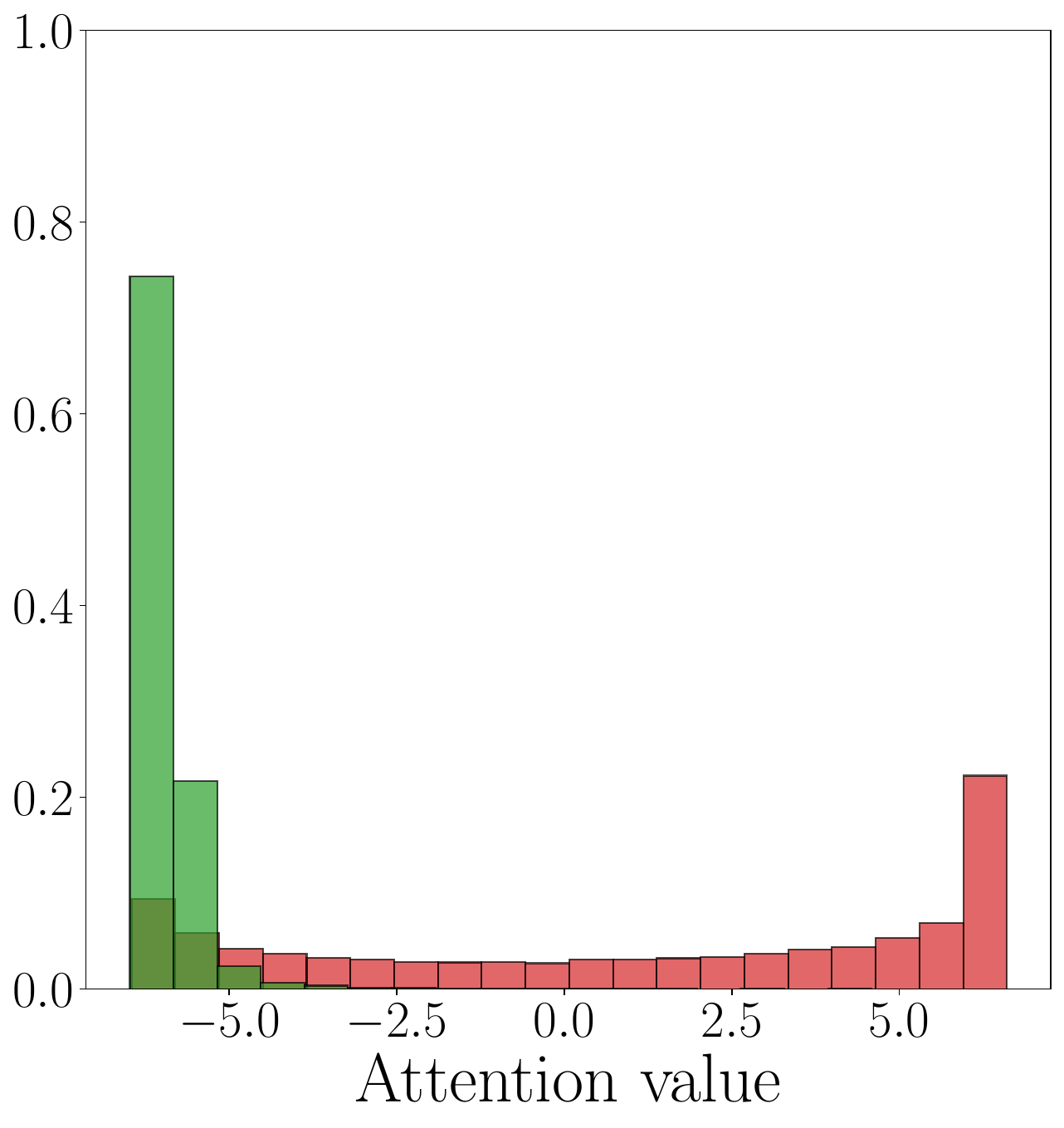}
        & 
        \includegraphics[trim={0cm 0cm 0cm 0cm},clip,width=0.18\textwidth]
        {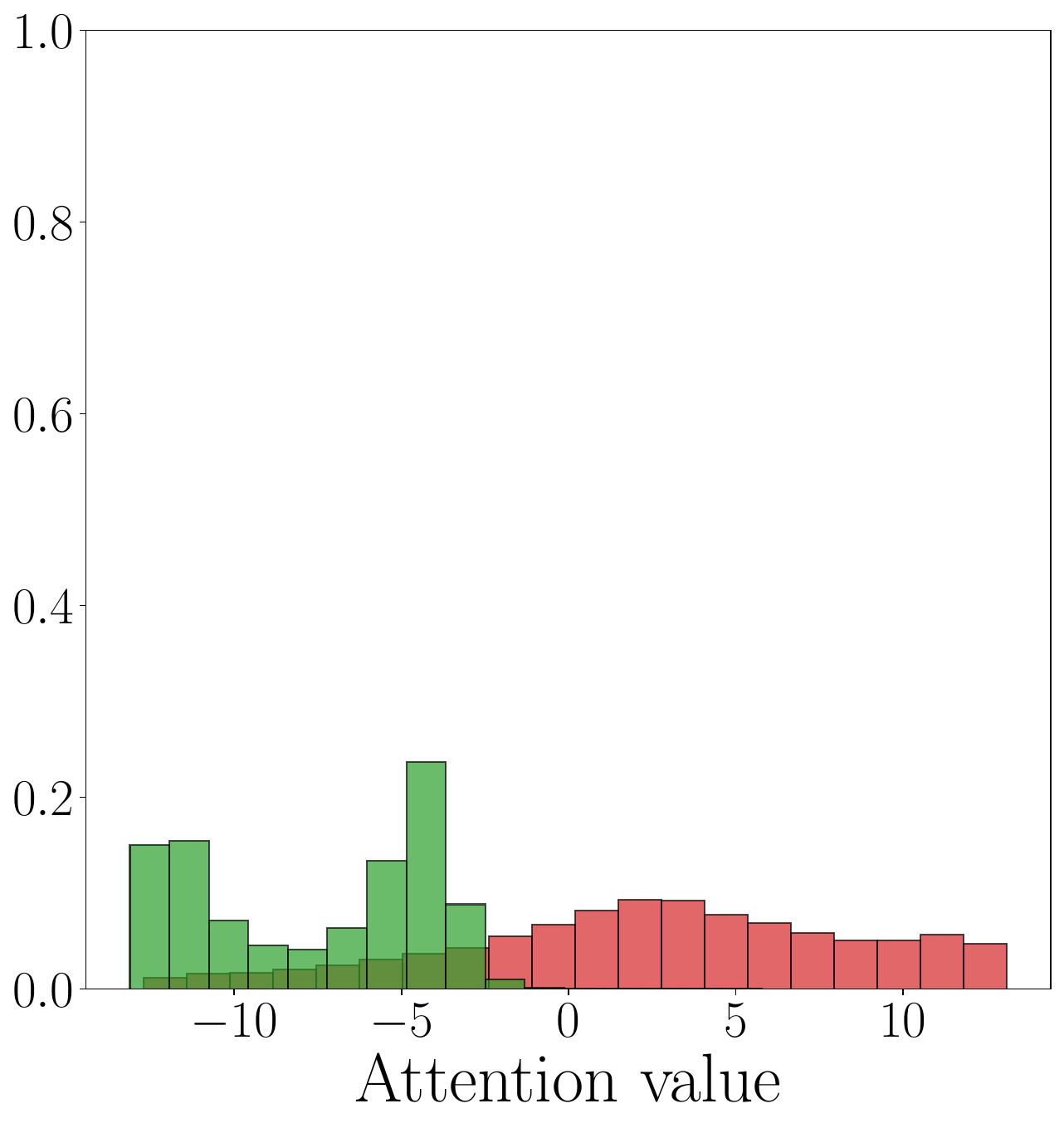}
        & 
        \includegraphics[trim={0cm 0cm 0cm 0cm},clip,width=0.18\textwidth]
        {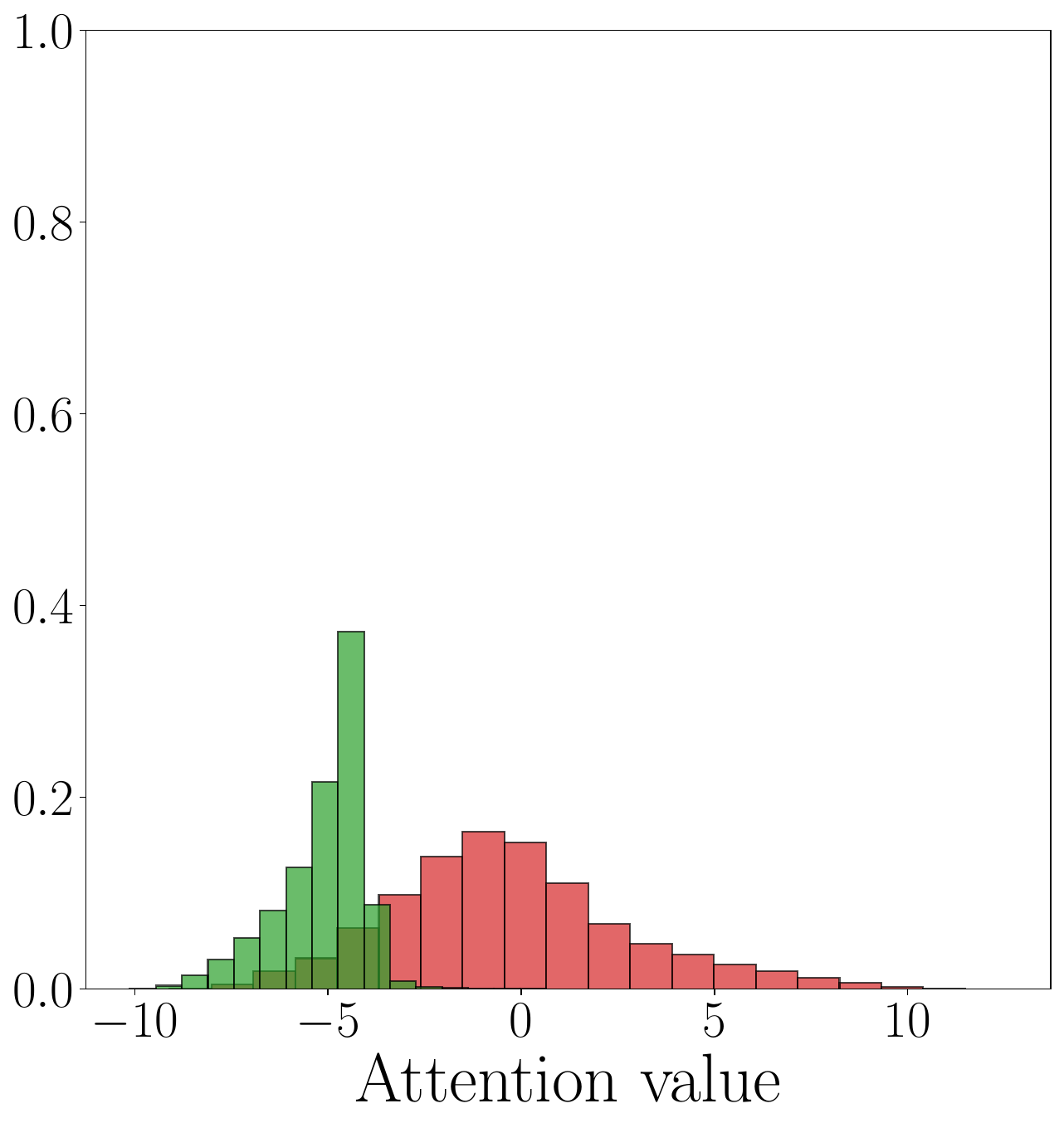}
        &
        \includegraphics[trim={0cm 0cm 0cm 0cm},clip,width=0.18\textwidth]{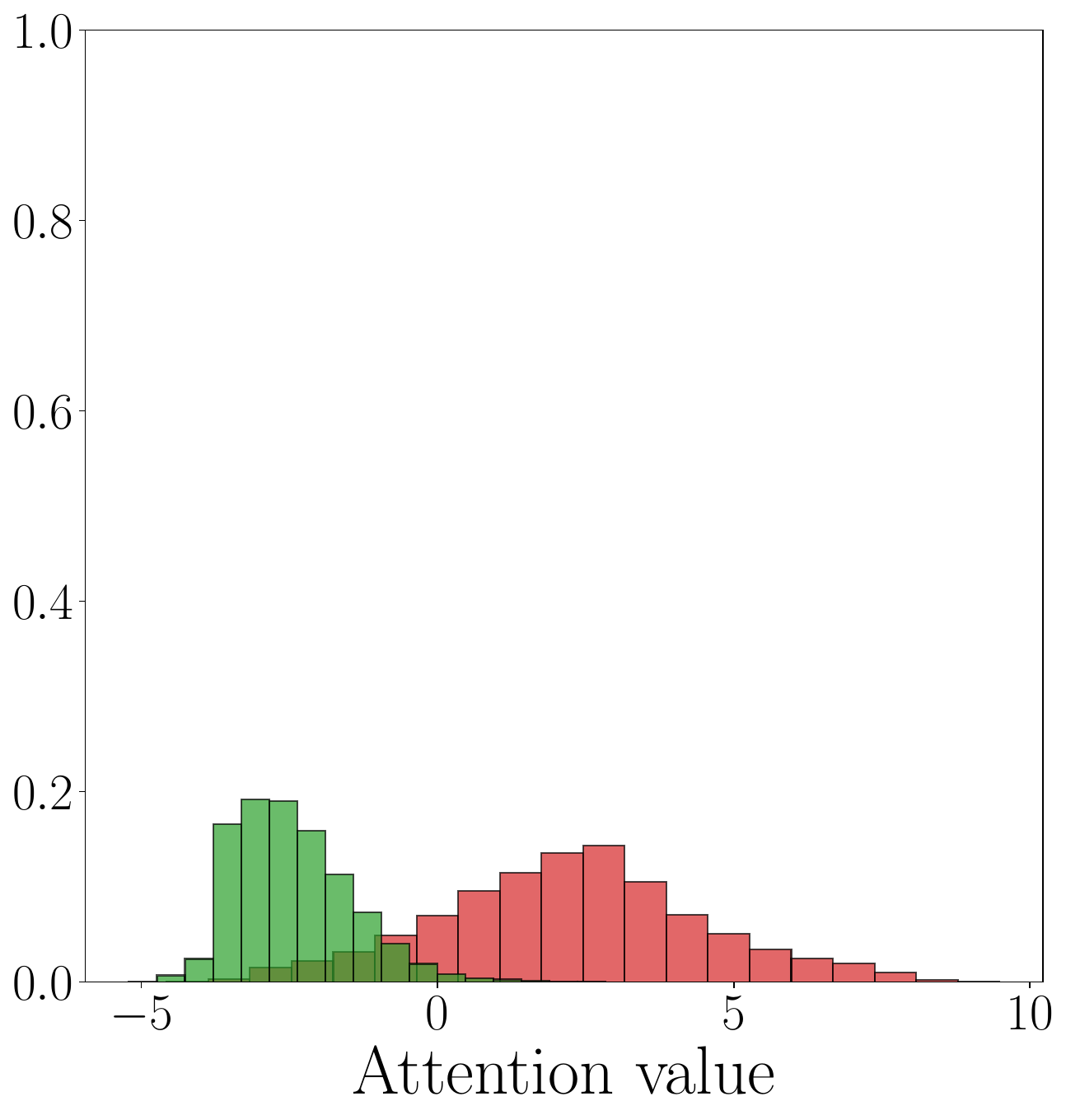}
        \\
        \smoothattpool & \abmil & \clam & \dsmil & \dftdmil \\
        \includegraphics[trim={0cm 0cm 0cm 0cm},clip,width=0.18\textwidth]
        {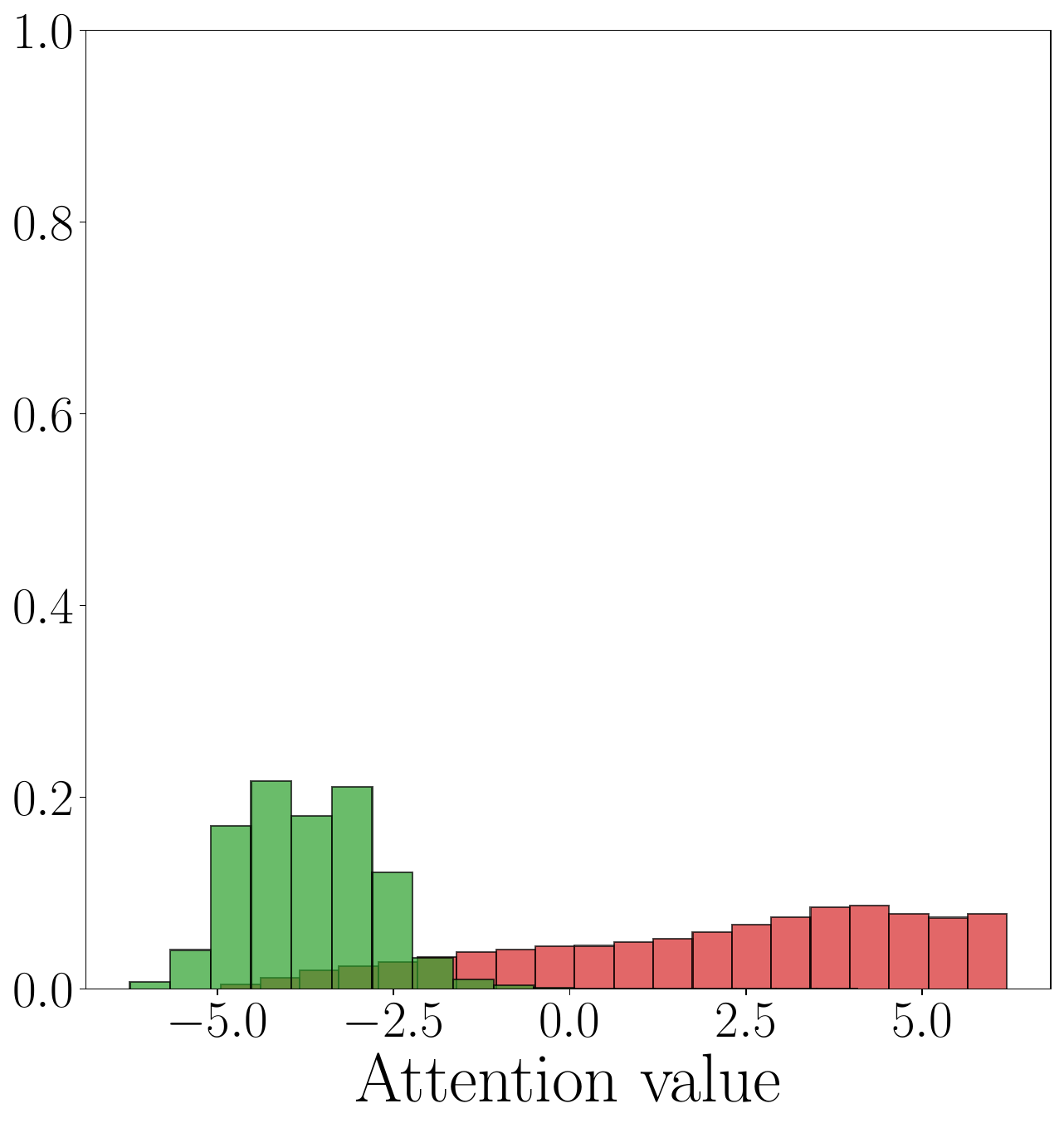}
        & 
        \includegraphics[trim={0cm 0cm 0cm 0cm},clip,width=0.18\textwidth]
        {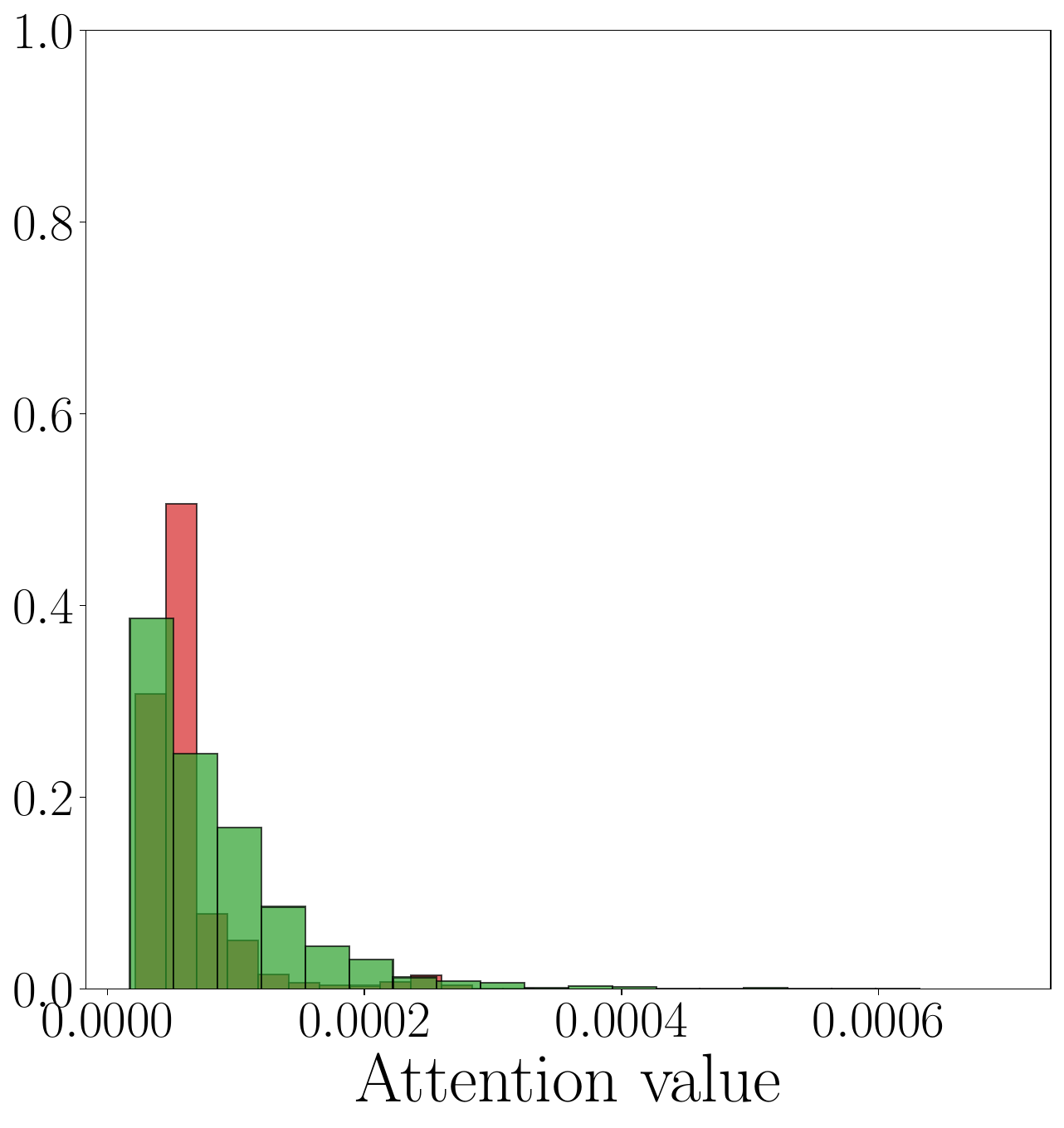}
        & 
        \includegraphics[trim={0cm 0cm 0cm 0cm},clip,width=0.18\textwidth]{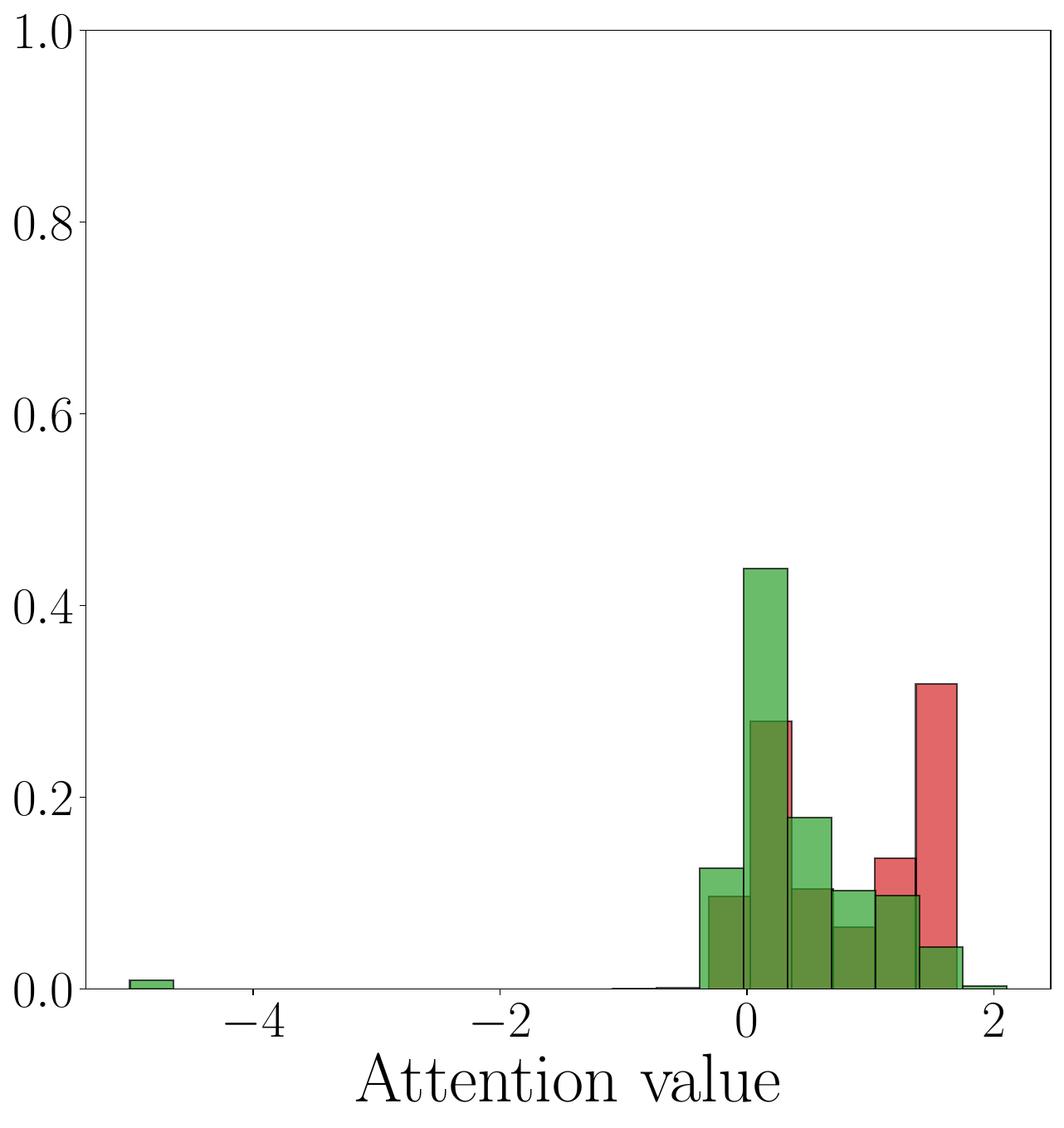}
        & 
        \includegraphics[trim={0cm 0cm 0cm 0cm},clip,width=0.18\textwidth]
        {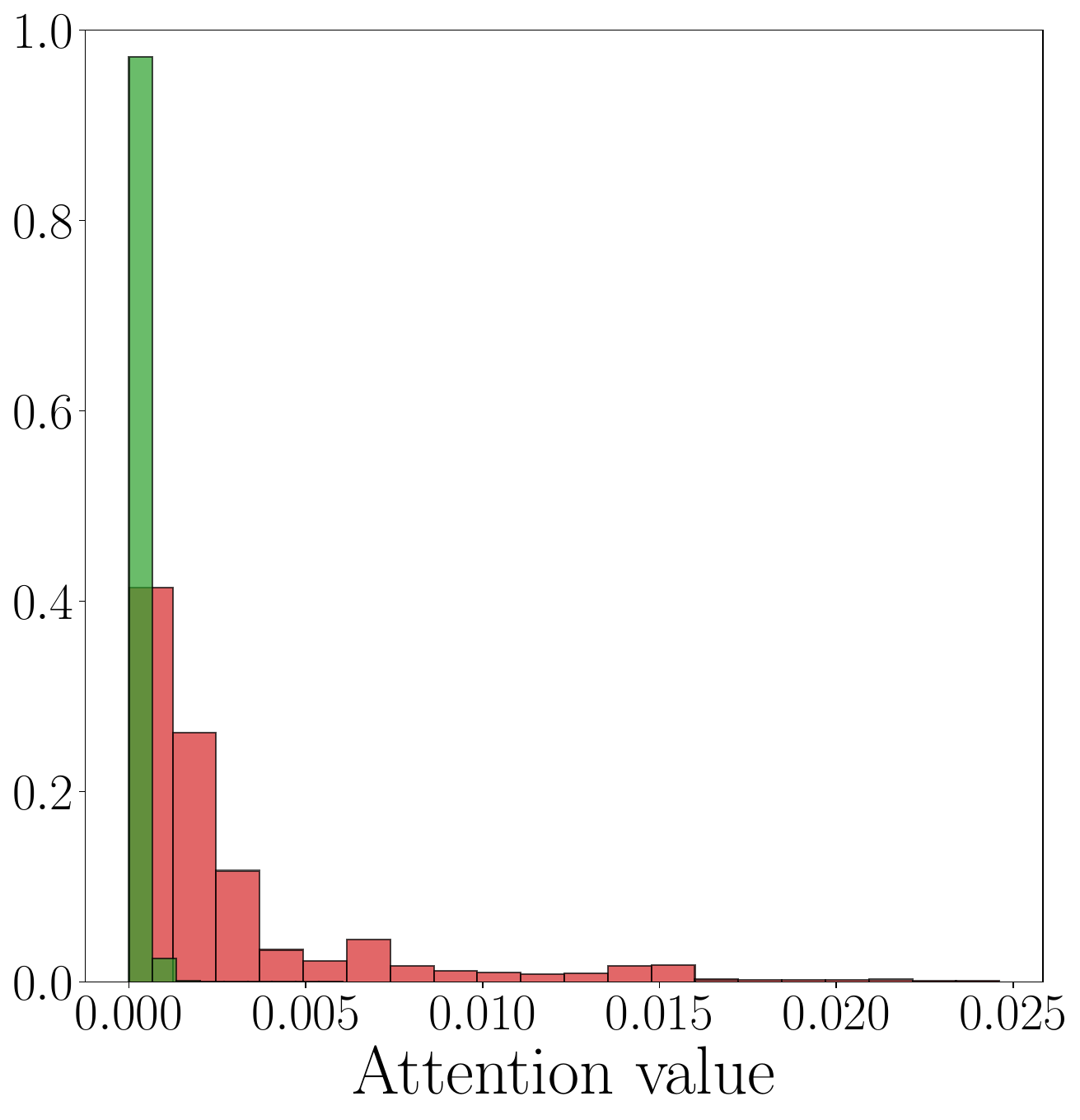}
        &
        \includegraphics[trim={0cm 0cm 0cm 0cm},clip,width=0.18\textwidth]
        {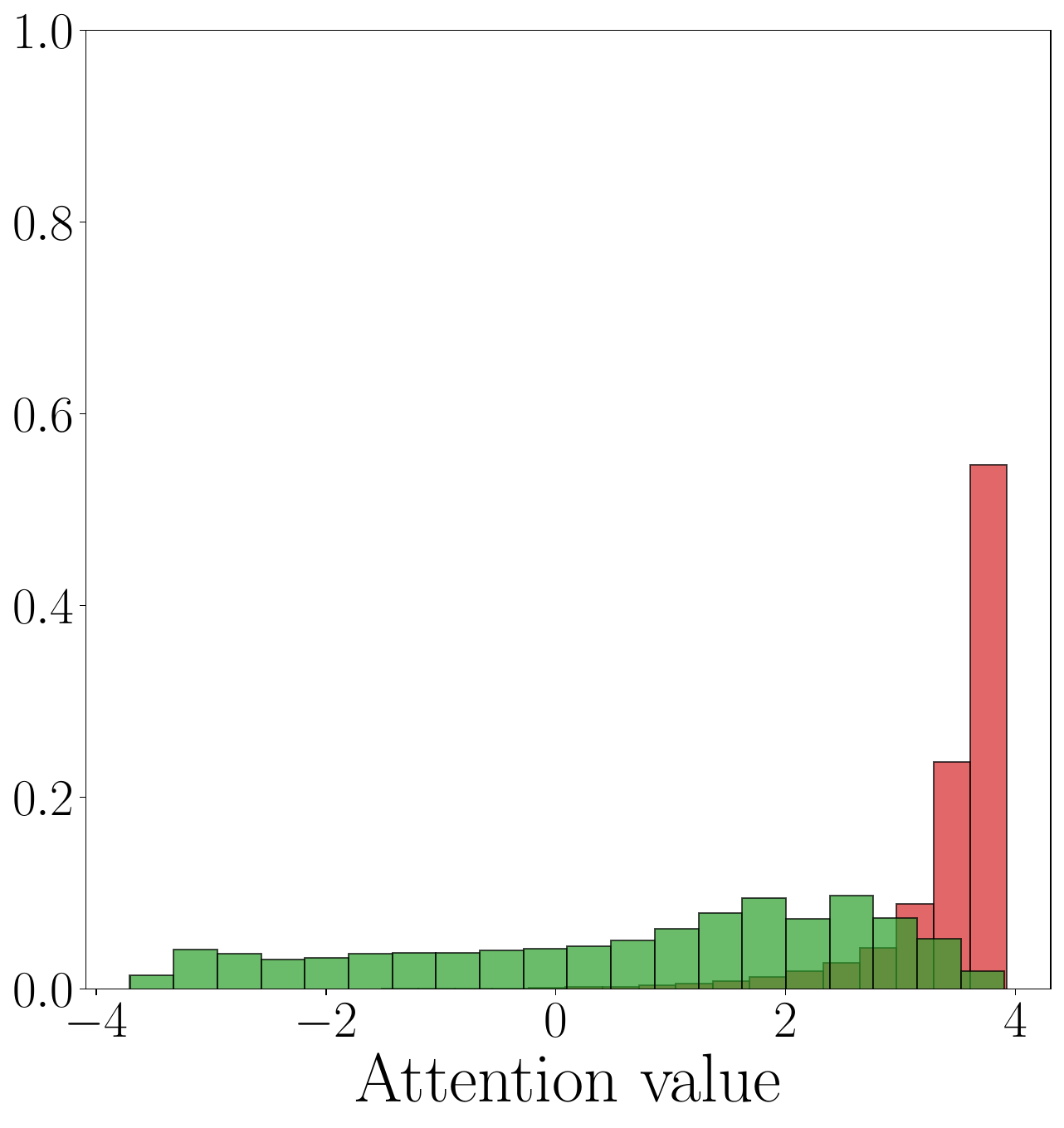}
        \\
        \smoothtransformerattpool & \transmil & \setmil & \gtp & \camil
    \end{tabular}
    \end{adjustbox}
    \caption{
    Attention histograms on CAMELYON16.
    First/second rows show models without/with global interactions. \smoothattpool\ and \smoothtransformerattpool\ stand out at separating positive and negative instances.
    }
    \label{fig:attval_histograms-camelyon}
    \vspace{-3mm}
\end{figure}

\begin{figure}%[h]
    \centering
    \begin{adjustbox}{width=\textwidth}
    \begin{tabular}{cccccc}
        % \includegraphics[trim={0cm 0cm 0cm 0cm},clip,width=0.12\textwidth]
        % {img/camelyon_wsi.pdf}
        % & 
        \includegraphics[trim={0cm 0cm 0cm 0cm},clip,width=0.17\textwidth]
        {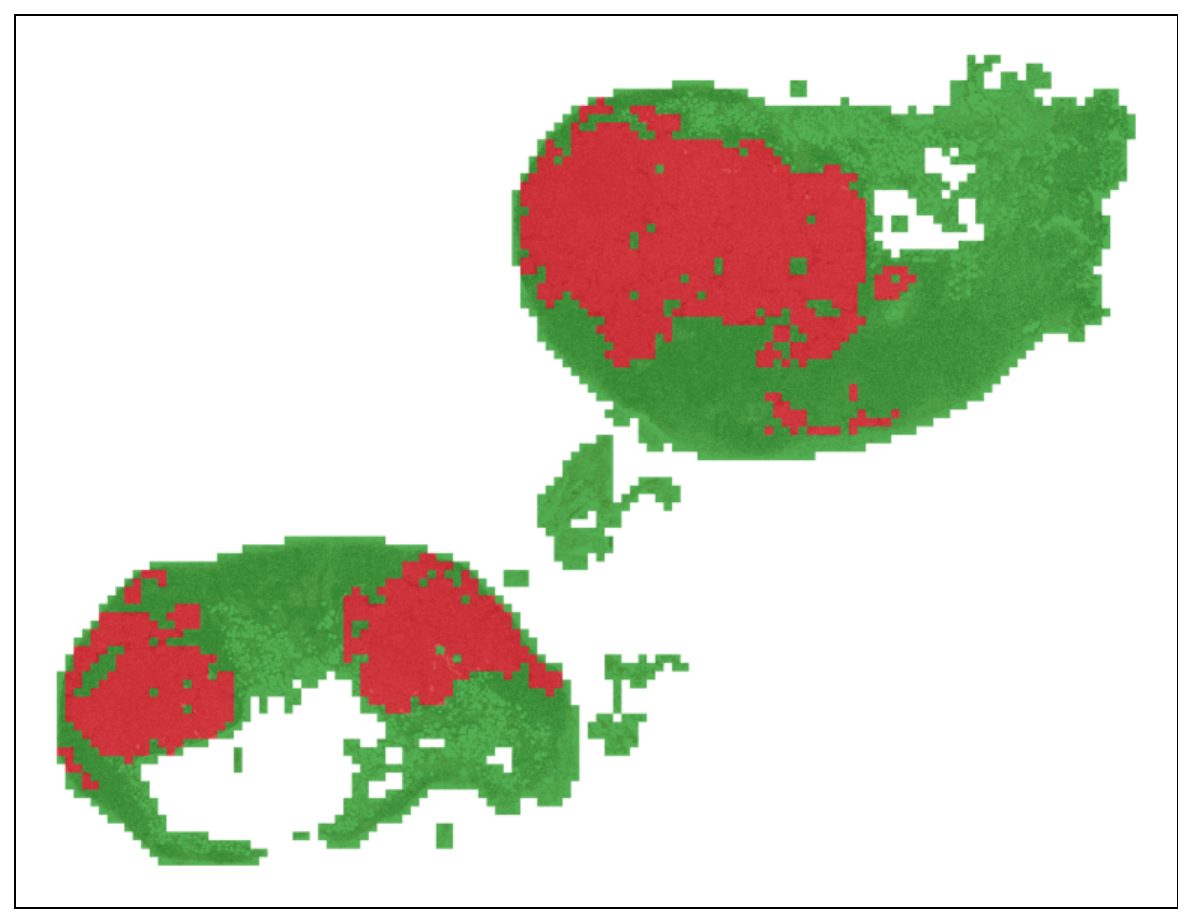}
        & 
        \includegraphics[trim={0cm 0cm 0cm 0cm},clip,width=0.17\textwidth]
        {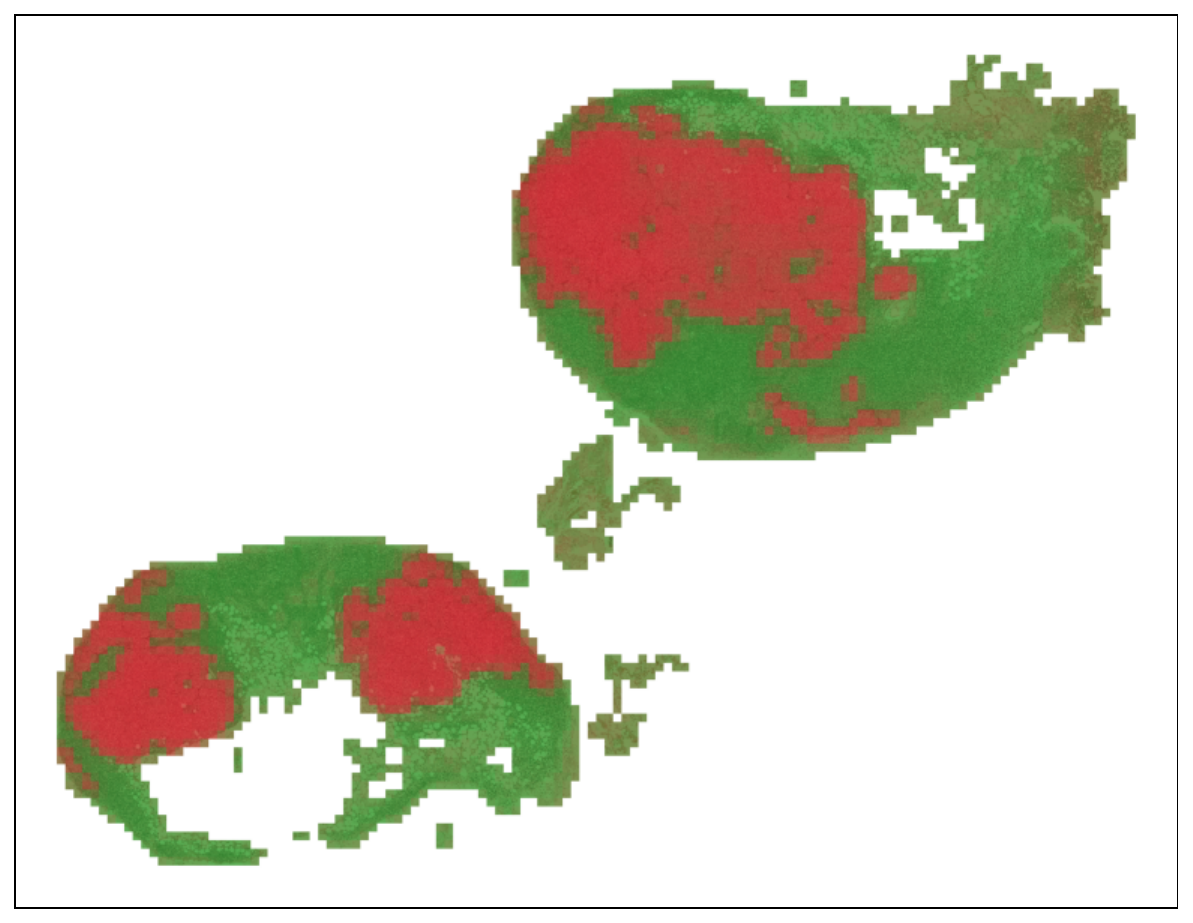}
        & 
        \includegraphics[trim={0cm 0cm 0cm 0cm},clip,width=0.17\textwidth]
        {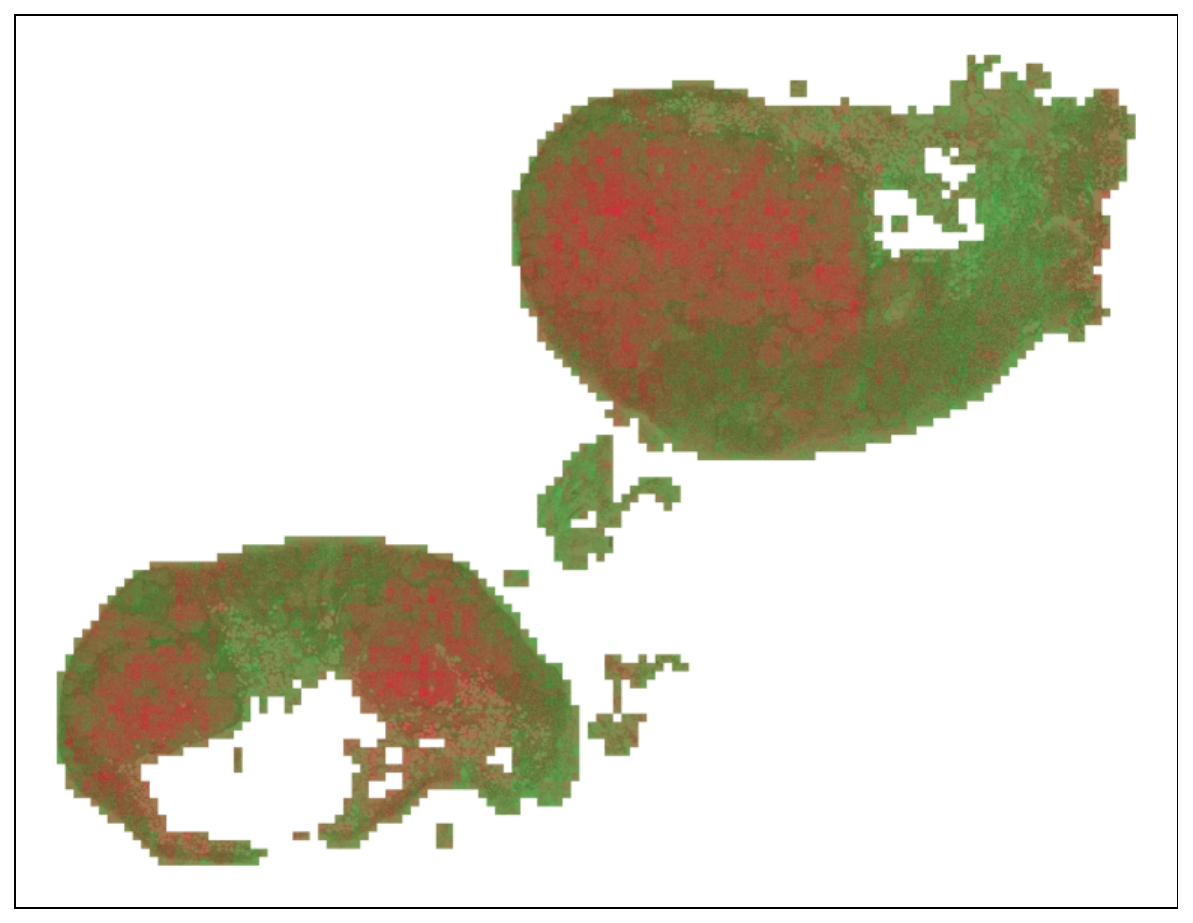}
        & 
        % \includegraphics[trim={0cm 0cm 0cm 0cm},clip,width=0.12\textwidth]
        % {img/camelyon_wsi_patched_setmil.pdf}
        % & 
        \includegraphics[trim={0cm 0cm 0cm 0cm},clip,width=0.17\textwidth]
        {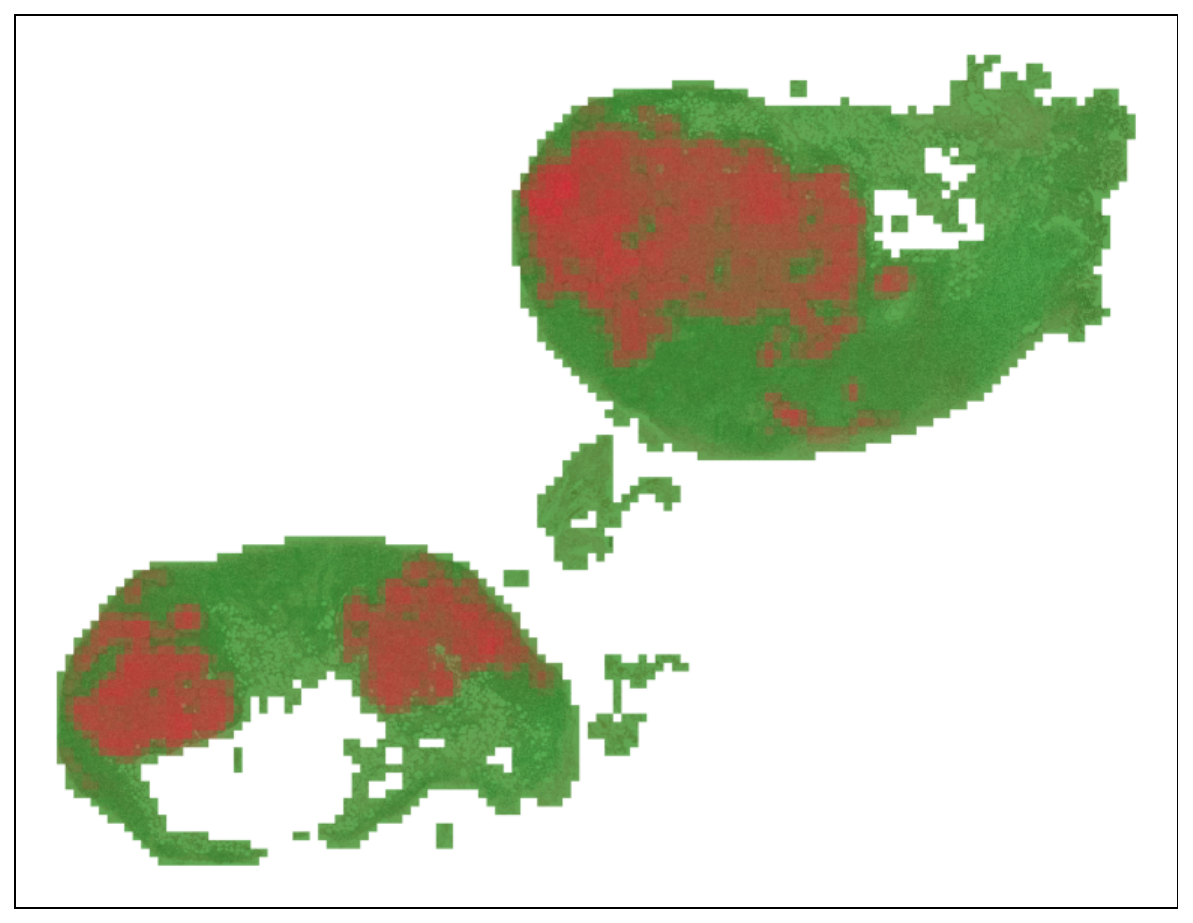}
        & 
        \includegraphics[trim={0cm 0cm 0cm 0cm},clip,width=0.17\textwidth]
        {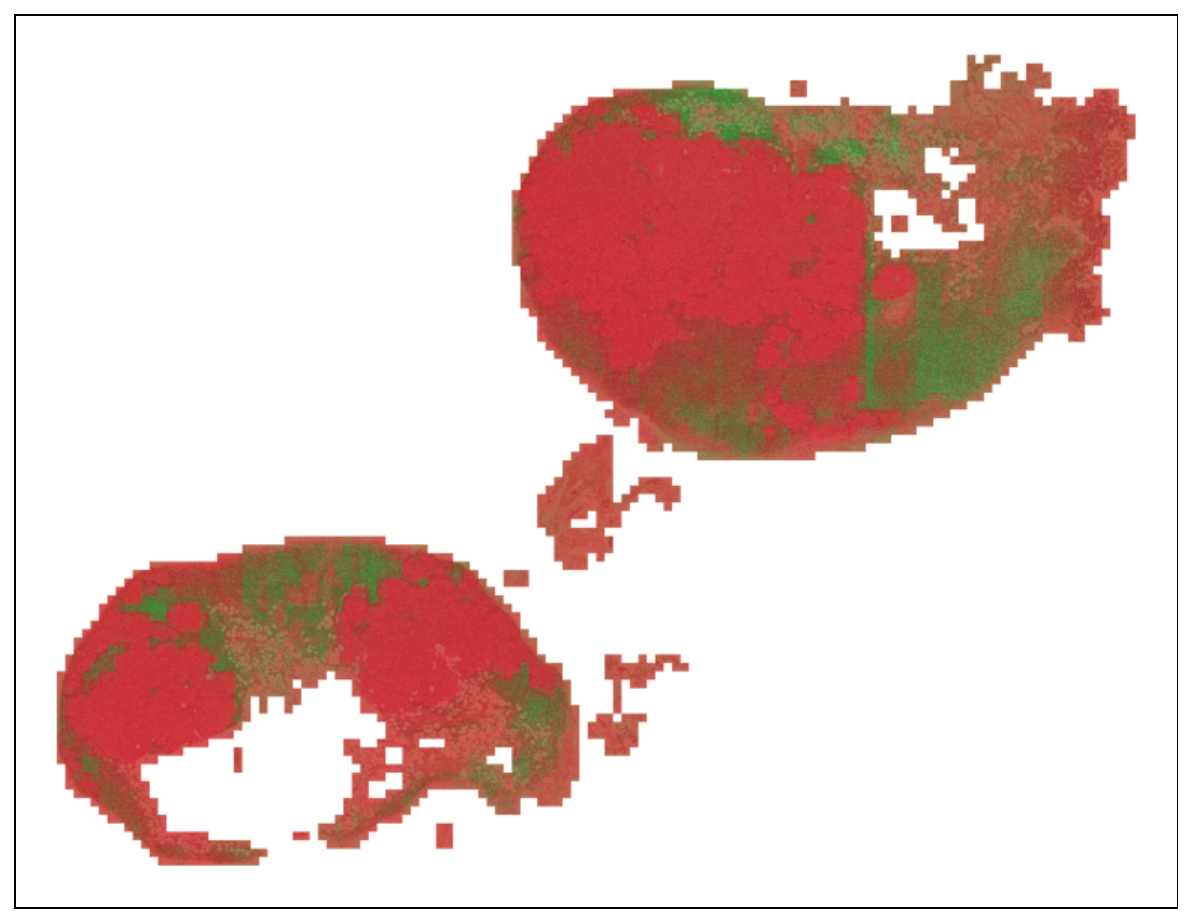}
        % &
        \includegraphics[trim={0cm 0cm 0cm 0cm},clip,height=54pt]
        {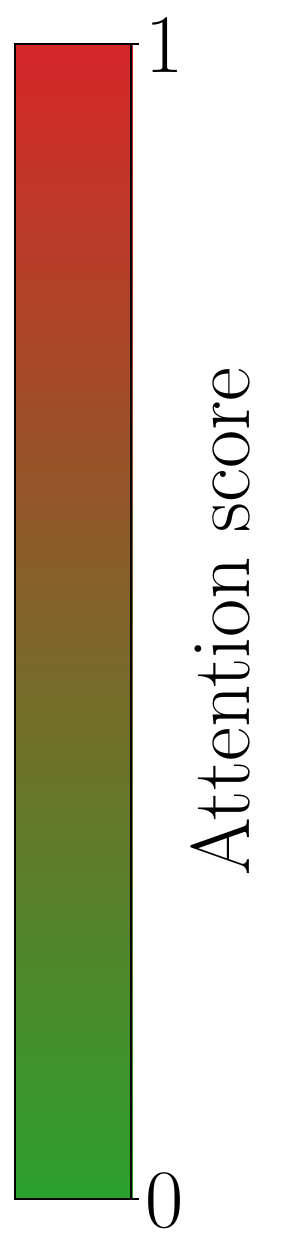}
        \\
        %  WSI &  Patch labels &  \smoothtransformerattpool. &  \transmil. &  \gtp. &  \camil. 
         Patch labels &  \smoothtransformerattpool &  \transmil &  \gtp &  \camil
    \end{tabular}
    \end{adjustbox}
    \caption{
    Attention maps on CAMELYON16.
    The novel \smoothtransformerattpool\ produces the most accurate one. 
    % Due to space limitations, \setmil\ is shown in \autoref{appendix:section:tables_figures}.
    }
    \label{fig:attmaps-camelyon}
     \vspace{-6mm}
\end{figure}

% heatmaps
\textbf{Attention maps.}
To visualize the localization differences, we show the attention maps generated by four of the transformer-based methods in a WSI from CAMELYON16, see \autoref{fig:attmaps-camelyon}. 
\smoothtransformerattpool\ attention map resembles the most to the ground truth. 
As noted in \autoref{fig:attval_histograms-camelyon}, \camil\ assigns high attention values to both positive and negative instances. 
\transmil\ and \gtp\ pinpoint the regions of interest, but the attention is relatively low in those areas, which produces unclear boundaries, especially in the case of \transmil.
The attention maps for the rest of the methods and datasets are in \autoref{appendix:section:tables_figures}.

The results shown in this subsection validate the utility of the smooth operator at the instance level. 
They suggest that having smooth attention maps is a powerful inductive bias that improves the instance-level performance.
In the following, we analyze its impact at the bag level. 

\vspace{-1mm}
\subsection{Classification: bag level results.}
\label{subsection:experiments-classification}

% In the previous subsection, we have shown that including the smooth operator produces more accurate attention maps, which improves the localization performance. 
% \pma{Probably this first sentence is not needed. If someone wants an overview of the experiments, they would go to the text in the beginning of section 5. But, once we have started subsections, there is no need to recall what the previous subsection was about, I think.}
In this subsection, we show that the use of the smooth operator does not deteriorate the bag classification results. 
On the contrary, in some cases, it improves them. 
Again, we focus on the AUROC and F1 scores, measured by comparing the true bag labels with the methods' bag label predictions. The threshold for the F1 score is $0.5$. We also report the mean rank achieved by each model.

\autoref{tab:bag-results} shows the results. 
The proposed models achieve the best performance overall.
As in the localization task, \abmil\ performs better than \smoothattpool\ in RSNA. Again, we believe it to be a consequence of the CT scan's low-complexity structure.
\dftdmil\ obtains the best result in CAMELYON16, but ranks second or third in the other two datasets.
\gtp\ and \setmil\ outperform the proposed \smoothtransformerattpool\ in PANDA, but their performance significantly decreases in CAMELYON16, obtaining the worst results.
Overall, our methods provide the most consistent performance, achieving an aggregated mean rank of 1.833. 

\begin{table}
\caption{
Classification results (mean and standard deviation from five independent runs).
The best is in bold and the second-best is underlined.
$(\downarrow)$/$(\uparrow)$ means lower/higher is better.
The models with the proposed operator achieve the best performance overall, ranking first or second in nine out of twelve dataset-score pairs.
}
\label{tab:bag-results}
\centering
\begin{adjustbox}{width=\textwidth}
\begin{tabular}{@{}ccccccccc@{}}
\toprule
& & \multicolumn{2}{c}{\textbf{RSNA}} & \multicolumn{2}{c}{\textbf{PANDA}} & \multicolumn{2}{c}{\textbf{CAMELYON16}} \\ \midrule
& & AUROC $(\uparrow)$ & F1 $(\uparrow)$ & AUROC $(\uparrow)$ & F1 $(\uparrow)$ & AUROC $(\uparrow)$ & F1 $(\uparrow)$ & Rank $(\downarrow)$\\ \midrule
\multirow{5}{*}{\makecell{Without\\global\\interactions}} & \smoothattpool & $0.888_{0.005}$ & $\underline{0.787}_{0.026}$ & $\mathbf{0.943}_{0.001}$ & $\mathbf{0.915}_{0.002}$ & $\underline{0.976}_{0.007}$ & $\underline{0.916}_{0.016}$ & $\mathbf{1.833}_{0.753}$ \\
& \abmil & $\underline{0.889}_{0.005}$ & $\mathbf{0.796}_{0.011}$ & $0.933_{0.002}$ & $\underline{0.909}_{0.001}$ & $0.956_{0.011}$ & $0.914_{0.021}$ & $2.500_{1.049}$ \\
& \clam & $0.674_{0.157}$ & $0.161_{0.291}$ & $0.893_{0.026}$ & $0.868_{0.034}$ & $0.960_{0.029}$ & $0.897_{0.012}$ & $4.500_{0.837}$ \\
& \dsmil & $0.689_{0.063}$ & $0.240_{0.012}$ & $0.921_{0.008}$ & $0.904_{0.008}$ & $0.947_{0.076}$ & $0.866_{0.123}$ & $4.167_{0.753}$ \\
& \dftdmil & $\mathbf{0.890}_{0.045}$ & $0.775_{0.282}$ & $\underline{0.940}_{0.001}$ & $0.903_{0.002}$ & $\mathbf{0.983}_{0.01}$ & $\mathbf{0.937}_{0.013}$ & $\underline{2.000}_{1.265}$ \\
% \multirow{7}{*}{\makecell{Without\\global\\interactions}} & \smoothattpool & $0.888_{0.005}$ & $\underline{0.787}_{0.026}$ & $\mathbf{0.943}_{0.001}$ & $\mathbf{0.915}_{0.002}$ & $\underline{0.976}_{0.007}$ & $\underline{0.916}_{0.016}$ & $\mathbf{2.000}_{1.095}$ \\
% & \abmil & $\underline{0.889}_{0.005}$ & $\mathbf{0.796}_{0.011}$ & $0.933_{0.002}$ & $\underline{0.909}_{0.001}$ & $0.956_{0.011}$ & $0.912_{0.027}$ & $2.500_{1.049}$ \\
% & \deepgraphsurv & $0.848_{0.017}$ & $0.719_{0.036}$ & $0.837_{0.020}$ & $0.823_{0.024}$ & $0.870_{0.070}$ & $0.772_{0.056}$ & $6.000_{0.894}$ \\
% & \clam & $0.674_{0.157}$ & $0.161_{0.291}$ & $0.893_{0.026}$ & $0.868_{0.034}$ & $0.960_{0.029}$ & $0.897_{0.012}$ & $5.167_{1.602}$ \\
% & \dsmil & $0.689_{0.063}$ & $0.240_{0.012}$ & $0.921_{0.008}$ & $0.904_{0.008}$ & $0.947_{0.085}$ & $0.866_{0.136}$ & $4.833_{1.169}$ \\
% & \pathgcn & $0.888_{0.007}$ & $0.782_{0.064}$ & $0.848_{0.005}$ & $0.857_{0.003}$ & $0.575_{0.206}$ & $0.345_{0.352}$ & $5.333_{1.862}$ \\
% & \dftdmil & $\mathbf{0.890}_{0.045}$ & $0.775_{0.282}$ & $\underline{0.940}_{0.001}$ & $0.903_{0.002}$ & $\mathbf{0.983}_{0.010}$ & $\mathbf{0.937}_{0.013}$ & $\underline{2.167}_{1.472}$ \\
\midrule
\multirow{5}{*}{\makecell{With\\global\\interactions}} & \smoothtransformerattpool& $\mathbf{0.906}_{0.007}$ & $\mathbf{0.825}_{0.026}$ & $0.946_{0.003}$ & $0.917_{0.002}$ & $\underline{0.976}_{0.014}$ & $\mathbf{0.948}_{0.02}$ & $\mathbf{1.833}_{0.983}$ \\
& \transmil & $0.883_{0.008}$ & $0.716_{0.031}$ & $0.933_{0.010}$ & $0.895_{0.029}$ & $0.973_{0.018}$ & $0.911_{0.028}$ & $4.083_{0.917}$ \\
& \setmil & $0.869_{0.011}$ & $0.716_{0.036}$ & $\mathbf{0.974}_{0.003}$ & $\mathbf{0.946}_{0.003}$ & $0.715_{0.155}$ & $0.471_{0.341}$ & $3.583_{2.010}$ \\
& \gtp & $\underline{0.901}_{0.008}$ & $\underline{0.805}_{0.017}$ & $\underline{0.949}_{0.004}$ & $\underline{0.920}_{0.003}$ & $0.748_{0.118}$ & $0.727_{0.143}$ & $2.750_{0.987}$ \\
& \camil & $0.889_{0.019}$ & $0.805_{0.028}$ & $0.938_{0.003}$ & $0.911_{0.004}$ & $\mathbf{0.984}_{0.007}$ & $\underline{0.918}_{0.018}$ & $\underline{2.750}_{1.173}$ \\
% \multirow{6}{*}{\makecell{With\\global\\interactions}} & \smoothtransformerattpool & $\mathbf{0.906}_{0.007}$ & $\mathbf{0.825}_{0.026}$ & $0.946_{0.003}$ & $0.917_{0.002}$ & $\underline{0.976}_{0.014}$ & $\mathbf{0.948}_{0.020}$ & $\mathbf{1.833}_{0.983}$ \\
% & \transmil & $0.883_{0.008}$ & $0.716_{0.031}$ & $0.933_{0.010}$ & $0.895_{0.029}$ & $0.973_{0.018}$ & $0.911_{0.028}$ & $4.333_{0.516}$ \\
% & \setmil & $0.869_{0.011}$ & $0.716_{0.036}$ & $\mathbf{0.974}_{0.003}$ & $\mathbf{0.946}_{0.003}$ & $0.715_{0.155}$ & $0.471_{0.341}$ & $4.000_{2.366}$ \\
% & \gtp & $\underline{0.901}_{0.008}$ & $\underline{0.805}_{0.017}$ & $\underline{0.949}_{0.004}$ & $\underline{0.920}_{0.003}$ & $0.748_{0.118}$ & $0.727_{0.143}$ & $3.167_{1.472}$ \\
% & \iibmil & $0.868_{0.013}$ & $0.621_{0.050}$ & $0.931_{0.004}$ & $0.881_{0.012}$ & $0.974_{0.002}$ & $\underline{0.922}_{0.010}$ & $4.833_{1.835}$ \\
% & \camil & $0.889_{0.019}$ & $0.805_{0.028}$ & $0.938_{0.003}$ & $0.911_{0.004}$ & $\mathbf{0.984}_{0.007}$ & $0.918_{0.018}$ & $\underline{2.833}_{1.169}$ \\
\bottomrule
\end{tabular}
\end{adjustbox}
\vspace{-3mm}
\end{table}

% \begin{table}[h]
% \caption{Classification results.}
% \label{tab:bag-results}
% \centering
% \begin{tabular}{@{}ccccccc@{}}
% \toprule
% & \multicolumn{2}{c}{\textbf{RSNA}} & \multicolumn{2}{c}{\textbf{PANDA}} & \multicolumn{2}{c}{\textbf{CAMELYON16}} \\ \midrule
% & AUROC & F1 & AUROC & F1 & AUROC & F1 \\ \midrule

% \bottomrule
% \end{tabular}
% \end{table}

\vspace{-1mm}
\subsection{Ablation study}
\label{subsection:experiments-ablation}

The proposed \smoothopp\ comes with different design choices and hyperparameters: the placement of \smoothopp, the trade-off parameter $\alpha$, the number of approximation steps $T$, and the use of spectral normalization. 
% The analysis for the first three is in \autoref{appendix:section:exp_details}. 
\textcolor{review}{
    We analyze them in the following, showing that \smoothopp\ leads to enhanced results almost under any choice. 
    This supports that our hypothesis --- that neighboring instances are likely to have the same label --- is a powerful inductive bias worth exploring.
}

\subsubsection{Placement of \smoothopp}

Recall that \smoothattpool\ leverages by default the \emph{early} variation, but we also described \smoothattpoolmid\ and \smoothattpoollate. Likewise, we discussed different variants for \smoothtransformerattpool. 
\autoref{tab:ablation-results} summarizes the impact of these choices on the final performance. 
% For each variant, we show the AUROC at bag and instance levels, along with the normalized Dirichlet energy of the attention values.
% The influence of the rest of the hyperparameters is studied in \autoref{appendix:section:exp_details}.
% For a more comprehensive table, see \autoref{appendix:section:tables_figures}. 

\textbf{\smoothopp\ without the transformer encoder (\smoothattpool).}
% \textbf{Variations on \smoothattpool.}
These variants differ in the place where \smoothopp\ is located inside the attention pool, recall \autoref{eq:smoothattpool_late}--\autoref{eq:smoothattpool_early}. 
We include \abmil\ since we build our model on top of it.
We see that using \smoothattpool\ improves the performance at both instance and bag levels. 
This improvement is more noticeable in PANDA and CAMELYON. 
We attribute it to the bag graph structure being more complex in WSIs than in CT scans. %\pma{(nice point)}
Also, the Dirichlet energy is lower when the smooth operator is used, as theoretically expected.
We observe that the proposed method is robust to different placement configurations, which is consistent with the theoretical guarantees presented in \autoref{subsection:method-modelling_smoothness}. 
However, none of the variants consistently outperforms the others. 
% \pma{The method is robust to the placement in different places, which is consistent with the theoretical guarantees based on Lipschitz etc. Partly, the idea here is to show that we are not doing ``tricks'' about where to place the operator so that it works.}

\textbf{\smoothopp\ with the transformer encoder (\smoothtransformerattpool).}
Recall that \smoothtransformerattpool\ leverages \smoothopp\ both after the transformer encoder and inside the attention pooling. Here we will refer to it as \smoothtransformer+\smoothattpool, and will compare it against T+\smoothattpool\ and \smoothtransformer+AP (using \smoothopp\ only in one of the components) and against T+AP (not using \smoothopp).
%In the second group, we ablate the use of the smooth operator both in the transformer encoder (T vs \smoothtransformer) and inside the attention pooling (AP vs \smoothattpool).
We observe that \smoothopp\ has no significant effect on bag-level performance.
%, which suggests that \smoothopp\ models local dependencies without interfering with global dependencies. 
At instance-level we do observe differences: the baseline T+AP is outperformed as long as \smoothopp\ is used within the attention pooling.
% The second row shows that placing \smoothopp\ only in the transformer encoder worsens the performance at the instance level.
% When \smoothopp\ is placed only in the attention pooling (third row), there is an improvement at the instance level. 
% Finally, placing \smoothopp\ in both the transformer encoder and the attention pooling (first row) improves the instance-level performance in PANDA, but not in CAMELYON16. 

\begin{table}
\caption{Ablation study on different configurations of our models. AUROC (at both instance and bag levels), and normalized Dirichlet energy of attention values are reported. Almost all configurations improve the results in both tasks against the baseline (not using \smoothopp).}
\label{tab:ablation-results}
\centering
\begin{adjustbox}{width=\textwidth}
\begin{tabular}{cccccccccc}
\toprule
& \multicolumn{3}{c}{\textbf{RSNA}} & \multicolumn{3}{c}{\textbf{PANDA}} & \multicolumn{3}{c}{\textbf{CAMELYON16}} \\ \midrule
& $\text{AUROC}_\text{Inst} (\uparrow)$ & $\text{AUROC}_\text{Bag} (\uparrow)$ & $\direnergy{\bff}$ & $\text{AUROC}_\text{Inst} (\uparrow)$ & $\text{AUROC}_\text{Bag} (\uparrow)$ & $\direnergy{\bff}$ & $\text{AUROC}_\text{Inst} (\uparrow)$ & $\text{AUROC}_\text{Bag} (\uparrow)$ & $\direnergy{\bff}$\\ \midrule
\smoothattpoolearly & $0.798$ & $0.888$ & $0.009$ & $0.799$ & $0.943$ & $0.106$ & $0.960$ & $0.976$ & $0.395$ \\
\smoothattpoolmid & $0.806$ & $0.888$ & $0.012$ & $0.792$ & $0.940$ & $0.135$ & $0.922$ & $0.964$ & $0.384$ \\
\smoothattpoollate & $0.811$ & $0.891$ & $0.011$ & $0.802$ & $0.944$ & $0.082$ & $0.819$ & $0.964$ & $0.321$ \\
\abmil & $0.806$ & $0.889$ & $0.023$ & $0.768$ & $0.933$ & $0.141$ & $0.819$ & $0.956$ & $0.419$ \\
\midrule
\smoothtransformer+\smoothattpool & $0.791$ & $0.910$ & $0.010$ & $0.813$ & $0.944$ & $0.306$ & $0.841$ & $0.986$ & $0.313$ \\
\smoothtransformer+\attpool & $0.791$ & $0.910$ & $0.010$ & $0.754$ & $0.940$ & $0.356$ & $0.754$ & $0.984$ & $0.320$ \\
\transformer+\smoothattpool & $0.792$ & $0.910$ & $0.010$ & $0.787$ & $0.944$ & $0.332$ & $0.915$ & $0.986$ & $0.343$ \\
\transformer+\attpool & $0.792$ & $0.910$ & $0.020$ & $0.760$ & $0.942$ & $0.391$ & $0.781$ & $0.984$ & $0.433$ \\
\bottomrule
\end{tabular}
\end{adjustbox}
% \vspace{-2mm}
\end{table}

\subsubsection{\smoothopp\ hyperparameters}

In the following we study the influence of the trade-off parameter $\alpha$ and of the spectral normalization. 
Due to space limitations, the analysis for the number of approximation steps $T$ is in \autoref{appendix:subsection:ablation}.

\begin{figure}
    \centering
    \begin{subfigure}{0.45\textwidth}
        \includegraphics[trim={0cm 0cm 0cm 0cm},clip,width=1.0\textwidth]{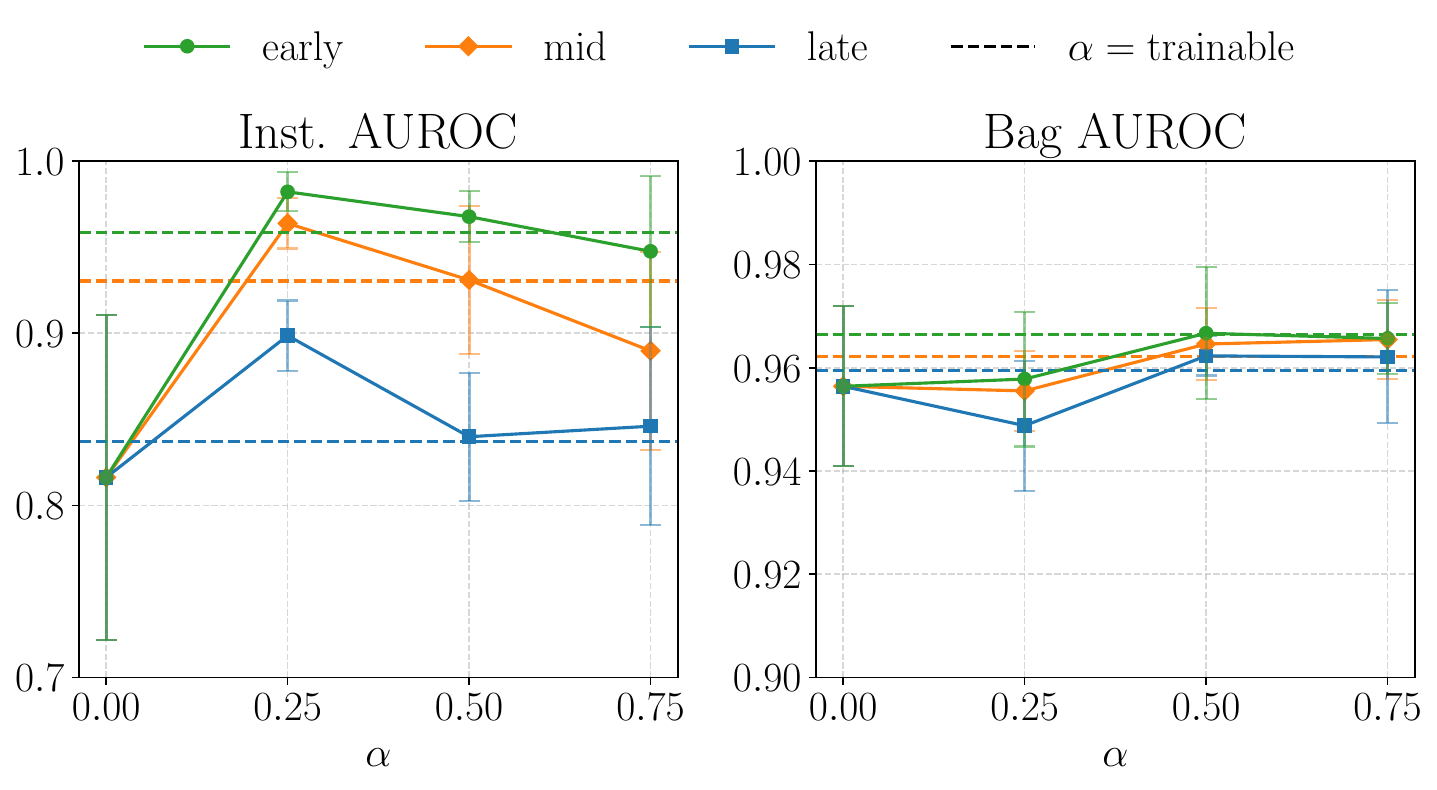}
        \vspace{-5mm}
        \caption{Trade-off parameter $\alpha$.}
        \label{fig:ablation_alpha}
        \vspace{-1mm}
    \end{subfigure}
    \hfill
    \begin{subfigure}{0.45\textwidth}
        \includegraphics[trim={0cm 0cm 0cm 0cm},clip,width=1.0\textwidth]{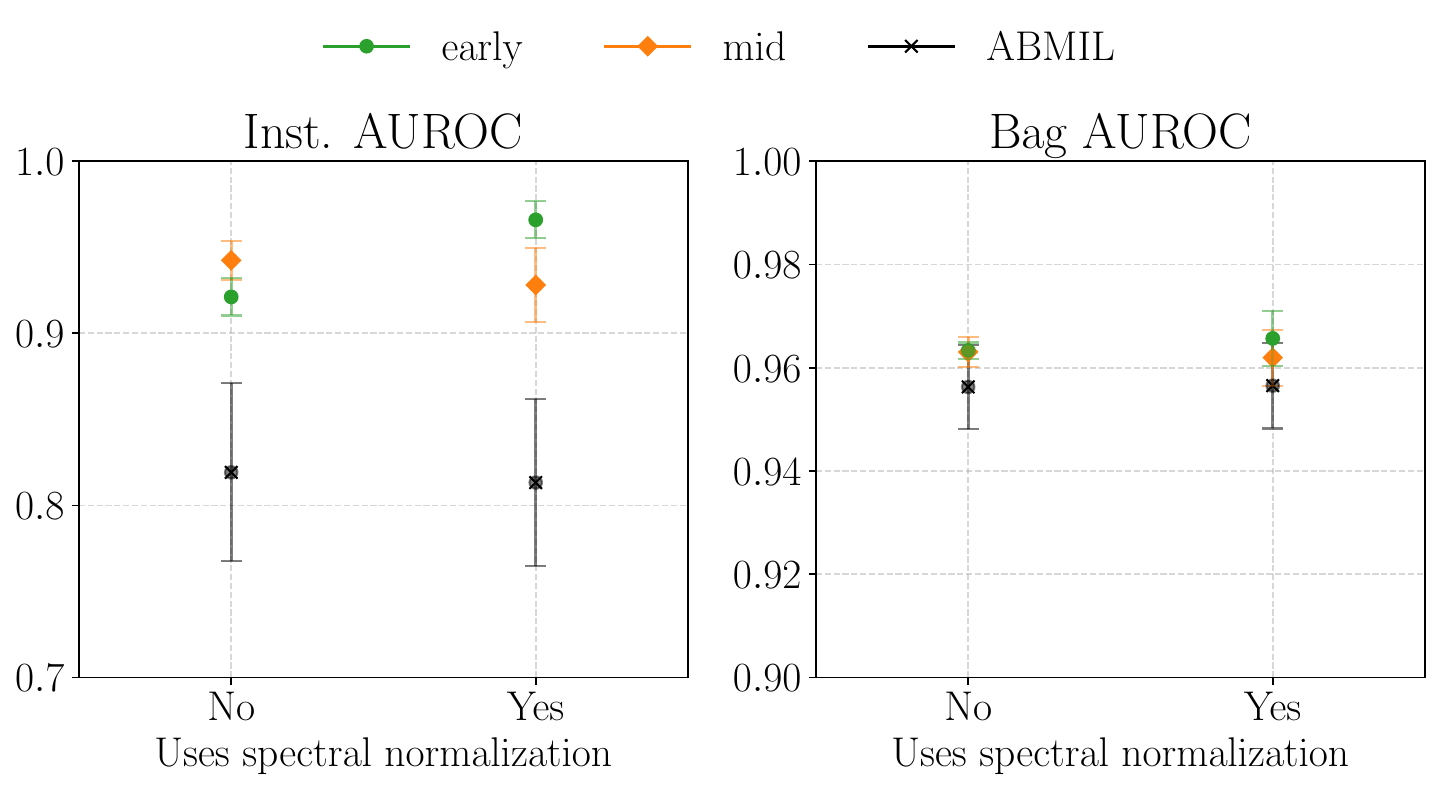}
        \vspace{-5mm}
        \caption{Spectral normalization.}
        \label{fig:ablation_spectralnorm}
        \vspace{-1mm}
    \end{subfigure}
    \caption{
        Influence of the trade-off parameter $\alpha$ (left) and of spectral normalization (right) in CAMELYON16.
        Setting $\alpha>0$ improves upon the baseline \abmil\ ($\alpha=0$) and is a trade-off between better localization results (lower $\alpha$) or better classification results (higher $\alpha$).
        Likewise, \smoothopp\ without spectral normalization already improves the results upon the baseline (\abmil), but the best performance is obtained when they are used together.
    }
    \label{fig:ablation_alpha_spectralnorm}
    \vspace{-3mm}
\end{figure}

\textbf{The trade-off parameter $\alpha$.} 
From \autoref{eq:energy} we see that $\alpha \in \left[0,1\right)$ controls the \textit{amount of smoothness} enforced by \smoothopp. 
Note that $\alpha=0$ in \autoref{eq:smoothop_exact} produces no smoothness, turning \smoothopp\ into the identity operator.
In \autoref{fig:ablation_alpha} we show the performance obtained for different values of $\alpha$ in CAMELYON16. 
Each choice of this hyperparameter improves upon the baseline \abmil\ ($\alpha=0$). 
We see that better localization results are obtained when $\alpha$ is lower, while better classification results are obtained when $\alpha$ is higher. 
Fixing $\alpha=0.5$ is a compromise between the two, and produces very similar results as setting it as a trainable parameter initialized at $\alpha=0.5$. 
\textcolor{review}{
    \autoref{fig:attmaps-camelyon-alpha-appendix} provides a visual comparison of the effect that $\alpha$ has on the attention maps. 
}

\textbf{The effect of spectral normalization.}
Spectral normalization forces the norm of the multi-layer perceptron weights to remain constant. 
In this work, this is a key design choice that helps \smoothopp\ to obtain attention maps with lower Dirichlet energy. 
In our experiments, we have used spectral normalization in the layers immediately after \smoothopp. 
Note that the late variant does not require spectral normalization, since it applies \smoothopp\ directly to the attention values. 
In \autoref{fig:ablation_spectralnorm} we show the results obtained with and without spectral normalization in CAMELYON16. 
We observe that, even without spectral normalization, \smoothopp\ improves upon the baseline. 
The improvement is more significant when \smoothopp\ is paired with spectral normalization, especially at the instance level. 

\vspace{-1mm}
\section{Discussion and conclusion}
\label{section:discussion_conclusion}

% What we have done:
% - A unified perspective on attention MIL
% - Smooth operator to introduce local interactions in a principled way. Our operator produces smooth attention maps. 
% - Exhaustive study of ours and other methods at both instance and bag level. 
% In this work, we showed that current attention-based MIL methods differ in how instances interact in the encoding and aggregation process (no interactions, global interactions, and global and local interactions). 
The main goal of this paper is to draw attention to the study of MIL methods at the instance level. 
To that end, we revised current deep MIL methods and provided a unified perspective on them.
We proposed the smooth operator \smoothopp\ to introduce local interactions in a principled way. 
By design, it produces smooth attention maps that resemble the ground truth instance labels. 
We conducted an exhaustive experimental validation with three real-world MIL datasets and up to eight state-of-the-art methods in both classification and localization tasks.
This study showed that our method provides the best performance in localization while being highly competitive (best or second best) at classification. 

Despite its advantages, our method has some limitations. 
The first is that, as with every other operator in GNNs, the computational costs of the smooth operator scale with the size of the bag. 
Fortunately, it can be paired with existing subgraph sampling techniques to mitigate this problem. 
The second limitation is that we do not have a definite answer for where it is better to use the proposed operator. We have shown that it leads to improvements in almost every place, but the optimal location may be problem-dependent and has to be tailored by the practitioner.  

Finally, we hope that our work will draw more attention to the localization problem, which is very important for the deployment of computer-aided systems in the real world. 
In this sense, safely deploying the proposed methods in clinical practice requires evaluating them in a wider range of medical problems and quantifying their uncertainty. 
For the latter, we believe that the smooth operator could also benefit from a probabilistic formulation. 

\section*{Acknowledgements}

\textcolor{review}{
    This work was supported by project PID2022-140189OB-C22 funded by MCIN / AEI / 10.13039 / 501100011033. 
    Francisco M. Castro-Macías acknowledges FPU contract FPU21/01874 funded by Ministerio de Universidades. 
    Pablo Morales-Álvarez acknowledges grant C-EXP-153-UGR23 funded by Consejería de Universidad, Investigación e Innovación and by the European Union (EU) ERDF Andalusia Program 2021-2027.
}

\bibliographystyle{plainnat}
\bibliography{references}

\begin{thebibliography}{44}
\providecommand{\natexlab}[1]{#1}
\providecommand{\url}[1]{\texttt{#1}}
\expandafter\ifx\csname urlstyle\endcsname\relax
  \providecommand{\doi}[1]{doi: #1}\else
  \providecommand{\doi}{doi: \begingroup \urlstyle{rm}\Url}\fi

\bibitem[Aggarwal et~al.(2021)Aggarwal, Sounderajah, Martin, Ting,
  Karthikesalingam, King, Ashrafian, and Darzi]{aggarwal2021diagnostic}
Ravi Aggarwal, Viknesh Sounderajah, Guy Martin, Daniel~SW Ting, Alan
  Karthikesalingam, Dominic King, Hutan Ashrafian, and Ara Darzi.
\newblock Diagnostic accuracy of deep learning in medical imaging: a systematic
  review and meta-analysis.
\newblock \emph{NPJ digital medicine}, 4\penalty0 (1):\penalty0 65, 2021.

\bibitem[Bejnordi et~al.(2017)Bejnordi, Veta, Van~Diest, Van~Ginneken,
  Karssemeijer, Litjens, Van Der~Laak, Hermsen, Manson, Balkenhol,
  et~al.]{bejnordi2017diagnostic}
Babak~Ehteshami Bejnordi, Mitko Veta, Paul~Johannes Van~Diest, Bram
  Van~Ginneken, Nico Karssemeijer, Geert Litjens, Jeroen~AWM Van Der~Laak,
  Meyke Hermsen, Quirine~F Manson, Maschenka Balkenhol, et~al.
\newblock Diagnostic assessment of deep learning algorithms for detection of
  lymph node metastases in women with breast cancer.
\newblock \emph{Jama}, 318\penalty0 (22):\penalty0 2199--2210, 2017.

\bibitem[Belkin and Niyogi(2002)]{belkin2002semi}
Mikhail Belkin and Partha Niyogi.
\newblock Semi-supervised learning on manifolds.
\newblock \emph{Machine Learning Journal}, 1, 2002.

\bibitem[Bishop and Bishop(2023)]{bishop2024deep}
Christopher~Michael Bishop and Hugh Bishop.
\newblock \emph{Deep Learning - Foundations and Concepts}.
\newblock 2023.

\bibitem[Bulten et~al.(2022)Bulten, Kartasalo, Chen, Str{\"o}m, Pinckaers,
  Nagpal, Cai, Steiner, Van~Boven, Vink, et~al.]{bulten2022artificial}
Wouter Bulten, Kimmo Kartasalo, Po-Hsuan~Cameron Chen, Peter Str{\"o}m, Hans
  Pinckaers, Kunal Nagpal, Yuannan Cai, David~F Steiner, Hester Van~Boven,
  Robert Vink, et~al.
\newblock Artificial intelligence for diagnosis and gleason grading of prostate
  cancer: the panda challenge.
\newblock \emph{Nature medicine}, 28\penalty0 (1):\penalty0 154--163, 2022.

\bibitem[Cai and Wang(2020)]{cai2020note}
Chen Cai and Yusu Wang.
\newblock A note on over-smoothing for graph neural networks.
\newblock \emph{arXiv preprint arXiv:2006.13318}, 2020.

\bibitem[Campanella et~al.(2019)Campanella, Hanna, Geneslaw, Miraflor, Werneck
  Krauss~Silva, Busam, Brogi, Reuter, Klimstra, and
  Fuchs]{campanella2019clinical}
Gabriele Campanella, Matthew~G Hanna, Luke Geneslaw, Allen Miraflor, Vitor
  Werneck Krauss~Silva, Klaus~J Busam, Edi Brogi, Victor~E Reuter, David~S
  Klimstra, and Thomas~J Fuchs.
\newblock Clinical-grade computational pathology using weakly supervised deep
  learning on whole slide images.
\newblock \emph{Nature medicine}, 25\penalty0 (8):\penalty0 1301--1309, 2019.

\bibitem[Carbonneau et~al.(2018)Carbonneau, Cheplygina, Granger, and
  Gagnon]{carbonneau2018multiple}
Marc-Andr{\'e} Carbonneau, Veronika Cheplygina, Eric Granger, and Ghyslain
  Gagnon.
\newblock Multiple instance learning: A survey of problem characteristics and
  applications.
\newblock \emph{Pattern Recognition}, 77:\penalty0 329--353, 2018.

\bibitem[Cersovsky et~al.(2023)Cersovsky, Mohammadi, Kainmueller, and
  Hoehne]{cersovsky2023towards}
Josef Cersovsky, Sadegh Mohammadi, Dagmar Kainmueller, and Johannes Hoehne.
\newblock Towards hierarchical regional transformer-based multiple instance
  learning.
\newblock In \emph{Proceedings of the IEEE/CVF International Conference on
  Computer Vision}, pages 3952--3960, 2023.

\bibitem[Chapelle et~al.(2002)Chapelle, Weston, and
  Sch{\"o}lkopf]{chapelle2002cluster}
Olivier Chapelle, Jason Weston, and Bernhard Sch{\"o}lkopf.
\newblock Cluster kernels for semi-supervised learning.
\newblock \emph{Advances in neural information processing systems}, 15, 2002.

\bibitem[Chen et~al.(2021)Chen, Lu, Shaban, Chen, Chen, Williamson, and
  Mahmood]{chen2021whole}
Richard~J Chen, Ming~Y Lu, Muhammad Shaban, Chengkuan Chen, Tiffany~Y Chen,
  Drew~FK Williamson, and Faisal Mahmood.
\newblock Whole slide images are 2d point clouds: Context-aware survival
  prediction using patch-based graph convolutional networks.
\newblock In \emph{Medical Image Computing and Computer Assisted
  Intervention--MICCAI 2021: 24th International Conference, Strasbourg, France,
  September 27--October 1, 2021, Proceedings, Part VIII 24}, pages 339--349.
  Springer, 2021.

\bibitem[Dietterich et~al.(1997)Dietterich, Lathrop, and
  Lozano-P{\'e}rez]{dietterich1997solving}
Thomas~G Dietterich, Richard~H Lathrop, and Tom{\'a}s Lozano-P{\'e}rez.
\newblock Solving the multiple instance problem with axis-parallel rectangles.
\newblock \emph{Artificial intelligence}, 89\penalty0 (1-2):\penalty0 31--71,
  1997.

\bibitem[Flanders et~al.(2020)Flanders, Prevedello, Shih, Halabi,
  Kalpathy-Cramer, Ball, Mongan, Stein, Kitamura, Lungren,
  et~al.]{flanders2020construction}
Adam~E Flanders, Luciano~M Prevedello, George Shih, Safwan~S Halabi, Jayashree
  Kalpathy-Cramer, Robyn Ball, John~T Mongan, Anouk Stein, Felipe~C Kitamura,
  Matthew~P Lungren, et~al.
\newblock Construction of a machine learning dataset through collaboration: the
  rsna 2019 brain ct hemorrhage challenge.
\newblock \emph{Radiology: Artificial Intelligence}, 2\penalty0 (3):\penalty0
  e190211, 2020.

\bibitem[Fourkioti et~al.(2024)Fourkioti, Vries, and Bakal]{fourkioti2023camil}
Olga Fourkioti, Matt~De Vries, and Chris Bakal.
\newblock {CAMIL}: Context-aware multiple instance learning for cancer
  detection and subtyping in whole slide images.
\newblock In \emph{The Twelfth International Conference on Learning
  Representations}, 2024.
\newblock URL \url{https://openreview.net/forum?id=rzBskAEmoc}.

\bibitem[Fu et~al.(2022)Fu, Zhao, and Bian]{fu2022p}
Guoji Fu, Peilin Zhao, and Yatao Bian.
\newblock $ p $-laplacian based graph neural networks.
\newblock In \emph{International Conference on Machine Learning}, pages
  6878--6917. PMLR, 2022.

\bibitem[Gasteiger et~al.(2018)Gasteiger, Bojchevski, and
  G{\"u}nnemann]{gasteiger2018predict}
Johannes Gasteiger, Aleksandar Bojchevski, and Stephan G{\"u}nnemann.
\newblock Predict then propagate: Graph neural networks meet personalized
  pagerank.
\newblock \emph{arXiv preprint arXiv:1810.05997}, 2018.

\bibitem[Ilse et~al.(2018)Ilse, Tomczak, and Welling]{ilse2018attention}
Maximilian Ilse, Jakub Tomczak, and Max Welling.
\newblock Attention-based deep multiple instance learning.
\newblock In \emph{International conference on machine learning}, pages
  2127--2136. PMLR, 2018.

\bibitem[Kang et~al.(2023)Kang, Song, Park, Yoo, and
  Pereira]{kang2023benchmarking}
Mingu Kang, Heon Song, Seonwook Park, Donggeun Yoo, and S{\'e}rgio Pereira.
\newblock Benchmarking self-supervised learning on diverse pathology datasets.
\newblock In \emph{Proceedings of the IEEE/CVF Conference on Computer Vision
  and Pattern Recognition}, pages 3344--3354, 2023.

\bibitem[Kingma and Ba(2014)]{kingma2014adam}
Diederik~P Kingma and Jimmy Ba.
\newblock Adam: A method for stochastic optimization.
\newblock \emph{arXiv preprint arXiv:1412.6980}, 2014.

\bibitem[Kipf and Welling(2016)]{kipf2016semi}
Thomas~N Kipf and Max Welling.
\newblock Semi-supervised classification with graph convolutional networks.
\newblock \emph{arXiv preprint arXiv:1609.02907}, 2016.

\bibitem[Li et~al.(2021{\natexlab{a}})Li, Li, and Eliceiri]{li2021dual}
Bin Li, Yin Li, and Kevin~W Eliceiri.
\newblock Dual-stream multiple instance learning network for whole slide image
  classification with self-supervised contrastive learning.
\newblock In \emph{Proceedings of the IEEE/CVF conference on computer vision
  and pattern recognition}, pages 14318--14328, 2021{\natexlab{a}}.

\bibitem[Li et~al.(2021{\natexlab{b}})Li, Yang, Zhao, Xing, Zhang, Gao, Huang,
  Wang, and Yao]{li2021dt}
Hang Li, Fan Yang, Yu~Zhao, Xiaohan Xing, Jun Zhang, Mingxuan Gao, Junzhou
  Huang, Liansheng Wang, and Jianhua Yao.
\newblock Dt-mil: deformable transformer for multi-instance learning on
  histopathological image.
\newblock In \emph{Medical Image Computing and Computer Assisted
  Intervention--MICCAI 2021: 24th International Conference, Strasbourg, France,
  September 27--October 1, 2021, Proceedings, Part VIII 24}, pages 206--216.
  Springer, 2021{\natexlab{b}}.

\bibitem[Li et~al.(2018{\natexlab{a}})Li, Han, and Wu]{li2018deeper}
Qimai Li, Zhichao Han, and Xiao-Ming Wu.
\newblock Deeper insights into graph convolutional networks for semi-supervised
  learning.
\newblock In \emph{Proceedings of the AAAI conference on artificial
  intelligence}, volume~32, 2018{\natexlab{a}}.

\bibitem[Li et~al.(2018{\natexlab{b}})Li, Yao, Zhu, Li, and Huang]{li2018graph}
Ruoyu Li, Jiawen Yao, Xinliang Zhu, Yeqing Li, and Junzhou Huang.
\newblock Graph cnn for survival analysis on whole slide pathological images.
\newblock In \emph{International Conference on Medical Image Computing and
  Computer-Assisted Intervention}, pages 174--182. Springer,
  2018{\natexlab{b}}.

\bibitem[Lu et~al.(2021)Lu, Williamson, Chen, Chen, Barbieri, and
  Mahmood]{lu2021data}
Ming~Y Lu, Drew~FK Williamson, Tiffany~Y Chen, Richard~J Chen, Matteo Barbieri,
  and Faisal Mahmood.
\newblock Data-efficient and weakly supervised computational pathology on
  whole-slide images.
\newblock \emph{Nature biomedical engineering}, 5\penalty0 (6):\penalty0
  555--570, 2021.

\bibitem[Miyato et~al.(2018)Miyato, Kataoka, Koyama, and
  Yoshida]{miyato2018spectral}
Takeru Miyato, Toshiki Kataoka, Masanori Koyama, and Yuichi Yoshida.
\newblock Spectral normalization for generative adversarial networks.
\newblock \emph{arXiv preprint arXiv:1802.05957}, 2018.

\bibitem[Ng et~al.(2001)Ng, Jordan, and Weiss]{ng2001spectral}
Andrew Ng, Michael Jordan, and Yair Weiss.
\newblock On spectral clustering: Analysis and an algorithm.
\newblock \emph{Advances in neural information processing systems}, 14, 2001.

\bibitem[Quellec et~al.(2017)Quellec, Cazuguel, Cochener, and
  Lamard]{quellec2017multiple}
Gwenol{\'e} Quellec, Guy Cazuguel, B{\'e}atrice Cochener, and Mathieu Lamard.
\newblock Multiple-instance learning for medical image and video analysis.
\newblock \emph{IEEE reviews in biomedical engineering}, 10:\penalty0 213--234,
  2017.

\bibitem[Ren et~al.(2023)Ren, Zhao, He, Wu, Mai, Xu, Huang, He, Huang, and
  Yao]{ren2023iib}
Qin Ren, Yu~Zhao, Bing He, Bingzhe Wu, Sijie Mai, Fan Xu, Yueshan Huang,
  Yonghong He, Junzhou Huang, and Jianhua Yao.
\newblock Iib-mil: Integrated instance-level and bag-level multiple instances
  learning with label disambiguation for pathological image analysis.
\newblock In \emph{International Conference on Medical Image Computing and
  Computer-Assisted Intervention}, pages 560--569. Springer, 2023.

\bibitem[Ripley(1981)]{ripley1981spatial}
Brian~D Ripley.
\newblock Spatial statistics.
\newblock \emph{Wiley Series in Probability and Statistics}, 1981.

\bibitem[Seeger(2000)]{seeger2000learning}
Matthias Seeger.
\newblock Learning with labeled and unlabeled data.
\newblock 2000.

\bibitem[Shao et~al.(2021)Shao, Bian, Chen, Wang, Zhang, Ji,
  et~al.]{shao2021transmil}
Zhuchen Shao, Hao Bian, Yang Chen, Yifeng Wang, Jian Zhang, Xiangyang Ji,
  et~al.
\newblock Transmil: Transformer based correlated multiple instance learning for
  whole slide image classification.
\newblock \emph{Advances in neural information processing systems},
  34:\penalty0 2136--2147, 2021.

\bibitem[Silva-Rodriguez et~al.(2021)Silva-Rodriguez, Colomer, Dolz, and
  Naranjo]{silva2021self}
Julio Silva-Rodriguez, Adri{\'a}n Colomer, Jose Dolz, and Valery Naranjo.
\newblock Self-learning for weakly supervised gleason grading of local
  patterns.
\newblock \emph{IEEE journal of biomedical and health informatics}, 25\penalty0
  (8):\penalty0 3094--3104, 2021.

\bibitem[Song et~al.(2023)Song, Jaume, Williamson, Lu, Vaidya, Miller, and
  Mahmood]{song2023artificial}
Andrew~H Song, Guillaume Jaume, Drew~FK Williamson, Ming~Y Lu, Anurag Vaidya,
  Tiffany~R Miller, and Faisal Mahmood.
\newblock Artificial intelligence for digital and computational pathology.
\newblock \emph{Nature Reviews Bioengineering}, 1\penalty0 (12):\penalty0
  930--949, 2023.

\bibitem[Song et~al.(2024)Song, Williams, Williamson, Chow, Jaume, Gao, Zhang,
  Chen, Baras, Serafin, Colling, Downes, Farré, Humphrey, Verrill, True,
  Parwani, Liu, and Mahmood]{song_analysis_2024}
Andrew~H. Song, Mane Williams, Drew F.~K. Williamson, Sarah S.~L. Chow,
  Guillaume Jaume, Gan Gao, Andrew Zhang, Bowen Chen, Alexander~S. Baras,
  Robert Serafin, Richard Colling, Michelle~R. Downes, Xavier Farré, Peter
  Humphrey, Clare Verrill, Lawrence~D. True, Anil~V. Parwani, Jonathan T.~C.
  Liu, and Faisal Mahmood.
\newblock Analysis of {3D} pathology samples using weakly supervised {AI}.
\newblock \emph{Cell}, 187\penalty0 (10):\penalty0 2502--2520.e17, May 2024.
\newblock ISSN 1097-4172.
\newblock \doi{10.1016/j.cell.2024.03.035}.

\bibitem[Wu et~al.(2021)Wu, Schmidt, Hern{\'a}ndez-S{\'a}nchez, Molina, and
  Katsaggelos]{wu2021combining}
Yunan Wu, Arne Schmidt, Enrique Hern{\'a}ndez-S{\'a}nchez, Rafael Molina, and
  Aggelos~K Katsaggelos.
\newblock Combining attention-based multiple instance learning and gaussian
  processes for ct hemorrhage detection.
\newblock In \emph{International Conference on Medical Image Computing and
  Computer-Assisted Intervention}, pages 582--591. Springer, 2021.

\bibitem[Xiong et~al.(2023)Xiong, Chen, Sung, and King]{xiong2023diagnose}
Conghao Xiong, Hao Chen, Joseph~JY Sung, and Irwin King.
\newblock Diagnose like a pathologist: Transformer-enabled hierarchical
  attention-guided multiple instance learning for whole slide image
  classification.
\newblock \emph{arXiv preprint arXiv:2301.08125}, 2023.

\bibitem[Zbontar et~al.(2021)Zbontar, Jing, Misra, LeCun, and
  Deny]{zbontar2021barlow}
Jure Zbontar, Li~Jing, Ishan Misra, Yann LeCun, and St{\'e}phane Deny.
\newblock Barlow twins: Self-supervised learning via redundancy reduction.
\newblock In \emph{International conference on machine learning}, pages
  12310--12320. PMLR, 2021.

\bibitem[Zhang et~al.(2022)Zhang, Meng, Zhao, Qiao, Yang, Coupland, and
  Zheng]{zhang2022dtfd}
Hongrun Zhang, Yanda Meng, Yitian Zhao, Yihong Qiao, Xiaoyun Yang, Sarah~E
  Coupland, and Yalin Zheng.
\newblock Dtfd-mil: Double-tier feature distillation multiple instance learning
  for histopathology whole slide image classification.
\newblock In \emph{Proceedings of the IEEE/CVF Conference on Computer Vision
  and Pattern Recognition}, pages 18802--18812, 2022.

\bibitem[Zhao et~al.(2022)Zhao, Lin, Sun, Zhang, Huang, Wang, and
  Yao]{zhao2022setmil}
Yu~Zhao, Zhenyu Lin, Kai Sun, Yidan Zhang, Junzhou Huang, Liansheng Wang, and
  Jianhua Yao.
\newblock Setmil: spatial encoding transformer-based multiple instance learning
  for pathological image analysis.
\newblock In \emph{International Conference on Medical Image Computing and
  Computer-Assisted Intervention}, pages 66--76. Springer, 2022.

\bibitem[Zheng et~al.(2022)Zheng, Gindra, Green, Burks, Betke, Beane, and
  Kolachalama]{zheng2022graph}
Yi~Zheng, Rushin~H Gindra, Emily~J Green, Eric~J Burks, Margrit Betke,
  Jennifer~E Beane, and Vijaya~B Kolachalama.
\newblock A graph-transformer for whole slide image classification.
\newblock \emph{IEEE transactions on medical imaging}, 41\penalty0
  (11):\penalty0 3003--3015, 2022.

\bibitem[Zhou and Sch{\"o}lkopf(2005)]{zhou2005regularization}
Dengyong Zhou and Bernhard Sch{\"o}lkopf.
\newblock Regularization on discrete spaces.
\newblock In \emph{Joint Pattern Recognition Symposium}, pages 361--368.
  Springer, 2005.

\bibitem[Zhou et~al.(2003)Zhou, Bousquet, Lal, Weston, and
  Sch{\"o}lkopf]{zhou2003learning}
Dengyong Zhou, Olivier Bousquet, Thomas Lal, Jason Weston, and Bernhard
  Sch{\"o}lkopf.
\newblock Learning with local and global consistency.
\newblock \emph{Advances in neural information processing systems}, 16, 2003.

\bibitem[Zhou et~al.(2021)Zhou, Greenspan, Davatzikos, Duncan, Van~Ginneken,
  Madabhushi, Prince, Rueckert, and Summers]{zhou2021review}
S~Kevin Zhou, Hayit Greenspan, Christos Davatzikos, James~S Duncan, Bram
  Van~Ginneken, Anant Madabhushi, Jerry~L Prince, Daniel Rueckert, and Ronald~M
  Summers.
\newblock A review of deep learning in medical imaging: Imaging traits,
  technology trends, case studies with progress highlights, and future
  promises.
\newblock \emph{Proceedings of the IEEE}, 109\penalty0 (5):\penalty0 820--838,
  2021.

\end{thebibliography}

\newpage
\appendix

\section{Proofs}
\label{appendix:section:proofs}

\subsection{Proof of \autoref{eq:ineq_chain_2}}

We present a general result for an arbitrary multilayer perceptron with Lipschitz activation functions. 
Note that assuming Lipschitzness is not a restriction, since most of the currently used activation functions meet this property \citep{cai2020note}.
The desired \autoref{eq:ineq_chain_2} is a particular case of this result.
Let $L \in \Nbb$. 
Consider a $L$-layer perceptron that, given $\bY \in \Rbb^{N \times D_0}$, outputs $\bY^L \in \Rbb^{N \times D_L}$ defined by the following rule
\begin{gather}
	\bY^0 = \bY, \\
	\bY^{\ell+1} = \varphi_{\ell}\left( \bY^\ell \bW_\ell + \bB_\ell \right), \quad \ell \in \left\{0, \ldots, L-1 \right\},
\end{gather}
where $\bW_\ell \in \Rbb^{D_{\ell} \times D_{\ell+1}}$, and $\bB_\ell = \left[ \bb_\ell, \ldots, \bb_\ell \right]^\top \in \Rbb^{N \times D_{\ell+1}}$ where $\bb_\ell \in \Rbb^{D_{\ell+1}}$ are trainable weights, and $\varphi_\ell \colon \Rbb \to \Rbb$ are activation functions applied element-wise. 
We suppose that each activation function $\varphi_\ell$ is $K_{\ell}$-Lipschitz.  
Then, we obtain the following inequality,
\begin{equation}\label{eq:ineq_layer}
    \direnergy{\bY^{\ell+1}} \leq K_{\ell}^2 \left\| \bW_{\ell} \right\|_{2}^{2}  \direnergy{\bY^{\ell}}.
\end{equation}

Before verifying it, we note that by applying this inequality to every layer, we arrive at
\begin{equation}\label{eq:ineq_chain_L}
    \hspace{-1pt}
	\direnergy{\bY^L} \leq \cdots \leq K_{L - 1 - \ell : 0}^2  \left\| \bW_{L - 1 - \ell : 0} \right\|^2 \direnergy{\bY^{\ell}} \leq \cdots \leq K_{L - 1 : 0}^2 \left\| \bW_{L-1 : 0} \right\|^2 \direnergy{\bY},
\end{equation}
where $\left\| \bW_{\ell:0} \right\|^2 = \prod_{j=0}^{\ell} \left\| \bW_{j} \right\|^2$ and  $K_{\ell:0} = \prod_{j=0}^{\ell} K_\ell$. 
Taking $L = 2$, $D_0=D$, $D_1=L$, $D_2 = 1$, $\bb_0 = \bb_1 = \bzero$, $\varphi_0 = \tanh$, and $\varphi_1 = \operatorname{Id}$, we recover \autoref{eq:ineq_chain_2}. 

To verify \autoref{eq:ineq_layer}, we write $\bY^{\ell+1} = \left[ \by^{\ell+1}_1, \ldots, \by^{\ell+1}_N \right]^\top$ and $\bY^{\ell} = \left[ \by^{\ell}_1, \ldots, \by^{\ell}_N \right]^\top$. We have, 
\begin{align}
    \direnergy{\bY^{\ell+1}} & = \frac{1}{2} \sum_{i=1}^N \sum_{j=1}^N A_{ij} \left\| \by^{\ell+1}_i - \by^{\ell+1}_j \right\|_2^2 = \\
    & = \frac{1}{2} \sum_{i=1}^N \sum_{j=1}^N A_{ij} \left\| \varphi_{\ell}\left( \bW_{\ell}^\top \by^{\ell}_i + \bB_\ell \right) - \varphi_{\ell}\left( \bW_{\ell}^\top \by^{\ell}_j + \bB_\ell \right) \right\|_2^2 \leq \label{eq:ineq_lipschitz_def_1}\\
    & \leq K_{\ell}^2 \frac{1}{2} \sum_{i=1}^N \sum_{j=1}^N A_{ij} \left\| \bW_{\ell}^\top \left( \by_i^\ell - \by_j^\ell \right) \right\|_2^2 \leq \label{eq:ineq_lipschitz_def_2}\\
    & \leq K_{\ell}^2 \left\| \bW_{\ell} \right\|_2^2 \frac{1}{2} \sum_{i=1}^N \sum_{j=1}^N A_{ij} \left\| \by_i^\ell - \by_j^\ell \right\|_2^2 = K_{\ell}^2 \left\| \bW_{\ell} \right\|_2^2 \direnergy{\bY^\ell}, \label{eq:norm_consistency_2}
\end{align}
where from \autoref{eq:ineq_lipschitz_def_1} to \autoref{eq:ineq_lipschitz_def_2} we have used the definition of Lipschitz function and from \autoref{eq:ineq_lipschitz_def_2} to \autoref{eq:norm_consistency_2} we have used the consistency between the spectral and Euclidean norms.

\subsection{Proof of \autoref{eq:smooth_ineq}}

In this section, we adapt the proof presented in \citep{cai2020note} for a similar result.
Let $\bU \in \Rbb^{N \times D}$. Our goal is to show that 
\begin{equation}
    \direnergy{ \left( \bI + \gamma \bL \right)^{-1} \bU } \leq \lambda_{\gamma}^* \direnergy{\bU},
\end{equation}
where $\gamma>0$ and $\lambda_{\gamma}^* = \max \left\{ \left( 1 + \gamma \lambda_n \right)^{-2} \colon \lambda_n \in \Lambda \setminus \left\{ 0 \right\}\right\}$, being $\Lambda = \left\{ \lambda_1, \ldots, \lambda_N \right\}$ the eigenvalues of the symmetric graph Laplacian matrix $\bL$.
First, we reduce the proof to univariate graph functions by looking at the rows of $\bU$ as univariate graph functions. 
% We have that $\bg_d = \left( \bI + \gamma \bL \right)^{-1} \bff_d$, where $\bff_d \in \Rbb^N$ are the rows of $\bF$, this is, $\bF = \left[ \bff_1, \ldots, \bff_D \right]$.
Denoting them as $\left\{ \bu_1, \ldots, \bu_D \right\}$, where each $\bu_d \in \Rbb^N$, we have $\direnergy{\left( \bI + \gamma \bL \right)^{-1} \bU} = \sum_{d=1}^D \direnergy{\left( \bI + \gamma \bL \right)^{-1} \bu_d}$. Therefore, it will be sufficient to show that, for any $\bu \in \Rbb^N$, 
\begin{equation}
    \direnergy{\left( \bI + \gamma \bL \right)^{-1} \bu} \leq \lambda_{\gamma}^* \direnergy{\bu}.
\end{equation}

Next, it is useful to note that if $\lambda_n$ is an eigenvalue of $\bL$ with associated eigenvector $\bv_n$, then $\left( 1 + \gamma \lambda_n \right)^{-1}$ is an eigenvalue of $\left( \bI + \gamma \bL \right)^{-1}$ with the same associated eigenvector.
Finally, let $\left\{ \bv_1, \ldots, \bv_N \right\}$ be a orthonormal eigenbasis of $\bL$, being each $\bv_n$ associated to the eigenvalue $\lambda_n$. 
This basis always exists since $\bL$ is a symmetric matrix.
Writing $\bu = \sum_{n=1}^N c_n \bv_n$, with $c_n \in \Rbb$, we have
\begin{equation}
    \left( \bI + \gamma \bL \right)^{-1} \bu = \sum_{n=1}^N c_n \left( 1 + \gamma \lambda_n \right)^{-1} \bv_n.
\end{equation}
Using that the eigenvectors are orthogonal to each other, we arrive at
\begin{equation}
    \direnergy{\left( \bI + \gamma \bL \right)^{-1} \bu} = \sum_{n=1}^N c_n^2 \lambda_n \left( 1 + \gamma \lambda_n \right)^{-2} \leq \lambda_{\gamma}^* \sum_{n=1}^N c_n^2 \lambda_n = \lambda_{\gamma}^* \direnergy{\bu}.
\end{equation}

% To this end, 
% It is easy to check that the eigenvalues of $\left( \bI + \gamma \bL \right)^{-1}$ are $\left\{ (1+\gamma\lambda_1)^{-1}, \ldots, \lambda_N \right\}$

% it is also a orthonormal base of eigenvectors of $\left( \bI + \gamma \bL \right)^{-1}$, each one associated to the eigenvalue 

% Suppose that . 

% We write $\bF = \left[ \bff_1, \ldots, \bff_D \right]$, where now we are using $\bff_d \in \Rbb^N$ to refer to the columns of $\bF$.
% Note that in \autoref{section:method} we used a similar notation to refer to the rows. 
% Writing $\bG = \left[ \bg_1, \ldots, \bg_D \right]$, we have that $\bg_d = \left( \bI + \gamma \bL \right)^{-1} \bff_d$. Since 

% We write 
% We observe that, if $\bF = \left[ \bff_1, \ldots, \bff_N \right]^\top \in \Rbb^{N \times D}$, then

% Let $ \Lambda = \left\{ \lambda_1, \ldots, \lambda_N \right\} $ be the eigenvalues of the graph Laplacian $\bL$, and  

\section{Experiments: details and further results}
\label{appendix:section:exp_details}

In this section, we provide the details of the datasets, architectures, and configurations used for each experiment. 
The code is available at \url{https://github.com/Franblueee/SmMIL}. 

\subsection{Datasets}
\label{appendix:subsection:datasets}

We provide insights into the datasets we have used: a description of the problem, the train/test splits, and preprocessing (instance selection and feature extraction). 
For all datasets, we obtain an initial train/test partition. Then, we split the initial train partition into five different train/validation splits. 
Every model is trained on each of these splits and then evaluated on the test set.
We report the average performance on this test set. 

\textbf{RSNA.} It was published by the Radiological Society of North America (RSNA) to detect acute intracranial hemorrhage and its subtypes \citep{flanders2020construction}.
It is available in Kaggle\footnote{\url{https://www.kaggle.com/c/rsna-intracranial-hemorrhage-detection}}.
We use the official train-test split. 
It includes a total of 1150 scans. 
There are a total amount of 39750 slices and the number in each scan varies from 24 to 57. 
Each slice is preprocessed following \citep{wu2021combining}. 
% Then, a 512 feature vector ($P=512$) is obtained using the pre-trained ResNet18 model available in \texttt{torchvision}\footnote{https://pytorch.org/vision/main/models/generated/torchvision.models.resnet18.html}.

\textbf{PANDA.} It is a public dataset for the classification of the severity of prostate cancer from microscopy scans of prostate biopsy samples \citep{bulten2022artificial}. 
It is available in Kaggle\footnote{\url{https://www.kaggle.com/c/prostate-cancer-grade-assessment/data}}.
Since the official test set is not publicly available, we use the train/test split proposed in \citep{silva2021self}. 
To extract the patches from each WSI, we follow the procedure described in \citep{silva2021self}, obtaining patches of size $512 \times 512$ at $10\times$ magnification.
This results in a total amount of 10503 WSIs and 1107931 patches.
% We use the pre-trained ResNet18 model from \texttt{torchvision} to extract a 512 feature vector ($P=512$) from each patch.

\textbf{CAMELYON16.} It is a public dataset for the detection of breast cancer metastasis \citep{bejnordi2017diagnostic}. 
It is available at the Registry of Open Data of AWS\footnote{\url{https://registry.opendata.aws/camelyon/}}.
The official repository contains 400 WSIs in total, including 270 for training and 130 for testing. 
From each WSI, we extract patches of size $512 \times 512$ at $20\times$ magnification using the method proposed by \citet{lu2021data}. 
% From each patch, a 2048 feature vector ($P=2048$) is obtained using a ResNet50 model with the pre-trained \texttt{bt\_rn50\_ep200.torch} weights provided by \citet{kang2023benchmarking} in their official repository\footnote{https://github.com/lunit-io/benchmark-ssl-pathology}. This feature extractor is pre-trained in a huge dataset of WSIs patches using the Barlow Twins Self-Supervised Learning method \citep{zbontar2021barlow}. 

\subsection{Model and training configuration}
\label{appendix:subsection:configuration}

We provide details about how we have implemented the proposed methods and how we have conducted the experiments. 

\textcolor{review}{
    \textbf{Feature extractor.} 
    Due to the limited memory of the GPU, it is necessary to extract features from each instance. Otherwise, the bags will not fit in memory.  In this work, we consider three options for the feature extractor, all of which are pre-trained in Imagenet: ResNet18 ($P=512$), ResNet50 ($P=2048$), and ViT-B-32 ($P=768$). In addition, for CAMELYON16 we also consider ResNet50-BT\footnote{Weights available at \url{https://github.com/lunit-io/benchmark-ssl-pathology}.} ($P=2048$), which is a ResNet50 model pre-trained using the Barlow Twins Self-Supervised Learning method on a huge dataset of WSIs patches \citep{zbontar2021barlow, kang2023benchmarking}. The results reported in the main text correspond to ResNet18 for RSNA and PANDA, and to ResNet50-BT for CAMELYON16. We study how the choice of the feature extractor affects the results in \autoref{appendix:subsection:ablation}. 
}

\textbf{Model architecture.} We describe the architecture we have used for the proposed methods (\smoothattpool\ and \smoothtransformerattpool). 
For the rest of the methods considered, we adopt their original implementations and default configurations, publicly available on their GitHub repositories. 
For the independent instance encoding part (see \autoref{fig:abmil} and \autoref{fig:smoothattpool}), the instance embeddings $\bh_n$ are obtained using one fully connected layer with 512 units ($D=512$). 
For the attention-based pooling described by \autoref{eq:attpool-f} and \autoref{eq:attpool-z}, we fix $D=512$ and $L=100$. 
The transformer encoder in \autoref{fig:smoothattpool_smoothtransformer} is implemented using two transformer layers. These layers use the standard multi-head attention mechanism equipped with skip connections and layer normalization \citep{bishop2024deep}. We fix the key, query, and value dimensions to $128$ and the number of heads to 8. 
We used the Pytorch's implementation of dot product attention\footnote{\url{https://pytorch.org/docs/stable/generated/torch.nn.functional.scaled\_dot\_product\_attention.html}}.
% After the last transformer layer, the resulting bag of embeddings is projected back to a $512$-dimensional feature space using a fully connected layer. 
Finally, the bag-embedding classifier was implemented using one fully connected layer.

\textbf{Training setup and hyperparameters.} To ensure fair and reproducible conditions, we trained every method under the same setup.
The number of epochs was set to $50$. 
We adopt the Adam optimizer \citep{kingma2014adam} with the default Pytorch configuration. 
For the base learning rate, we considered two different values, $10^{-4}$ and $10^{-5}$, since we noticed that models that do not use transformers obtained better results when the learning rate was higher. 
We report the best results for each model.
We adopted a slow start using Pytorch's \texttt{LinearLR} scheduler with \texttt{start\_factor=0.1} and \texttt{total\_iters=5}. 
During training, we monitored the bag AUROC and cross-entropy loss in the validation set and kept the weights that obtained the best results. 
The batch size was set to 32 in RSNA and PANDA.
In CAMELYON16, it was set to 4 for no-transformer methods, and to 1 for transformer-based methods. 
However, for SETMIL, we had to set it to 1 in PANDA and CAMELYON16 due to its high GPU memory requirements. 
In RSNA we weighted the loss function to account for the imbalance between positive and negative bags since we observed it to improve the results. 
All the experiments were performed on one NVIDIA GeForce RTX 3090.

\subsection{Further ablation studies.}
\label{appendix:subsection:ablation}

We complete the ablation study presented in the main paper, \autoref{subsection:experiments-ablation}, by looking at the rest of the design choices or hyperparameters associated with our \smoothopp. 
% The idea is to isolate the contribution of these elements, showing that \smoothopp\ leads to enhanced results under any choice of hyperparameters. 
% This confirms that our hypothesis --- that neighboring instances are likely to have the same label --- is a powerful inductive bias worth exploring.

\begin{figure}
    \centering
    \begin{subfigure}{0.49\textwidth}
        \centering
        \includegraphics[trim={0cm 0cm 0cm 0cm},clip,width=1.0\textwidth]{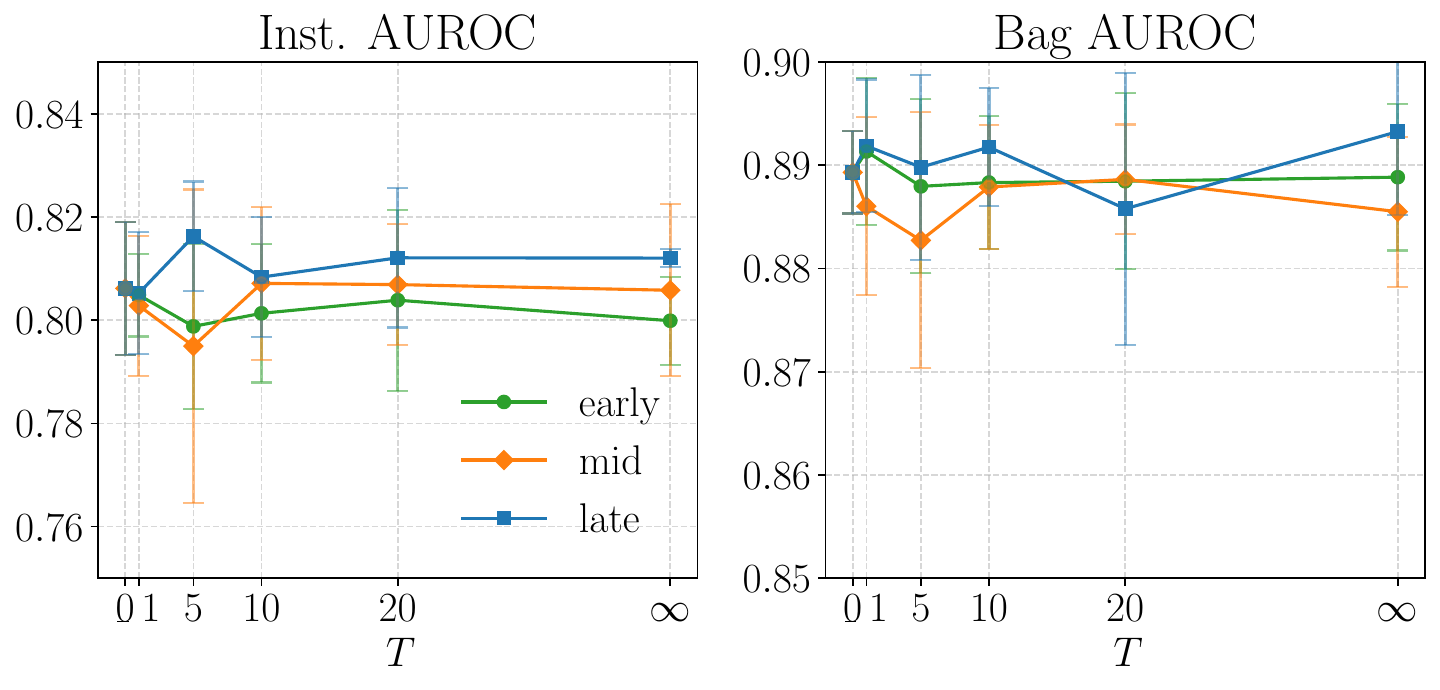}
        \caption{RSNA.}
        \label{fig:ablation_T-rsna}
    \end{subfigure}
    \hfill
    \begin{subfigure}{0.49\textwidth}
        \centering
        \includegraphics[trim={0cm 0cm 0cm 0cm},clip,width=1.0\textwidth]{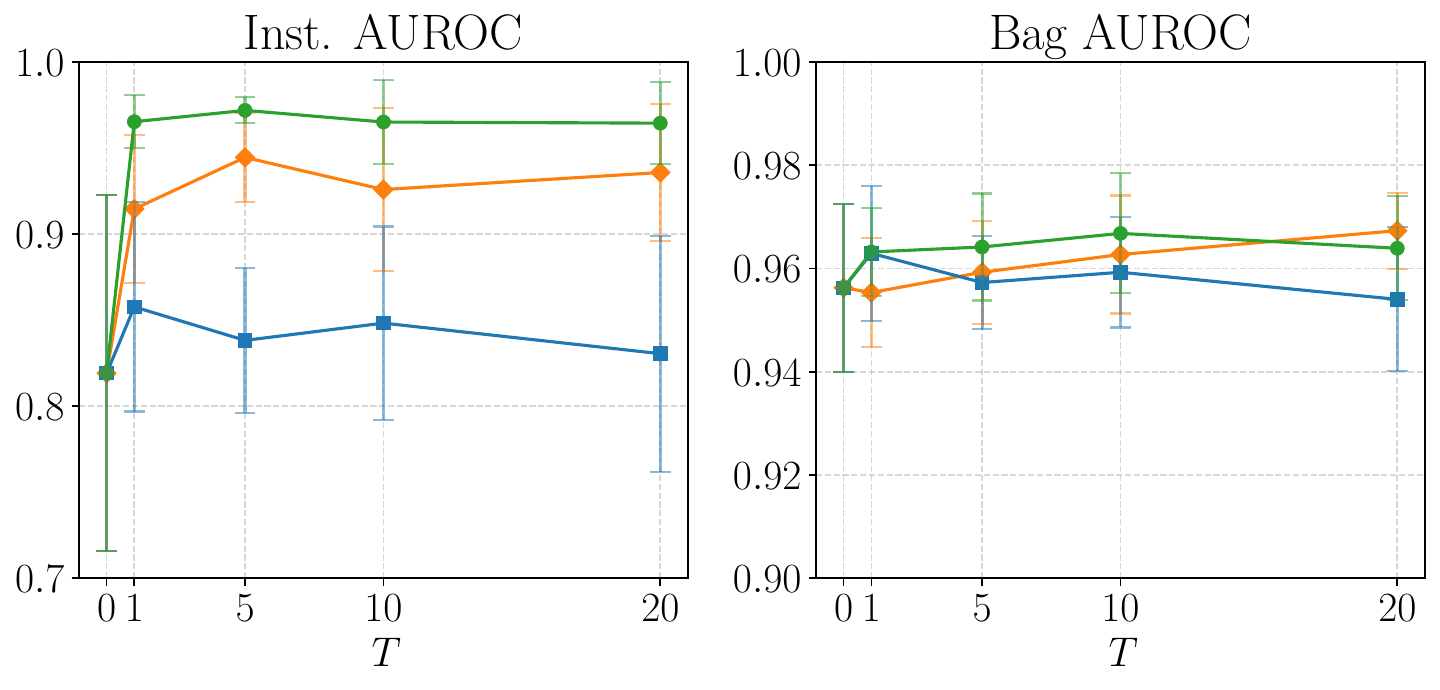}
        % \caption{}
        \caption{CAMELYON16.}
        \label{fig:ablation_T-camelyon}
    \end{subfigure}
    \caption{
        Influence of the number of steps $T$ used to approximate \smoothopp\ in RSNA and CAMELYON16. 
        \abmil\ corresponds to $T=0$. 
        Using $T=10$ is enough to closely match the performance of the exact form ($T=\infty$).  
    }
    \label{fig:ablation_T}
    \vspace{-15pt}
\end{figure}

\textbf{Smooth operator approximation.}
The exact form of \smoothopp\, given by \autoref{eq:smoothop_exact} becomes computationally infeasible for large bag sizes. 
The quality of the approximation, given by \autoref{eq:smoothop_approx}, is controlled by the number of steps $T$.
\autoref{fig:ablation_T} shows the results for different values of this hyperparameter in RSNA and CAMELYON16. 
In RSNA, since the bags are smaller, we can compute the closed-form solution, which we represent as $T=\infty$. 
Almost any choice of $T>0$ improves upon \abmil. 
This improvement is particularly noticeable in CAMELYON16. 
Moreover, in most cases, the performance stabilizes at $T=10$, which is the value we used in our experiments. 

\textcolor{review}{
    \textbf{\smoothopp\ on top of other models.} 
    We have proposed two new models (\smoothattpool\ and \smoothtransformerattpool) by applying \smoothopp\ on top of two baselines (\abmil\ and Transformer+\abmil, respectively).
    Instead, the \smoothopp\ can be applied on top of other existing approaches. 
    In \autoref{tab:smoothop_others} we explore how other approaches behave when combined with the proposed \smoothopp\ in the CAMELYON16 dataset. 
    Instance-level performance is enhanced (greatly in some cases, e.g. an increase from 0.76 to 0.96 in AUROC for DSMIL), whereas bag-level results are competitive. 
    The decrease in bag-level results for DFTD-MIL is explained by the fact that this method randomly splits each bag into different chunks. This may lead to the loss of local interactions exploited by Sm (e.g. if two adjacent instances end in different chunks).
}

\begin{table}
    \vspace{-20pt}
	\centering
    \caption{
		Using \smoothopp\ on top of other models (CAMELYON16 with ResNet50-BT features). Improvements are highlighted in green. Using the proposed \smoothopp\ increases the instance-level performance, while the bag-level performance remains competitive.  
	}
    \label{tab:smoothop_others}
    \begin{adjustbox}{width=0.68\textwidth}
        \begin{tabular}{ccccc}
            \toprule
             & \multicolumn{2}{c}{Instance} & \multicolumn{2}{c}{Bag} \\ \midrule
             & AUROC $(\uparrow)$ & F1 $(\uparrow)$ & AUROC $(\uparrow)$ & F1 $(\uparrow)$ \\
             \midrule
            \clam & $0.849_{0.044}$ & $0.821_{0.046}$ & $0.96_{0.029}$ & $0.897_{0.012}$ \\
            \smoothopp\clam & ${\color{improve}{0.928}}_{0.028}$ & ${\color{improve}{0.873}}_{0.018}$ & ${\color{improve}{0.966}}_{0.007}$ & $0.889_{0.017}$ \\
            \midrule
            \dsmil & $0.76_{0.078}$ & $0.654_{0.203}$ & $0.947_{0.085}$ & $0.866_{0.136}$ \\
            \smoothopp\dsmil & ${\color{improve}{0.960}}_{0.013}$ & ${\color{improve}{0.776}}_{0.088}$ & ${\color{improve}{0.967}}_{0.011}$ & ${\color{improve}{0.919}}_{0.018}$ \\
            \midrule
            \dftdmil & $0.984_{0.002}$ & $0.742_{0.040}$ & $0.983_{0.010}$ & $0.937_{0.013}$ \\
            \smoothopp\dftdmil & $0.984_{0.183}$ & ${\color{improve}{0.836}}_{0.222}$ & $0.978_{0.158}$ & $0.903_{0.183}$ \\ \bottomrule
        \end{tabular}
    \end{adjustbox}
\end{table}

\textcolor{review}{
    \textbf{An alternative smoothing strategy.} 
    Introducing a penalty term in the loss function to favor smoothness is a natural alternative to the proposed operator. 
    However, there is an important difference: the use of a penalty term does not modify the model architecture. 
    The penalty term favors that the learned weights encode such a property, but it is not explicitly encoded in the model.
    For instance, note that the penalty term is not used at inference time. 
    We compare the penalty-based approach and the proposed \smoothopp\ in \autoref{tab:alternative-smooth}. 
    Although differences are not large, \smoothopp\ obtains superior performance.
}

\begin{table}
\caption{
    Instance and bag AUROC (higher is better) in CAMELYON16 using ResNet50-BT features for the proposed methods and the penalty-based approach. The best in each column is highlighted in bold. \smoothopp\ obtains superior performance, although the differences are not large. 
}
\label{tab:alternative-smooth}
\centering
\begin{adjustbox}{width=\textwidth}
\begin{tabular}{@{}cccccccc@{}}
\toprule
& & \multicolumn{2}{c}{\textbf{RSNA}} & \multicolumn{2}{c}{\textbf{PANDA}} & \multicolumn{2}{c}{\textbf{CAMELYON16}} \\ \midrule
& & Inst. & Bag & Inst. & Bag & Inst. & Bag \\ \midrule
\multirow{2}{*}{\makecell{W/o\\global int.}} & SmAP & $\mathbf{0.798}_{0.033}$ & $0.888_{0.005}$ & $\mathbf{0.799}_{0.005}$ & $\mathbf{0.943}_{0.001}$ & $0.961_{0.007}$ & $\mathbf{0.965}_{0.007}$ \\
& \abmil+PENALTY & $0.782_{0.050}$ & $\mathbf{0.889}_{0.043}$ & $0.780_{0.003}$ & $0.935_{0.001}$ & $\mathbf{0.979}_{0.013}$ & $0.963_{0.012}$ \\
\midrule
\multirow{2}{*}{\makecell{W/\\global int.}} & \smoothtransformerattpool & $\mathbf{0.767}_{0.046}$ & $\mathbf{0.906}_{0.007}$ & $\mathbf{0.790}_{0.007}$ & $0.946_{0.003}$ & $\mathbf{0.789}_{0.008}$ & $0.976_{0.014}$ \\
& \transformer+PENALTY & $0.737_{0.045}$ & $0.905_{0.005}$ & $0.772_{0.011}$ & $\mathbf{0.947}_{0.001}$ & $0.769_{0.099}$ & $\mathbf{0.988}_{0.004}$ \\
\bottomrule
\end{tabular}
\end{adjustbox}
\end{table}

\textcolor{review}{
    \textbf{Feature extractor.} 
    We investigate whether the choice of the feature extractor influences the results and conclusions presented in the main text. 
    We have evaluated each of the considered methods in each dataset using the feature extractors mentioned above (ResNet18, ResNet50, ViT-B-32, and ResNet50-BT).
    The results are shown in Tables \ref{tab:feat_ext_inst_auroc}--\ref{tab:feat_ext_bag_f1}.
    We summarize them in \autoref{tab:feat_ext_ranks}, where we collect the average instance and bag rank of each method for each feature extractor. 
    We observe that the proposed smooth operator \smoothopp\ obtains in almost all cases the highest rank. 
    This supports the idea that the improvement introduced by \smoothopp\ does not depend on the used features.
}

\begin{table}
\centering
\caption{
    Instance and bag average ranks (lower is better) obtained by each method for different choices of the feature extractor. The best result within each group is bolded, and the second-best is underlined. SmAP and SmTAP obtain in almost all cases the highest rank.
}
\label{tab:feat_ext_ranks}
\begin{adjustbox}{width=\textwidth}\begin{tabular}{@{}cc|cc|cc|cc|cc@{}}
\toprule
& & \multicolumn{2}{c|}{\textbf{ResNet18}} & \multicolumn{2}{c|}{\textbf{ResNet50}} & \multicolumn{2}{c|}{\textbf{ViT-B-32}} & \multicolumn{2}{c}{\textbf{ResNet50-BT}} \\
& & Inst. & Bag & Inst. & Bag & Inst. & Bag & Inst. & Bag \\
\midrule
\multirow{7}{*}{\makecell{Without\\global\\interactions}} & \smoothattpool & $\mathbf{2.000}_{0.632}$ & $\underline{2.000}_{1.095}$ & $\mathbf{1.625}_{0.744}$ & $\mathbf{1.750}_{0.707}$ & $\mathbf{1.667}_{0.816}$ & $\mathbf{1.500}_{1.225}$ & $\mathbf{1.000}_{0.000}$ & $\underline{2.000}_{0.000}$ \\
& \abmil & $2.667_{1.366}$ & $\mathbf{1.667}_{0.816}$ & $3.750_{1.581}$ & $3.250_{0.707}$ & $4.333_{2.160}$ & $3.000_{0.894}$ & $4.500_{0.707}$ & $3.500_{0.707}$ \\
& \deepgraphsurv & $4.000_{2.000}$ & $5.500_{1.225}$ & $\underline{2.500}_{1.195}$ & $5.625_{0.916}$ & $3.333_{1.211}$ & $5.000_{0.000}$ & $\underline{2.500}_{0.707}$ & $6.000_{0.000}$ \\
& \clam & $6.167_{0.983}$ & $5.500_{1.225}$ & $4.750_{2.053}$ & $4.500_{2.070}$ & $6.333_{0.816}$ & $5.000_{2.449}$ & $3.000_{1.414}$ & $3.500_{0.707}$ \\
& \dsmil & $5.167_{0.983}$ & $5.333_{1.506}$ & $6.375_{0.518}$ & $6.000_{0.926}$ & $5.667_{1.033}$ & $6.667_{0.516}$ & $6.000_{0.000}$ & $5.000_{0.000}$ \\
& \pathgcn & $5.833_{1.472}$ & $5.167_{1.722}$ & $5.625_{1.302}$ & $4.625_{2.134}$ & $4.500_{1.643}$ & $4.000_{1.673}$ & $7.000_{0.000}$ & $7.000_{0.000}$ \\
& \dftdmil & $\underline{2.167}_{0.983}$ & $2.833_{1.169}$ & $3.375_{1.302}$ & $\underline{2.250}_{1.488}$ & $\underline{2.167}_{0.983}$ & $\underline{2.833}_{0.983}$ & $4.000_{1.414}$ & $\mathbf{1.000}_{0.000}$ \\
\midrule
\multirow{6}{*}{\makecell{With\\global\\interactions}} & \smoothtransformerattpool & $\mathbf{2.167}_{1.835}$ & $\mathbf{1.667}_{1.033}$ & $\mathbf{2.375}_{1.847}$ & $\mathbf{1.875}_{0.835}$ & $\mathbf{1.833}_{0.983}$ & $\mathbf{2.500}_{0.837}$ & $\mathbf{1.500}_{0.707}$ & $\mathbf{1.500}_{0.707}$ \\
& \transmil & $3.167_{1.329}$ & $3.833_{1.169}$ & $3.500_{1.069}$ & $3.625_{1.061}$ & $4.167_{1.329}$ & $4.500_{1.975}$ & $4.000_{1.414}$ & $4.000_{0.000}$ \\
& \setmil & $3.667_{0.816}$ & $3.500_{1.975}$ & $3.625_{1.768}$ & $4.125_{2.031}$ & $3.667_{1.506}$ & $3.333_{1.862}$ & $4.500_{0.707}$ & $6.000_{0.000}$ \\
& \gtp & $4.167_{0.983}$ & $3.333_{1.751}$ & $4.500_{1.069}$ & $3.375_{1.598}$ & $5.000_{1.265}$ & $4.833_{1.329}$ & $6.000_{0.000}$ & $5.000_{0.000}$ \\
& \iibmil & $5.167_{2.041}$ & $5.667_{0.816}$ & $4.500_{2.138}$ & $5.125_{1.642}$ & $4.167_{1.722}$ & $3.167_{2.041}$ & $\underline{2.000}_{1.414}$ & $2.500_{0.707}$ \\
& \camil & $\underline{2.667}_{1.751}$ & $\underline{3.000}_{0.894}$ & $\underline{2.500}_{1.309}$ & $\underline{2.875}_{1.356}$ & $\underline{2.167}_{1.472}$ & $\underline{2.667}_{1.211}$ & $3.000_{1.414}$ & $\underline{2.000}_{1.414}$ \\
\bottomrule
\end{tabular}\end{adjustbox}
\end{table}

% \begin{table}[]
\begin{sidewaystable}
    \caption{Instance AUROC (higher is better) for different choices of the feature extractor.}
    \label{tab:feat_ext_inst_auroc}
	\begin{adjustbox}{width=1.0\textwidth}
		\begin{tabular}{@{}cc|ccc|ccc|ccc|c@{}}
			\toprule
			&               & \multicolumn{3}{c|}{ResNet18} & \multicolumn{3}{c|}{ResNet50} & \multicolumn{3}{c|}{ViT-B-32} & ResNet50+BT \\
			&               & RSNA  & PANDA  & CAMELYON16  & RSNA  & PANDA  & CAMELYON16  & RSNA  & PANDA  & CAMELYON16  & CAMELYON16  \\
			\midrule
			\multirow{7}{*}{\makecell{Without\\global\\interactions}} & \smoothattpool & $\underline{0.798}_{0.033}$ & $\mathbf{0.799}_{0.005}$ & $0.798_{0.037}$ & $0.783_{0.026}$ & $\underline{0.786}_{0.005}$ & $\underline{0.854}_{0.116}$ & $\underline{0.804}_{0.017}$ & $\mathbf{0.810}_{0.007}$ & $0.773_{0.062}$ & $\mathbf{0.961}_{0.007}$ \\
            & \abmil & $\mathbf{0.806}_{0.012}$ & $0.768_{0.002}$ & $0.679_{0.082}$ & $\mathbf{0.796}_{0.027}$ & $0.774_{0.009}$ & $0.806_{0.130}$ & $0.797_{0.023}$ & $0.773_{0.004}$ & $0.755_{0.143}$ & $0.816_{0.055}$ \\
            & \deepgraphsurv & $0.681_{0.054}$ & $0.720_{0.011}$ & $\underline{0.868}_{0.094}$ & $0.768_{0.013}$ & $\mathbf{0.806}_{0.002}$ & $0.814_{0.027}$ & $0.755_{0.063}$ & $\underline{0.809}_{0.008}$ & $0.756_{0.104}$ & $\underline{0.959}_{0.033}$ \\
            & \clam & $0.523_{0.069}$ & $0.727_{0.046}$ & $0.516_{0.102}$ & $0.497_{0.005}$ & $0.785_{0.004}$ & $0.559_{0.056}$ & $0.500_{0.000}$ & $0.777_{0.004}$ & $0.463_{0.034}$ & $0.849_{0.044}$ \\
            & \dsmil & $0.554_{0.004}$ & $0.765_{0.008}$ & $0.628_{0.181}$ & $0.568_{0.015}$ & $0.747_{0.006}$ & $0.670_{0.111}$ & $0.702_{0.029}$ & $0.779_{0.002}$ & $0.661_{0.115}$ & $0.760_{0.078}$ \\
            & \pathgcn & $0.711_{0.049}$ & $0.664_{0.019}$ & $0.618_{0.214}$ & $0.692_{0.047}$ & $0.772_{0.011}$ & $0.851_{0.219}$ & $0.749_{0.046}$ & $0.769_{0.032}$ & $\underline{0.813}_{0.073}$ & $0.443_{0.138}$ \\
            & \dftdmil & $0.747_{0.070}$ & $\underline{0.795}_{0.004}$ & $\mathbf{0.920}_{0.074}$ & $\underline{0.795}_{0.018}$ & $0.784_{0.007}$ & $\mathbf{0.952}_{0.013}$ & $\mathbf{0.807}_{0.030}$ & $0.785_{0.008}$ & $\mathbf{0.952}_{0.011}$ & $0.884_{0.002}$ \\
			\midrule
			\multirow{6}{*}{\makecell{With\\global\\interactions}} & \smoothtransformerattpool & $\mathbf{0.767}_{0.046}$ & $\mathbf{0.790}_{0.007}$ & $0.750_{0.134}$ & $\mathbf{0.802}_{0.016}$ & $0.756_{0.012}$ & $0.716_{0.105}$ & $\mathbf{0.795}_{0.027}$ & $\mathbf{0.822}_{0.010}$ & $0.819_{0.162}$ & $\underline{0.789}_{0.008}$ \\
            & \transmil & $0.732_{0.013}$ & $0.751_{0.011}$ & $\underline{0.820}_{0.038}$ & $0.707_{0.023}$ & $0.743_{0.021}$ & $\underline{0.844}_{0.023}$ & $0.749_{0.019}$ & $0.749_{0.040}$ & $0.779_{0.062}$ & $0.781_{0.024}$ \\
            & \setmil & $0.726_{0.025}$ & $0.774_{0.007}$ & $0.792_{0.032}$ & $0.678_{0.004}$ & $\mathbf{0.774}_{0.071}$ & $0.787_{0.005}$ & $0.755_{0.001}$ & $0.789_{0.089}$ & $0.569_{0.009}$ & $0.615_{0.231}$ \\
            & \gtp & $0.736_{0.017}$ & $0.768_{0.022}$ & $0.594_{0.228}$ & $0.736_{0.024}$ & $0.754_{0.019}$ & $0.528_{0.149}$ & $0.760_{0.013}$ & $0.720_{0.039}$ & $0.477_{0.095}$ & $0.442_{0.091}$ \\
            & \iibmil & $0.675_{0.017}$ & $0.740_{0.020}$ & $0.500_{0.000}$ & $0.672_{0.024}$ & $0.729_{0.031}$ & $0.500_{0.000}$ & $0.690_{0.017}$ & $0.740_{0.031}$ & $\underline{0.863}_{0.038}$ & $\mathbf{0.873}_{0.138}$ \\
            & \camil & $\underline{0.760}_{0.036}$ & $\underline{0.785}_{0.011}$ & $\mathbf{0.917}_{0.043}$ & $\underline{0.796}_{0.013}$ & $\underline{0.766}_{0.027}$ & $\mathbf{0.964}_{0.014}$ & $\underline{0.783}_{0.007}$ & $\underline{0.806}_{0.015}$ & $\mathbf{0.939}_{0.007}$ & $0.742_{0.028}$ \\
			 \bottomrule
		\end{tabular}%
	\end{adjustbox}
    \vspace{10pt}
    \caption{Instance F1 (higher is better) for different choices of the feature extractor.}
    \label{tab:feat_ext_inst_f1}
	\begin{adjustbox}{width=1.0\textwidth}
		\begin{tabular}{@{}cc|ccc|ccc|ccc|c@{}}
			\toprule
			&               & \multicolumn{3}{c|}{ResNet18} & \multicolumn{3}{c|}{ResNet50} & \multicolumn{3}{c|}{ViT-B-32} & ResNet50+BT \\ 
			&               & RSNA  & PANDA  & CAMELYON16  & RSNA  & PANDA  & CAMELYON16  & RSNA  & PANDA  & CAMELYON16  & CAMELYON16  \\
			\midrule
			\multirow{7}{*}{\makecell{Without\\global\\interactions}} & \smoothattpool & $\underline{0.477}_{0.014}$ & $\underline{0.635}_{0.006}$ & $\underline{0.591}_{0.059}$ & $\mathbf{0.473}_{0.015}$ & $\underline{0.630}_{0.009}$ & $\mathbf{0.675}_{0.077}$ & $\underline{0.494}_{0.019}$ & $\mathbf{0.645}_{0.009}$ & $\mathbf{0.580}_{0.053}$ & $\mathbf{0.839}_{0.053}$ \\
            & \abmil & $\mathbf{0.486}_{0.033}$ & $0.602_{0.004}$ & $0.428_{0.049}$ & $\underline{0.470}_{0.031}$ & $0.611_{0.007}$ & $0.654_{0.067}$ & $\mathbf{0.498}_{0.021}$ & $0.605_{0.006}$ & $0.419_{0.029}$ & $0.767_{0.039}$ \\
            & \deepgraphsurv & $0.293_{0.168}$ & $0.581_{0.026}$ & $\mathbf{0.595}_{0.129}$ & $0.464_{0.022}$ & $\mathbf{0.641}_{0.002}$ & $\underline{0.663}_{0.038}$ & $0.479_{0.043}$ & $\underline{0.642}_{0.006}$ & $0.465_{0.059}$ & $0.771_{0.070}$ \\
            & \clam & $0.076_{0.154}$ & $0.568_{0.038}$ & $0.406_{0.238}$ & $0.000_{0.000}$ & $0.621_{0.007}$ & $0.584_{0.087}$ & $0.000_{0.000}$ & $0.610_{0.005}$ & $0.373_{0.054}$ & $\underline{0.821}_{0.046}$ \\
            & \dsmil & $0.180_{0.000}$ & $0.598_{0.006}$ & $0.155_{0.180}$ & $0.271_{0.019}$ & $0.592_{0.005}$ & $0.290_{0.174}$ & $0.399_{0.031}$ & $0.610_{0.004}$ & $0.255_{0.134}$ & $0.654_{0.203}$ \\
            & \pathgcn & $0.447_{0.014}$ & $0.526_{0.019}$ & $0.150_{0.211}$ & $0.431_{0.020}$ & $0.608_{0.010}$ & $0.371_{0.211}$ & $0.481_{0.039}$ & $0.610_{0.023}$ & $0.414_{0.119}$ & $0.077_{0.114}$ \\
            & \dftdmil & $0.453_{0.194}$ & $\mathbf{0.637}_{0.006}$ & $0.563_{0.132}$ & $0.447_{0.026}$ & $0.617_{0.011}$ & $0.591_{0.098}$ & $0.489_{0.033}$ & $0.616_{0.013}$ & $\underline{0.552}_{0.055}$ & $0.742_{0.040}$ \\
			\midrule
			\multirow{6}{*}{\makecell{With\\global\\interactions}} & \smoothtransformerattpool & $\mathbf{0.474}_{0.023}$ & $0.622_{0.010}$ & $\mathbf{0.581}_{0.061}$ & $\mathbf{0.517}_{0.020}$ & $0.606_{0.015}$ & $\mathbf{0.630}_{0.070}$ & $0.475_{0.034}$ & $\underline{0.658}_{0.013}$ & $\mathbf{0.552}_{0.138}$ & $\mathbf{0.600}_{0.067}$ \\
            & \transmil & $\underline{0.471}_{0.014}$ & $\underline{0.636}_{0.008}$ & $0.174_{0.080}$ & $0.442_{0.024}$ & $0.622_{0.023}$ & $0.196_{0.115}$ & $\underline{0.480}_{0.046}$ & $0.630_{0.041}$ & $0.194_{0.090}$ & $0.127_{0.078}$ \\
            & \setmil & $0.438_{0.027}$ & $0.631_{0.010}$ & $0.237_{0.058}$ & $0.405_{0.021}$ & $\mathbf{0.821}_{0.022}$ & $0.036_{0.021}$ & $0.467_{0.008}$ & $\mathbf{0.822}_{0.012}$ & $0.159_{0.039}$ & $0.134_{0.267}$ \\
            & \gtp & $0.425_{0.018}$ & $0.636_{0.011}$ & $0.168_{0.132}$ & $0.431_{0.013}$ & $0.621_{0.014}$ & $0.150_{0.122}$ & $0.447_{0.021}$ & $0.641_{0.017}$ & $0.084_{0.048}$ & $0.037_{0.036}$ \\
            & \iibmil & $0.420_{0.016}$ & $\mathbf{0.645}_{0.007}$ & $0.000_{0.000}$ & $0.403_{0.014}$ & $\underline{0.641}_{0.019}$ & $0.000_{0.000}$ & $0.443_{0.010}$ & $0.655_{0.006}$ & $0.295_{0.015}$ & $0.352_{0.100}$ \\
            & \camil & $0.456_{0.013}$ & $0.621_{0.013}$ & $\underline{0.403}_{0.157}$ & $\underline{0.483}_{0.024}$ & $0.615_{0.014}$ & $\underline{0.563}_{0.153}$ & $\mathbf{0.504}_{0.025}$ & $0.641_{0.014}$ & $\underline{0.426}_{0.055}$ & $\underline{0.479}_{0.175}$ \\
			 \bottomrule
		\end{tabular}%
	\end{adjustbox}
% \end{table}
\end{sidewaystable}

% \begin{table}[]
\begin{sidewaystable}
    \caption{Bag AUROC (higher is better) for different choices of the feature extractor.}
    \label{tab:feat_ext_bag_auroc}
	\begin{adjustbox}{width=1.0\textwidth}
		\begin{tabular}{@{}cc|ccc|ccc|ccc|c@{}}
			\toprule
			&               & \multicolumn{3}{c|}{ResNet18} & \multicolumn{3}{c|}{ResNet50} & \multicolumn{3}{c|}{ViT-B-32} & ResNet50+BT \\ 
			&               & RSNA  & PANDA  & CAMELYON16  & RSNA  & PANDA  & CAMELYON16  & RSNA  & PANDA  & CAMELYON16  & CAMELYON16  \\
			\midrule
			\multirow{7}{*}{\makecell{Without\\global\\interactions}} & \smoothattpool & $0.888_{0.005}$ & $\mathbf{0.943}_{0.001}$ & $\underline{0.729}_{0.037}$ & $\mathbf{0.890}_{0.007}$ & $\underline{0.944}_{0.001}$ & $\mathbf{0.777}_{0.046}$ & $\mathbf{0.897}_{0.005}$ & $\mathbf{0.947}_{0.002}$ & $0.775_{0.023}$ & $\underline{0.976}_{0.007}$ \\
            & \abmil & $\underline{0.889}_{0.005}$ & $0.933_{0.002}$ & $\mathbf{0.731}_{0.030}$ & $0.886_{0.013}$ & $0.942_{0.003}$ & $0.752_{0.023}$ & $\underline{0.893}_{0.007}$ & $0.943_{0.002}$ & $0.792_{0.022}$ & $0.956_{0.011}$ \\
            & \deepgraphsurv & $0.848_{0.017}$ & $0.837_{0.020}$ & $0.673_{0.017}$ & $0.877_{0.003}$ & $0.925_{0.002}$ & $0.695_{0.007}$ & $0.870_{0.010}$ & $0.938_{0.002}$ & $0.747_{0.039}$ & $0.870_{0.070}$ \\
            & \clam & $0.674_{0.157}$ & $0.893_{0.026}$ & $0.683_{0.082}$ & $0.802_{0.054}$ & $0.930_{0.002}$ & $\underline{0.775}_{0.041}$ & $0.735_{0.047}$ & $0.927_{0.001}$ & $\mathbf{0.832}_{0.030}$ & $0.960_{0.029}$ \\
            & \dsmil & $0.689_{0.063}$ & $0.921_{0.008}$ & $0.672_{0.110}$ & $0.761_{0.026}$ & $0.926_{0.002}$ & $0.693_{0.036}$ & $0.792_{0.041}$ & $0.925_{0.004}$ & $0.628_{0.063}$ & $0.947_{0.085}$ \\
            & \pathgcn & $0.888_{0.007}$ & $0.848_{0.005}$ & $0.585_{0.180}$ & $\underline{0.890}_{0.017}$ & $0.943_{0.006}$ & $0.708_{0.064}$ & $0.880_{0.023}$ & $\underline{0.945}_{0.006}$ & $0.741_{0.121}$ & $0.575_{0.206}$ \\
            & \dftdmil & $\mathbf{0.890}_{0.045}$ & $\underline{0.940}_{0.001}$ & $0.706_{0.022}$ & $0.886_{0.009}$ & $\mathbf{0.945}_{0.002}$ & $0.720_{0.031}$ & $0.870_{0.020}$ & $0.945_{0.001}$ & $\underline{0.801}_{0.015}$ & $\mathbf{0.983}_{0.010}$ \\
			\midrule
			\multirow{6}{*}{\makecell{With\\global\\interactions}} & \smoothtransformerattpool & $\mathbf{0.906}_{0.007}$ & $0.946_{0.003}$ & $\mathbf{0.783}_{0.056}$ & $\underline{0.893}_{0.009}$ & $0.944_{0.002}$ & $\mathbf{0.805}_{0.057}$ & $0.896_{0.009}$ & $0.946_{0.004}$ & $0.754_{0.032}$ & $\underline{0.976}_{0.014}$ \\
            & \transmil & $0.883_{0.008}$ & $0.933_{0.010}$ & $\underline{0.771}_{0.050}$ & $0.885_{0.008}$ & $0.942_{0.002}$ & $\underline{0.791}_{0.027}$ & $\mathbf{0.900}_{0.013}$ & $0.939_{0.003}$ & $0.655_{0.086}$ & $0.973_{0.018}$ \\
            & \setmil & $0.869_{0.011}$ & $\mathbf{0.974}_{0.003}$ & $0.628_{0.039}$ & $0.870_{0.008}$ & $\mathbf{0.977}_{0.005}$ & $0.657_{0.030}$ & $0.895_{0.012}$ & $\mathbf{0.970}_{0.005}$ & $0.469_{0.105}$ & $0.715_{0.155}$ \\
            & \gtp & $\underline{0.901}_{0.008}$ & $\underline{0.949}_{0.004}$ & $0.577_{0.075}$ & $\mathbf{0.896}_{0.016}$ & $\underline{0.952}_{0.002}$ & $0.459_{0.056}$ & $0.890_{0.015}$ & $0.945_{0.003}$ & $0.456_{0.110}$ & $0.748_{0.118}$ \\
            & \iibmil & $0.868_{0.013}$ & $0.931_{0.004}$ & $0.641_{0.012}$ & $0.861_{0.006}$ & $0.939_{0.004}$ & $0.455_{0.042}$ & $\underline{0.897}_{0.006}$ & $0.939_{0.002}$ & $\mathbf{0.791}_{0.049}$ & $0.974_{0.002}$ \\
            & \camil & $0.889_{0.019}$ & $0.938_{0.003}$ & $0.746_{0.041}$ & $0.892_{0.010}$ & $0.941_{0.002}$ & $0.738_{0.039}$ & $0.892_{0.008}$ & $\underline{0.947}_{0.004}$ & $\underline{0.772}_{0.034}$ & $\mathbf{0.984}_{0.007}$ \\
			 \bottomrule
		\end{tabular}%
	\end{adjustbox}
    \vspace{10pt}
    \caption{Bag F1 (higher is better) for different choices of the feature extractor.}
    \label{tab:feat_ext_bag_f1}
	\begin{adjustbox}{width=1.0\textwidth}
		\begin{tabular}{@{}cc|ccc|ccc|ccc|c@{}}
			\toprule
			&               & \multicolumn{3}{c|}{ResNet18} & \multicolumn{3}{c|}{ResNet50} & \multicolumn{3}{c|}{ViT-B-32} & ResNet50+BT \\ 
			&               & RSNA  & PANDA  & CAMELYON16  & RSNA  & PANDA  & CAMELYON16  & RSNA  & PANDA  & CAMELYON16  & CAMELYON16  \\
			\midrule
			\multirow{7}{*}{\makecell{Without\\global\\interactions}} & \smoothattpool & $\underline{0.787}_{0.026}$ & $\mathbf{0.915}_{0.002}$ & $\underline{0.661}_{0.056}$ & $0.788_{0.031}$ & $\mathbf{0.918}_{0.005}$ & $\mathbf{0.713}_{0.044}$ & $\mathbf{0.805}_{0.012}$ & $\mathbf{0.918}_{0.002}$ & $\mathbf{0.701}_{0.023}$ & $\underline{0.916}_{0.016}$ \\
            & \abmil & $\mathbf{0.796}_{0.011}$ & $\underline{0.909}_{0.001}$ & $\mathbf{0.667}_{0.022}$ & $\underline{0.800}_{0.024}$ & $0.912_{0.007}$ & $0.661_{0.019}$ & $0.788_{0.023}$ & $0.912_{0.001}$ & $\underline{0.673}_{0.038}$ & $0.912_{0.027}$ \\
            & \deepgraphsurv & $0.719_{0.036}$ & $0.823_{0.024}$ & $0.560_{0.021}$ & $0.770_{0.013}$ & $0.905_{0.003}$ & $0.590_{0.014}$ & $0.776_{0.019}$ & $0.908_{0.004}$ & $0.606_{0.067}$ & $0.772_{0.056}$ \\
            & \clam & $0.161_{0.291}$ & $0.868_{0.034}$ & $0.485_{0.272}$ & $0.016_{0.024}$ & $0.904_{0.005}$ & $\underline{0.676}_{0.041}$ & $0.000_{0.000}$ & $0.904_{0.002}$ & $0.671_{0.033}$ & $0.897_{0.012}$ \\
            & \dsmil & $0.240_{0.012}$ & $0.904_{0.008}$ & $0.252_{0.239}$ & $0.374_{0.064}$ & $0.907_{0.002}$ & $0.212_{0.118}$ & $0.683_{0.036}$ & $0.902_{0.004}$ & $0.308_{0.080}$ & $0.866_{0.136}$ \\
            & \pathgcn & $0.782_{0.064}$ & $0.857_{0.003}$ & $0.287_{0.324}$ & $0.757_{0.089}$ & $0.915_{0.004}$ & $0.507_{0.177}$ & $0.776_{0.012}$ & $\underline{0.914}_{0.006}$ & $0.606_{0.082}$ & $0.345_{0.352}$ \\
            & \dftdmil & $0.775_{0.282}$ & $0.903_{0.002}$ & $0.576_{0.105}$ & $\mathbf{0.806}_{0.009}$ & $\underline{0.917}_{0.002}$ & $0.599_{0.043}$ & $\underline{0.798}_{0.024}$ & $0.914_{0.002}$ & $0.668_{0.017}$ & $\mathbf{0.937}_{0.013}$ \\
			\midrule
			\multirow{6}{*}{\makecell{With\\global\\interactions}} & \smoothtransformerattpool & $\mathbf{0.825}_{0.026}$ & $0.917_{0.002}$ & $\mathbf{0.677}_{0.062}$ & $\underline{0.809}_{0.016}$ & $0.914_{0.003}$ & $\mathbf{0.707}_{0.020}$ & $\mathbf{0.807}_{0.032}$ & $0.914_{0.004}$ & $\underline{0.679}_{0.044}$ & $\mathbf{0.948}_{0.020}$ \\
            & \transmil & $0.716_{0.031}$ & $0.895_{0.029}$ & $0.636_{0.019}$ & $0.758_{0.045}$ & $0.905_{0.013}$ & $\underline{0.635}_{0.075}$ & $0.719_{0.027}$ & $0.892_{0.024}$ & $0.453_{0.098}$ & $0.911_{0.028}$ \\
            & \setmil & $0.716_{0.036}$ & $\mathbf{0.946}_{0.003}$ & $0.540_{0.024}$ & $0.734_{0.027}$ & $\mathbf{0.951}_{0.011}$ & $0.013_{0.072}$ & $0.730_{0.014}$ & $\mathbf{0.953}_{0.004}$ & $0.451_{0.085}$ & $0.471_{0.341}$ \\
            & \gtp & $\underline{0.805}_{0.017}$ & $\underline{0.920}_{0.003}$ & $0.458_{0.082}$ & $0.807_{0.019}$ & $\underline{0.923}_{0.003}$ & $0.382_{0.076}$ & $0.773_{0.015}$ & $0.912_{0.003}$ & $0.384_{0.105}$ & $0.727_{0.143}$ \\
            & \iibmil & $0.621_{0.050}$ & $0.881_{0.012}$ & $0.000_{0.000}$ & $0.667_{0.011}$ & $0.889_{0.011}$ & $0.000_{0.000}$ & $0.723_{0.061}$ & $0.893_{0.008}$ & $\mathbf{0.686}_{0.016}$ & $\underline{0.922}_{0.010}$ \\
            & \camil & $0.805_{0.028}$ & $0.911_{0.004}$ & $\underline{0.649}_{0.054}$ & $\mathbf{0.811}_{0.014}$ & $0.913_{0.003}$ & $0.619_{0.039}$ & $\underline{0.792}_{0.013}$ & $\underline{0.917}_{0.002}$ & $0.639_{0.039}$ & $0.918_{0.018}$ \\
			 \bottomrule
		\end{tabular}%
	\end{adjustbox}
% \end{table}
\end{sidewaystable}

\newpage
\section{Additional figures}
\label{appendix:section:tables_figures}

\begin{figure}[h]
    \centering
    \centering
    \begin{subfigure}[b]{0.4\textwidth}
        \centering
        \includegraphics[trim={0cm 0cm 0cm 0cm},clip,height=90pt]{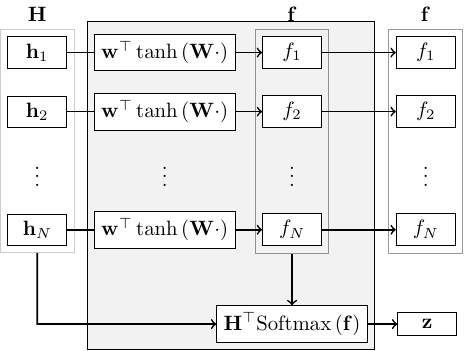}
        \caption{\abmil.}
        \label{fig:abmil_closerlook}
    \end{subfigure}
    \hfill
    \begin{subfigure}[b]{0.4\textwidth}
        \centering
        \includegraphics[trim={0cm 0cm 0cm 0cm},clip,height=90pt]{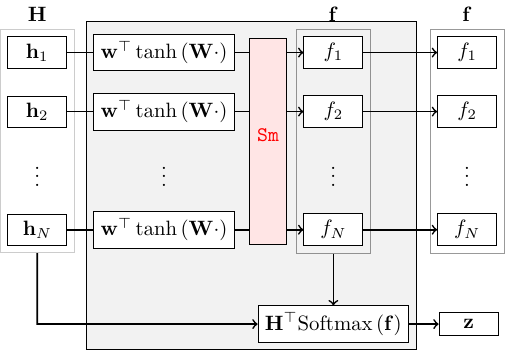}
        \caption{\smoothattpoollate.}
        \label{fig:smoothatpool_late}
    \end{subfigure}
    \hfill
    \begin{subfigure}[b]{0.4\textwidth}
        \centering
        \includegraphics[trim={0cm 0cm 0cm 0cm},clip,height=90pt]{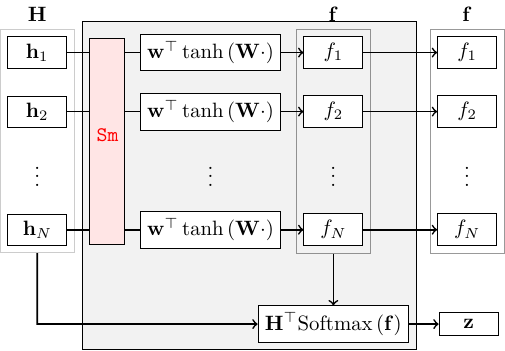}
        \caption{\smoothattpoolmid.}
        \label{fig:smoothatpool_mid}
    \end{subfigure} 
    \hfill
    \begin{subfigure}[b]{0.4\textwidth}
        \centering
        \includegraphics[trim={0cm 0cm 0cm 0cm},clip,height=90pt]{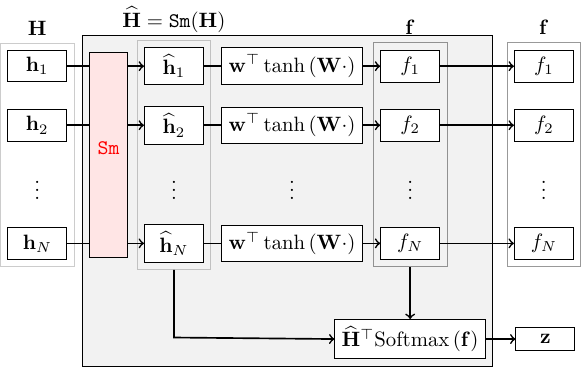}
        \caption{\smoothattpoolearly.}
        \label{fig:smoothatpool_early}
    \end{subfigure}
    \caption{Graphical representation of the different variants \smoothattpoollate, \smoothattpoolmid, \smoothattpoolearly. The well-known ABMIL, which we build upon, is shown in (a).} % The proposed \smoothattpool\ builds upon \abmil\ (a) by introducing \smoothopp\ in three different places: late (b), mid (c), and early (d).}
    \label{fig:attpooling_variants}
    % \vspace{-3mm}
\end{figure}

% \subsection{Attention maps}

% \begin{figure}

    \begin{center}
        \begin{adjustbox}{width=\textwidth}
        \centering
        \begin{tabular}{c}
            \includegraphics[trim={0cm 0cm 0cm 0cm},clip,width=0.85\textwidth]
            {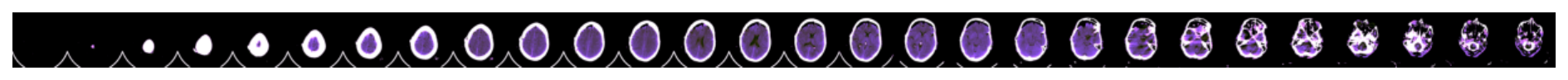} \\
            CT scan \\
            \includegraphics[trim={0cm 0cm 0cm 0cm},clip,width=0.85\textwidth]
            {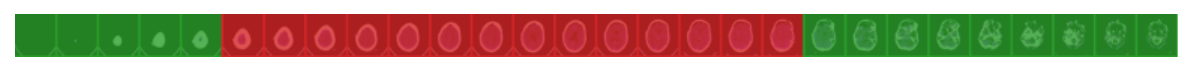} \\
            Slice labels \\
            \includegraphics[trim={0cm 0cm 0cm 0cm},clip,width=0.85\textwidth]{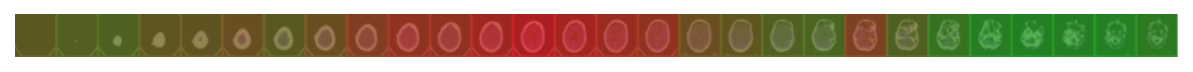} \\
            \smoothattpool \\
            \includegraphics[trim={0cm 0cm 0cm 0cm},clip,width=0.85\textwidth]{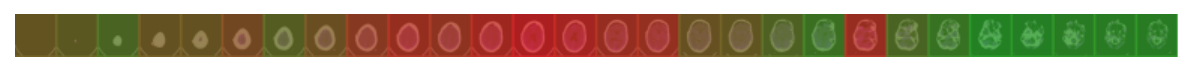} \\
            \abmil \\
            \includegraphics[trim={0cm 0cm 0cm 0cm},clip,width=0.85\textwidth]{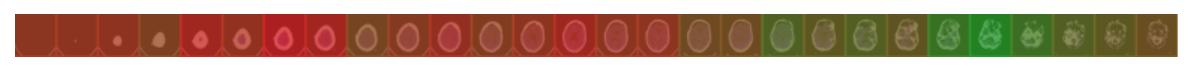} \\
            \clam \\
            \includegraphics[trim={0cm 0cm 0cm 0cm},clip,width=0.85\textwidth]{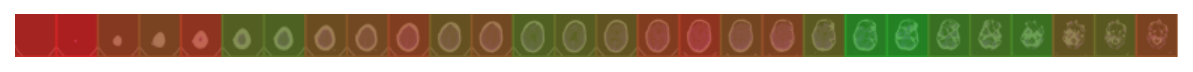} \\
            \dsmil \\
            \includegraphics[trim={0cm 0cm 0cm 0cm},clip,width=0.85\textwidth]{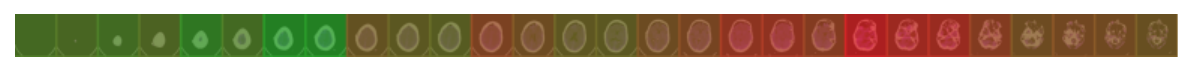} \\
            \dftdmil \\
            \includegraphics[trim={0cm 0cm 0cm 0cm},clip,width=0.85\textwidth]{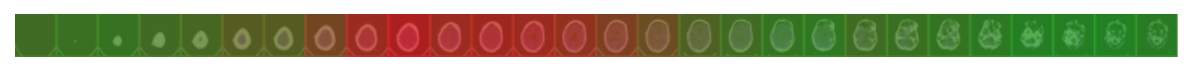} \\
            \smoothtransformerattpool \\
            \includegraphics[trim={0cm 0cm 0cm 0cm},clip,width=0.85\textwidth]{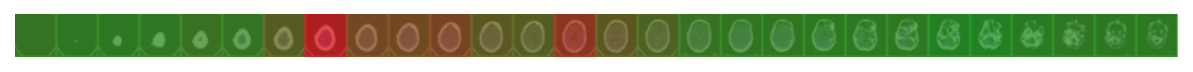} \\
            \transmil \\
            \includegraphics[trim={0cm 0cm 0cm 0cm},clip,width=0.85\textwidth]{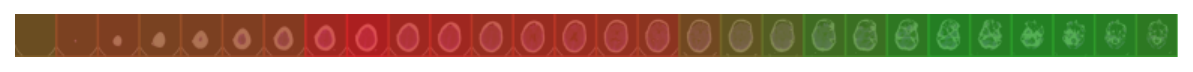} \\
            \setmil \\
            \includegraphics[trim={0cm 0cm 0cm 0cm},clip,width=0.85\textwidth]{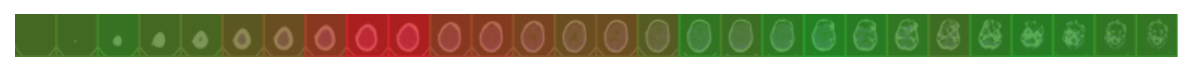} \\
            \gtp \\
            \includegraphics[trim={0cm 0cm 0cm 0cm},clip,width=0.85\textwidth]{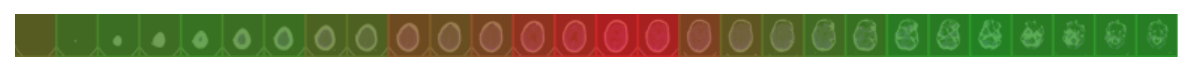} \\
            \camil \\
        \end{tabular}
        \end{adjustbox}
        \captionof{figure}{RSNA attention maps.}
        \label{fig:attmaps-rsna}
    \end{center}
% \end{figure}

\begin{figure}
    \begin{center}
        \centering
        \begin{tabular}{cccccc}
            \includegraphics[trim={0cm 0cm 0cm 0cm},clip,width=0.12\textwidth]{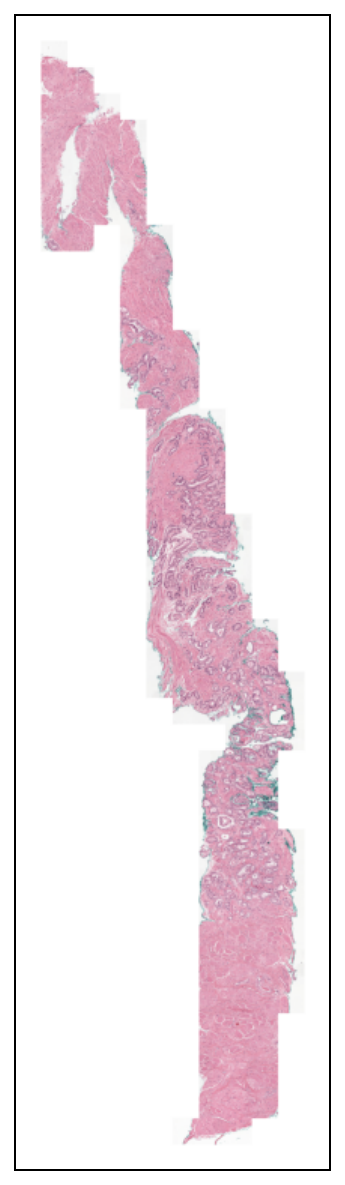}
            & 
            \includegraphics[trim={0cm 0cm 0cm 0cm},clip,width=0.12\textwidth]{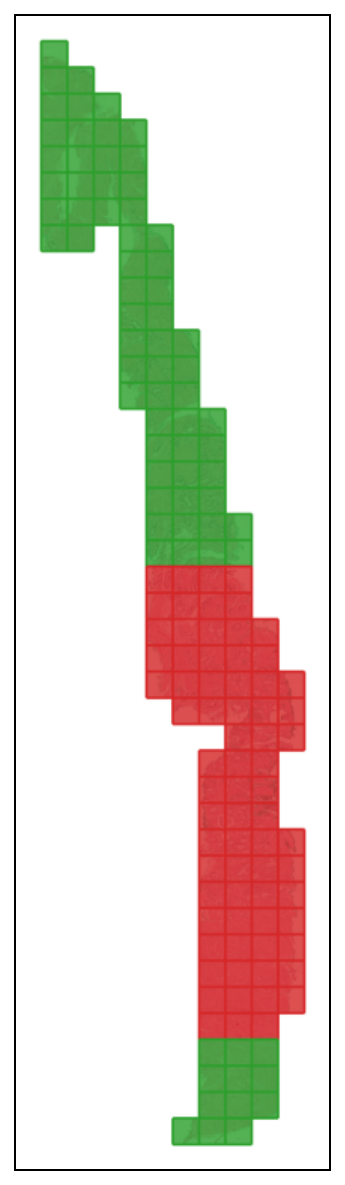}
            &
            \multicolumn{1}{l}{
                \includegraphics[trim={0cm 0cm 0cm 0cm},clip,height=168pt]
            {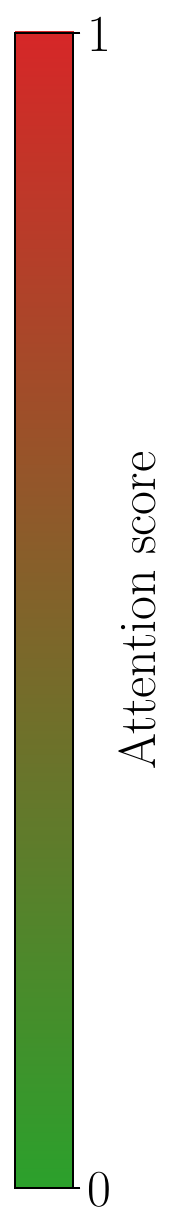}
            }
            \\
             WSI &  Patch labels \\
            \includegraphics[trim={0cm 0cm 0cm 0cm},clip,width=0.12\textwidth]
            {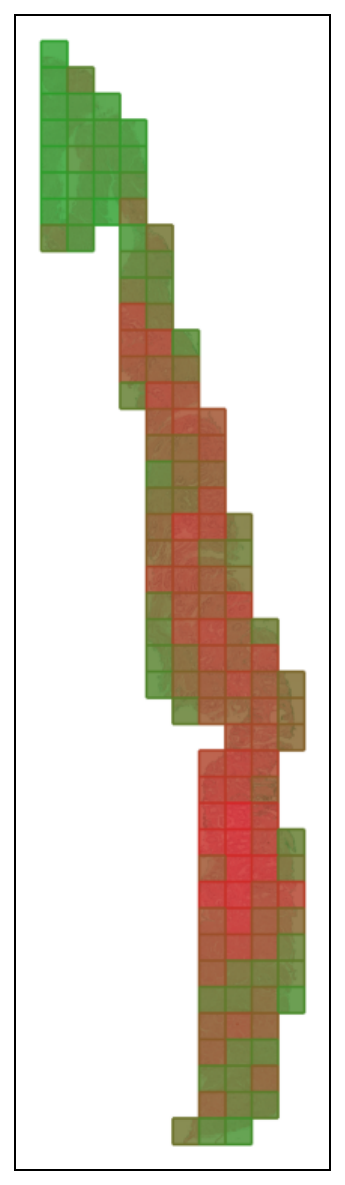}
            & 
            \includegraphics[trim={0cm 0cm 0cm 0cm},clip,width=0.12\textwidth]
            {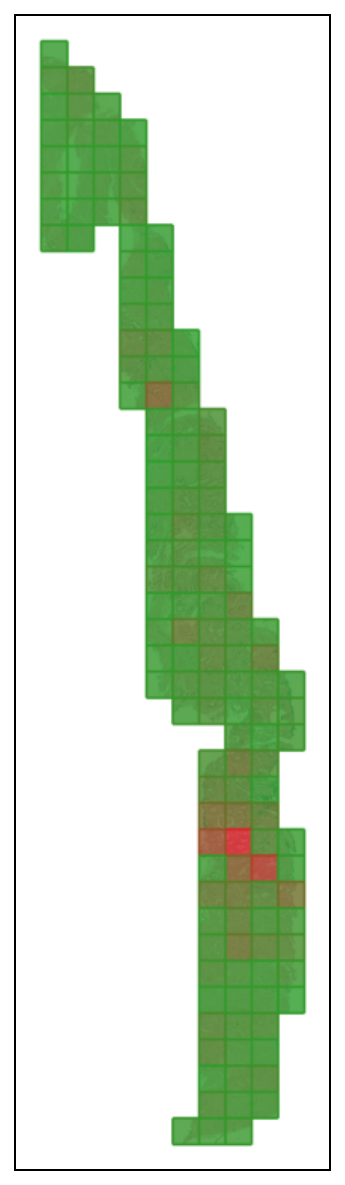}
            & 
            \includegraphics[trim={0cm 0cm 0cm 0cm},clip,width=0.12\textwidth]{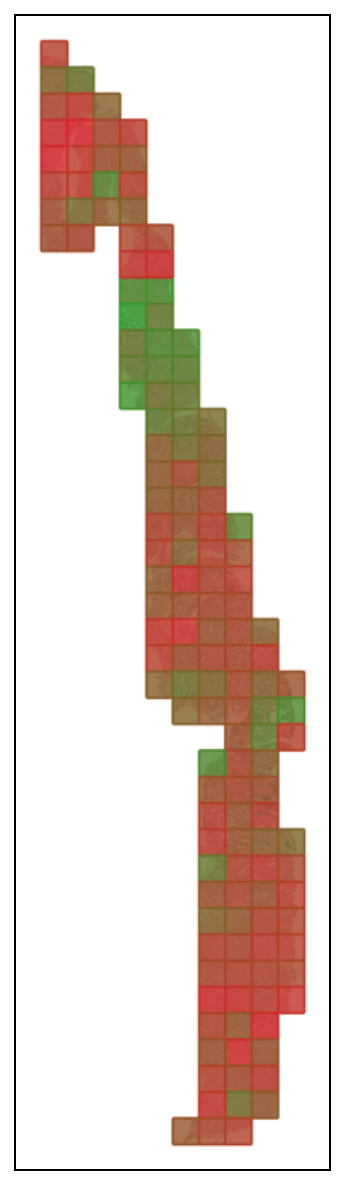}        & 
            \includegraphics[trim={0cm 0cm 0cm 0cm},clip,width=0.12\textwidth]
            {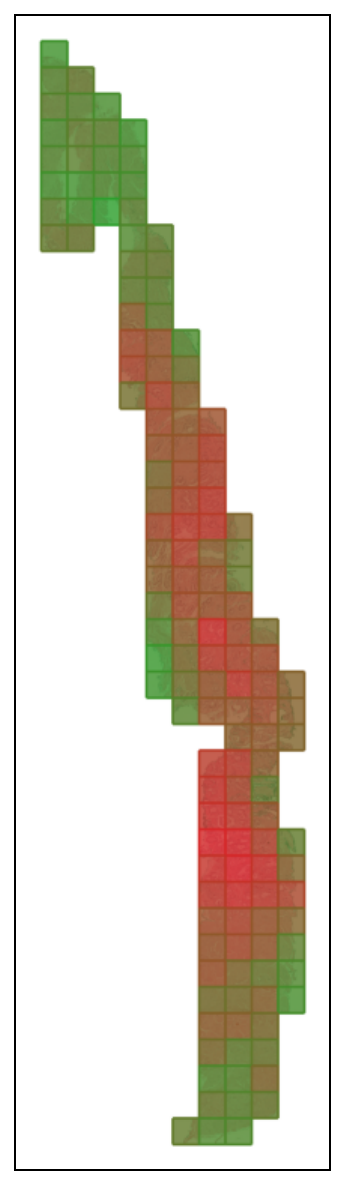}
            & 
            \includegraphics[trim={0cm 0cm 0cm 0cm},clip,width=0.12\textwidth]
            {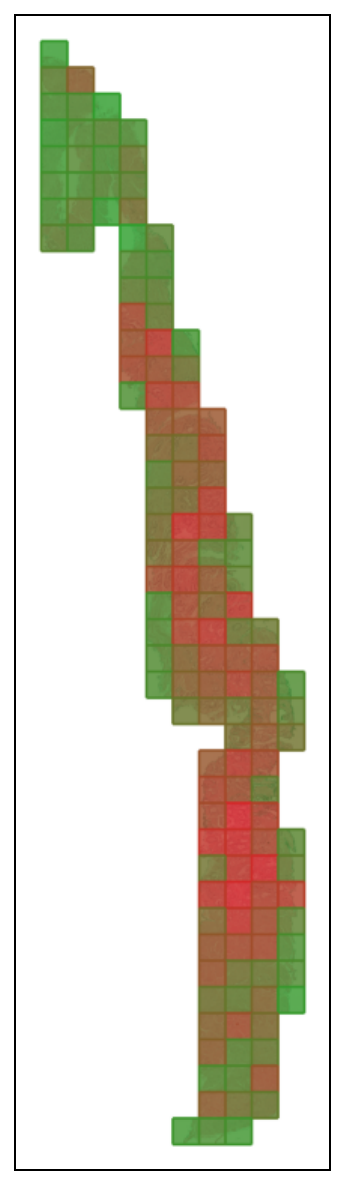}
            \\
             \smoothtransformerattpool &  \transmil &  \setmil &  \gtp &  \camil \\
            \includegraphics[trim={0cm 0cm 0cm 0cm},clip,width=0.12\textwidth]
            {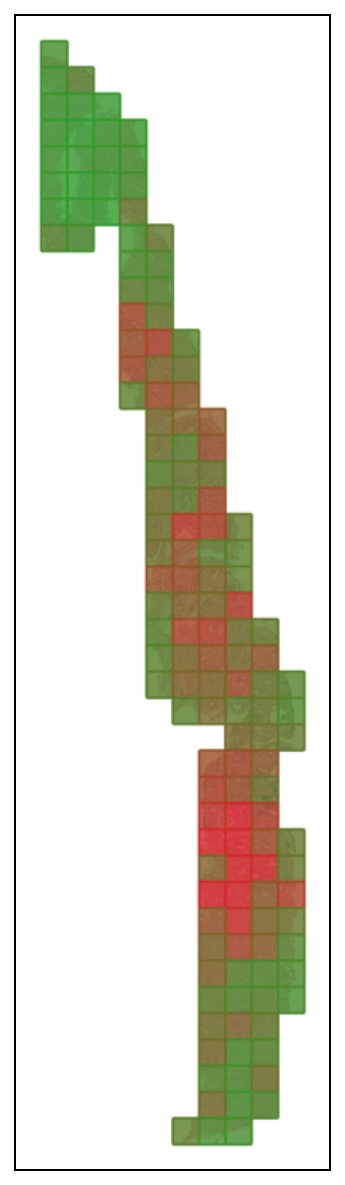}
            & 
            \includegraphics[trim={0cm 0cm 0cm 0cm},clip,width=0.12\textwidth]
            {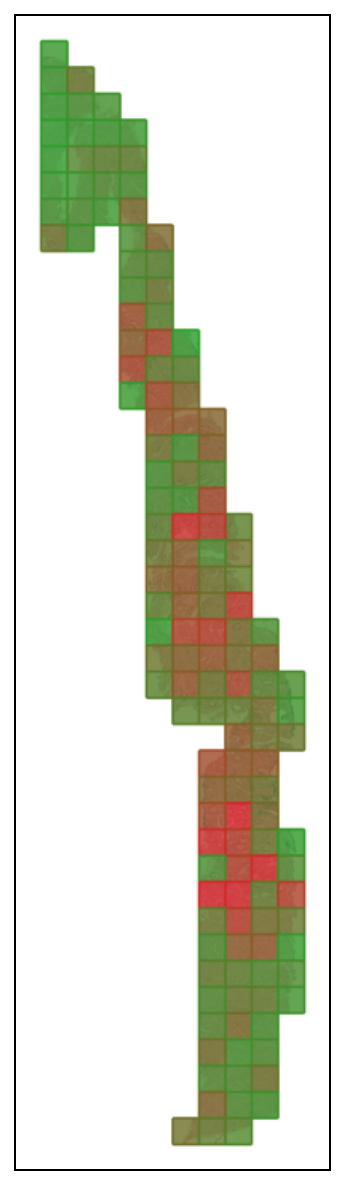}
            & 
            \includegraphics[trim={0cm 0cm 0cm 0cm},clip,width=0.12\textwidth]{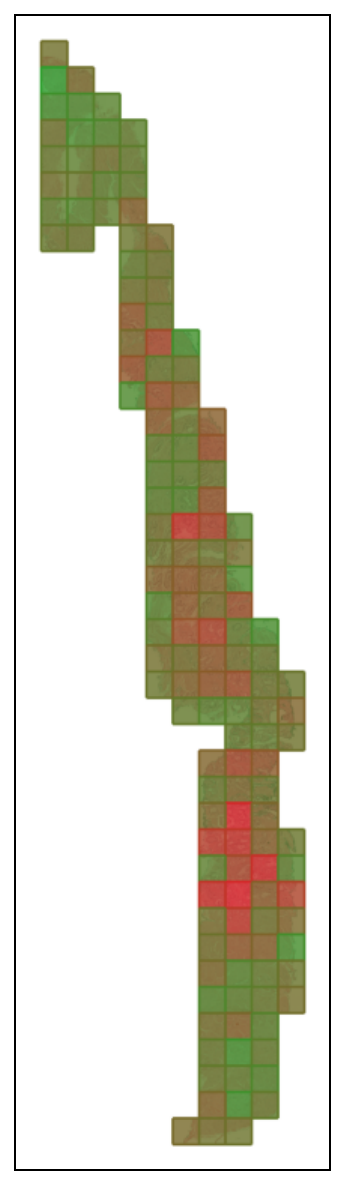}
            & 
            \includegraphics[trim={0cm 0cm 0cm 0cm},clip,width=0.12\textwidth]
            {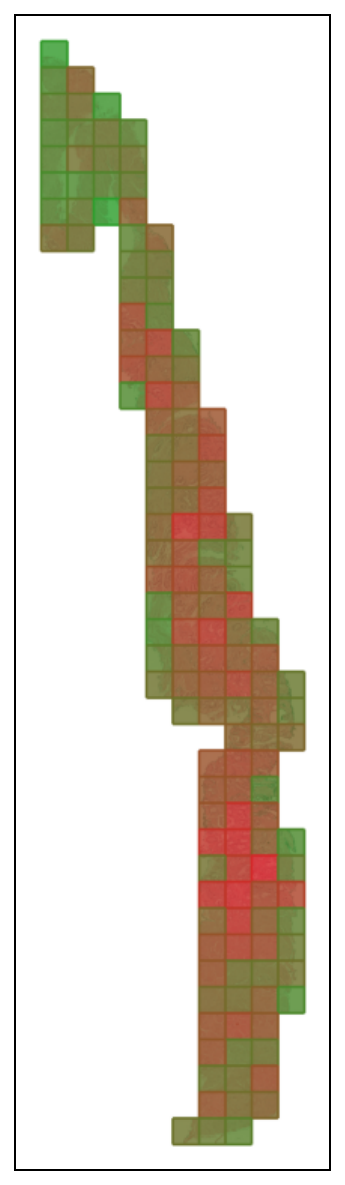}
            & 
            \includegraphics[trim={0cm 0cm 0cm 0cm},clip,width=0.12\textwidth]{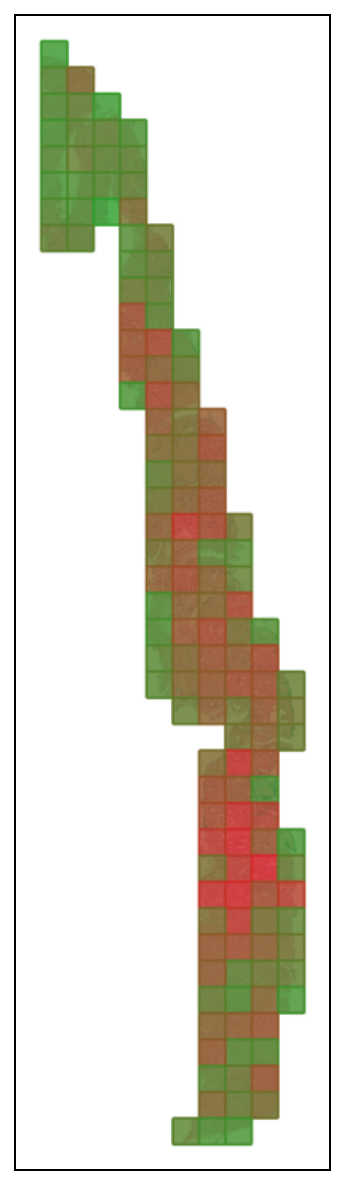}
            \\        
             \smoothattpool &  \abmil &  \clam &  \dsmil &  \dftdmil
        \end{tabular}
        \captionof{figure}{PANDA attention maps.}
        \label{fig:attmaps-panda-appendix}
    \end{center}
\end{figure}

\begin{figure}
    \begin{center}
        \centering
        \begin{adjustbox}{width=\textwidth}
        \begin{tabular}{ccccc}
            \includegraphics[trim={0cm 0cm 0cm 0cm},clip,width=0.18\textwidth]{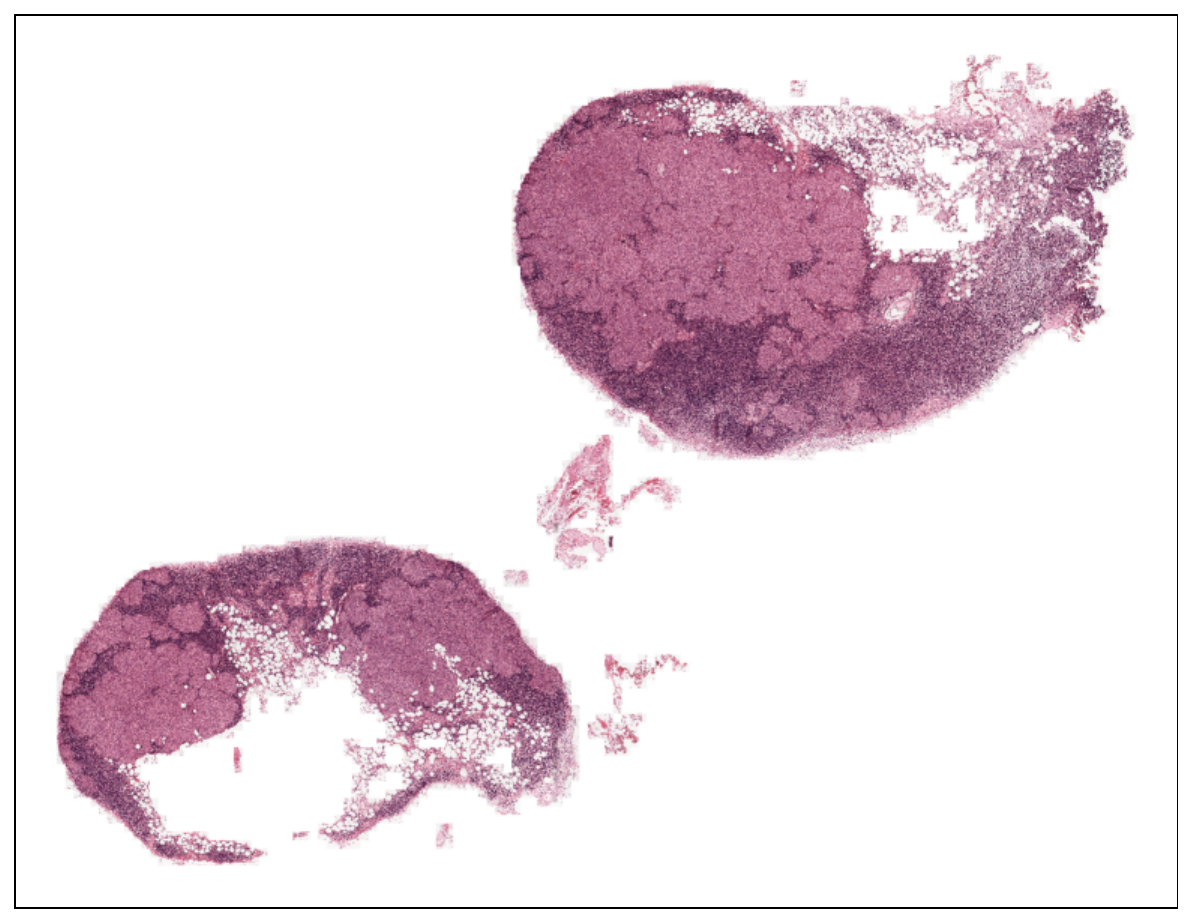}
            & 
            \includegraphics[trim={0cm 0cm 0cm 0cm},clip,width=0.18\textwidth]
            {img/camelyon_wsi_patched_labels.pdf}
            & 
            \multicolumn{1}{l}{
                \includegraphics[trim={0cm 0cm 0cm 0cm},clip,height=58pt]
            {img/camelyon_attmap_bar.pdf}
            }        
            \\
             WSI &  Patch labels \\
            \includegraphics[trim={0cm 0cm 0cm 0cm},clip,width=0.18\textwidth]
            {img/camelyon_wsi_patched_smoothtransformer.pdf}
            & 
            \includegraphics[trim={0cm 0cm 0cm 0cm},clip,width=0.18\textwidth]
            {img/camelyon_wsi_patched_transmil.pdf}
            & 
            \includegraphics[trim={0cm 0cm 0cm 0cm},clip,width=0.18\textwidth]{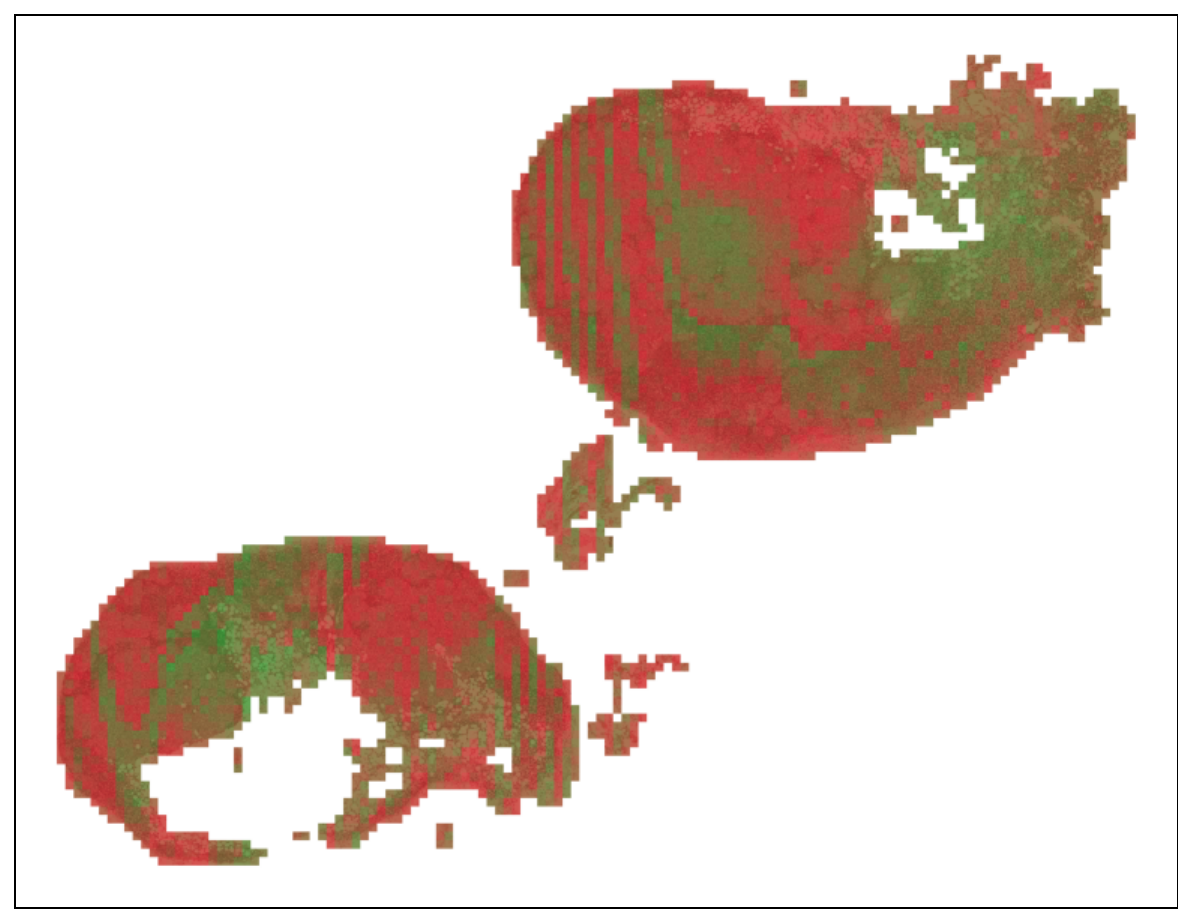}
            & 
            \includegraphics[trim={0cm 0cm 0cm 0cm},clip,width=0.18\textwidth]
            {img/camelyon_wsi_patched_gtp.pdf}
            & 
            \includegraphics[trim={0cm 0cm 0cm 0cm},clip,width=0.18\textwidth]
            {img/camelyon_wsi_patched_camil.pdf}
            \\
             \smoothtransformerattpool &  \transmil &  \setmil &  \gtp &  \camil \\
            \includegraphics[trim={0cm 0cm 0cm 0cm},clip,width=0.18\textwidth]
            {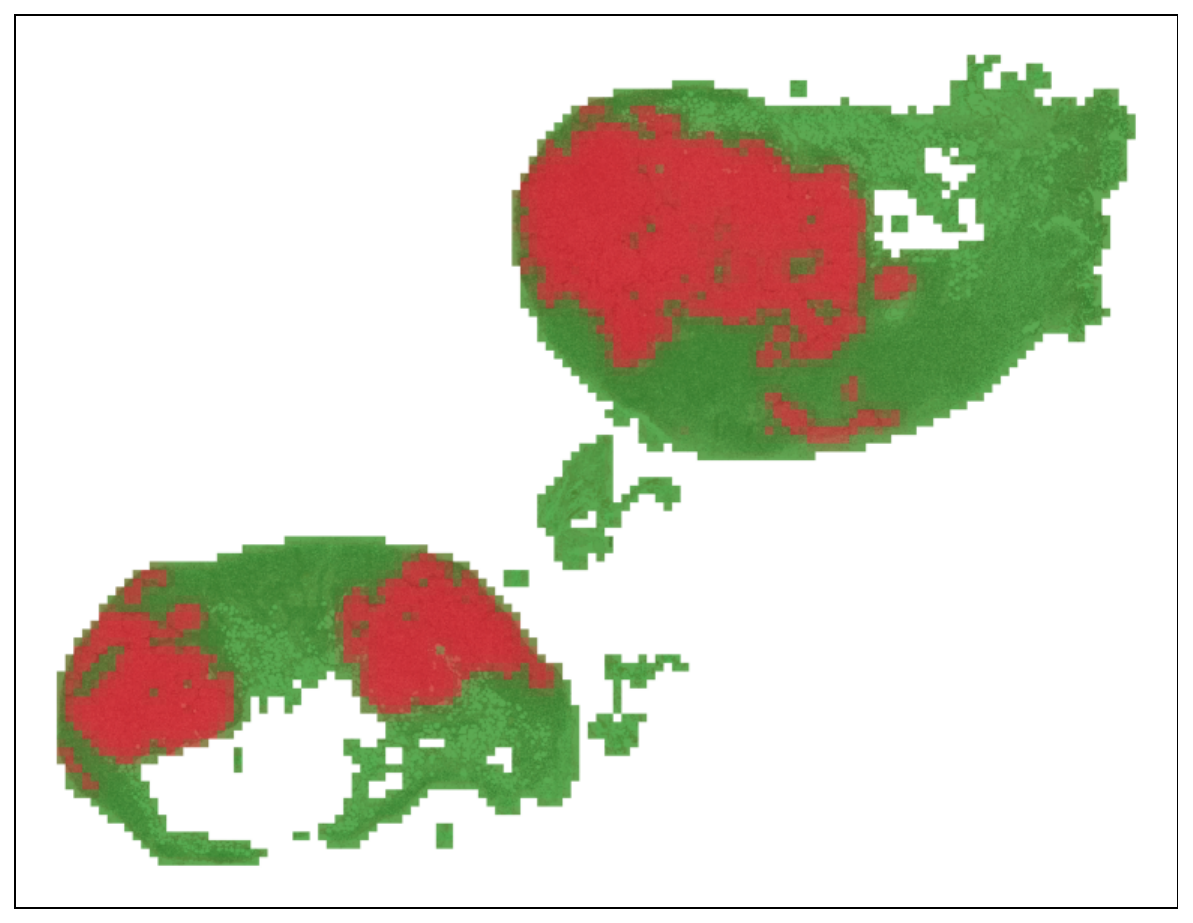}
            & 
            \includegraphics[trim={0cm 0cm 0cm 0cm},clip,width=0.18\textwidth]
            {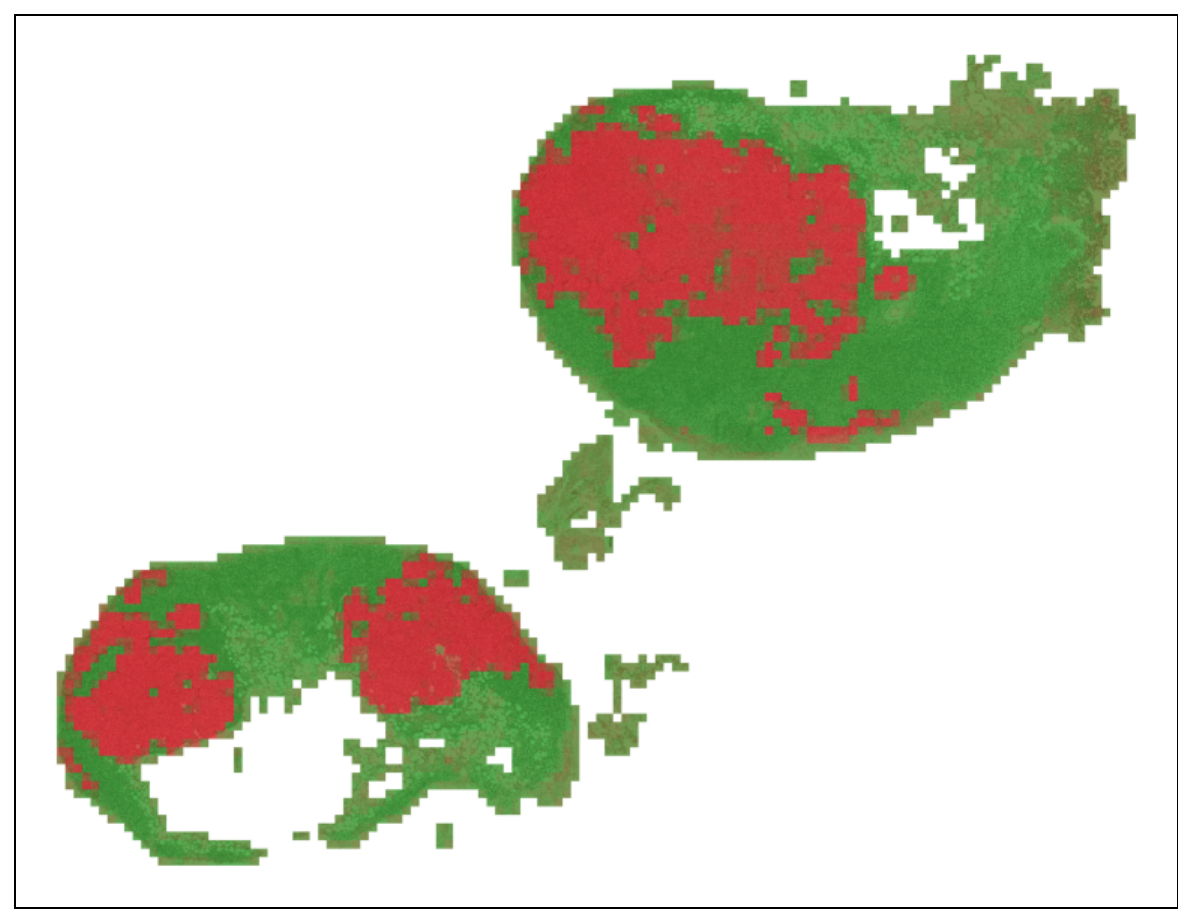}
            & 
            \includegraphics[trim={0cm 0cm 0cm 0cm},clip,width=0.18\textwidth]{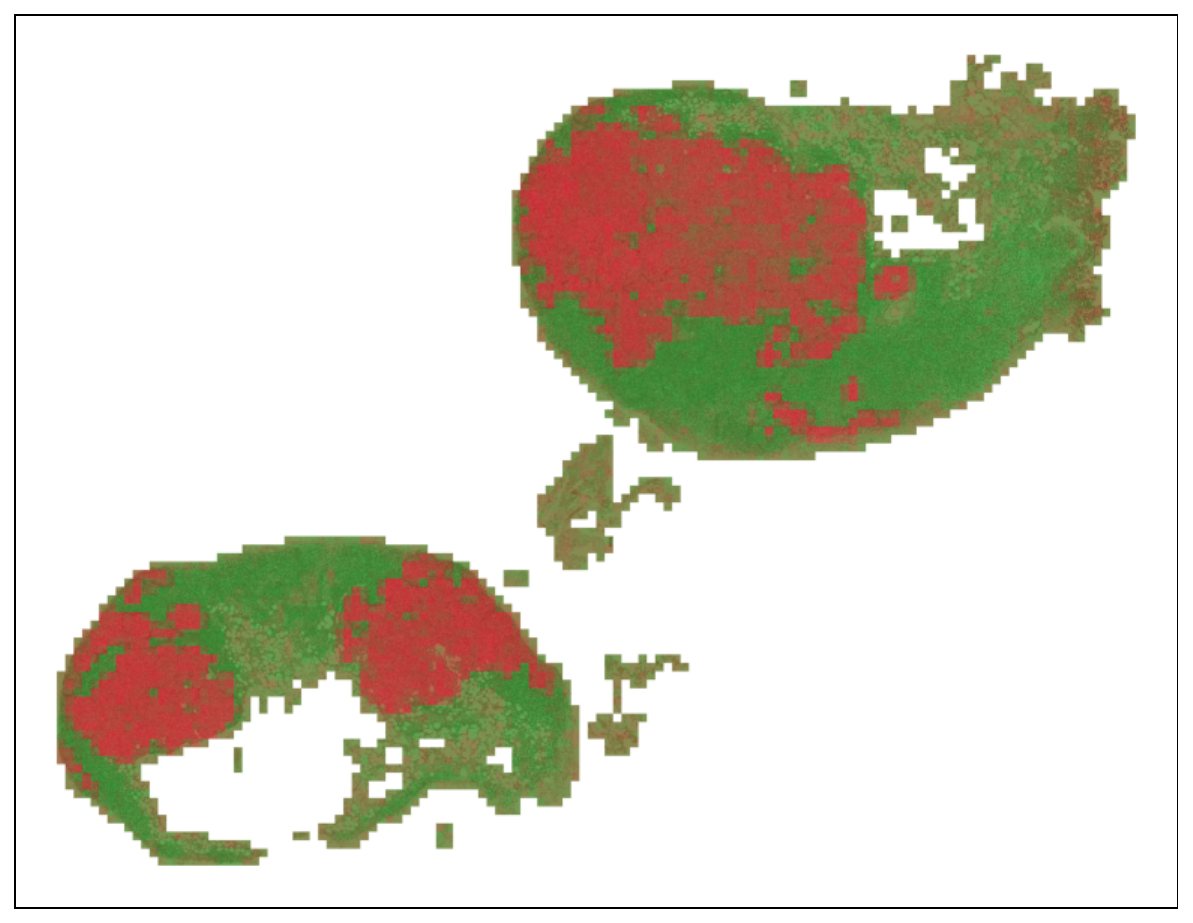}
            & 
            \includegraphics[trim={0cm 0cm 0cm 0cm},clip,width=0.18\textwidth]
            {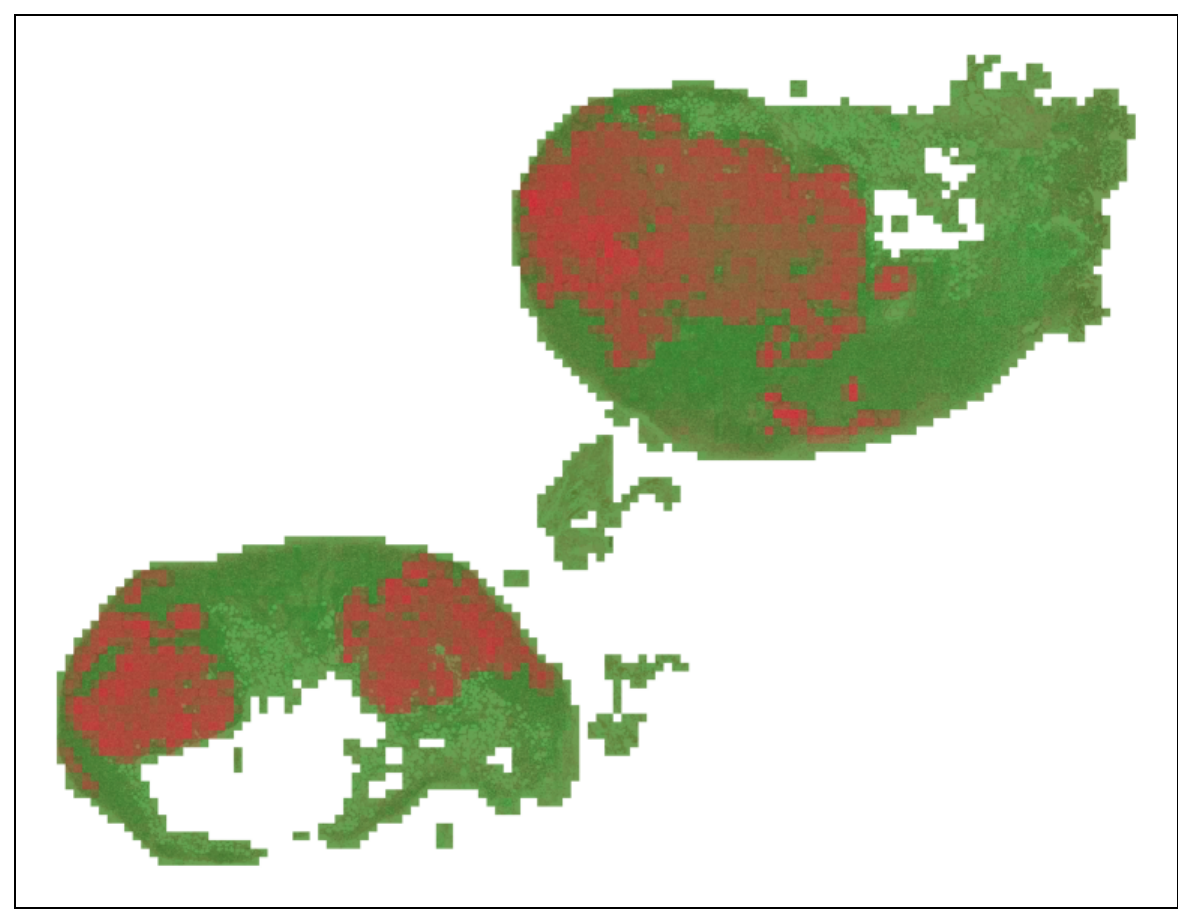}
            & 
            \includegraphics[trim={0cm 0cm 0cm 0cm},clip,width=0.18\textwidth]{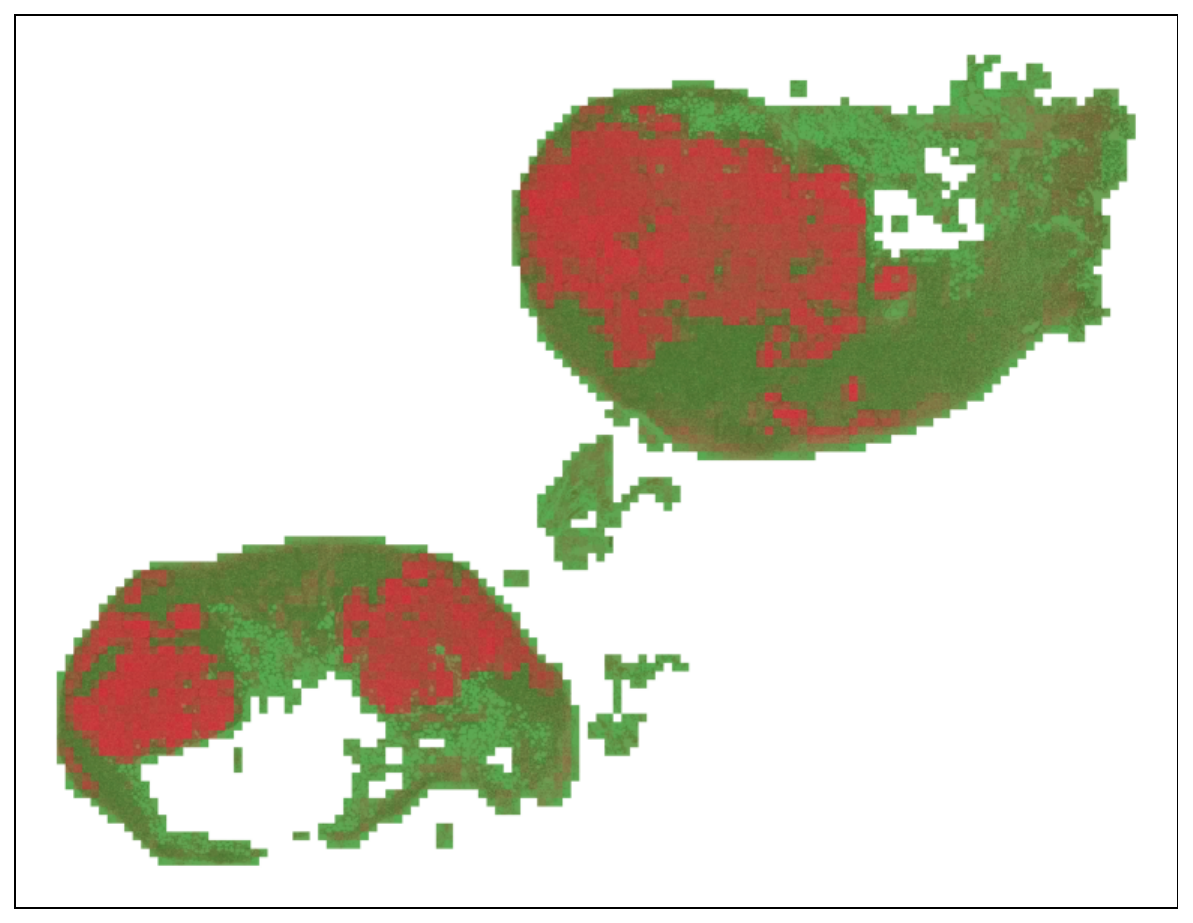}
            \\        
             \smoothattpool &  \abmil &  \clam & \dsmil &  \dftdmil
        \end{tabular}
        \end{adjustbox}
        \captionof{figure}{CAMELYON16 attention maps.}
        \label{fig:attmaps-camelyon-appendix}
    \end{center}
\end{figure}

\begin{figure}
    \begin{center}
    	\centering
    	\begin{adjustbox}{width=\textwidth}
    		\begin{tabular}{ccccc}
    			%			\includegraphics[trim={1cm 1cm 0.5cm 6cm},clip,width=0.19\textwidth]
    			%			{img/camelyon_wsi-test_021.pdf}
    			%			& 
    			Groung truth & ABMIL & SmAP($\alpha=0.1$) & SmAP($\alpha=0.5$) & SmAP($\alpha=0.9$) \\
    			\toprule
    			\includegraphics[trim={1cm 1cm 0.5cm 6.5cm},clip,width=0.19\textwidth]
    			{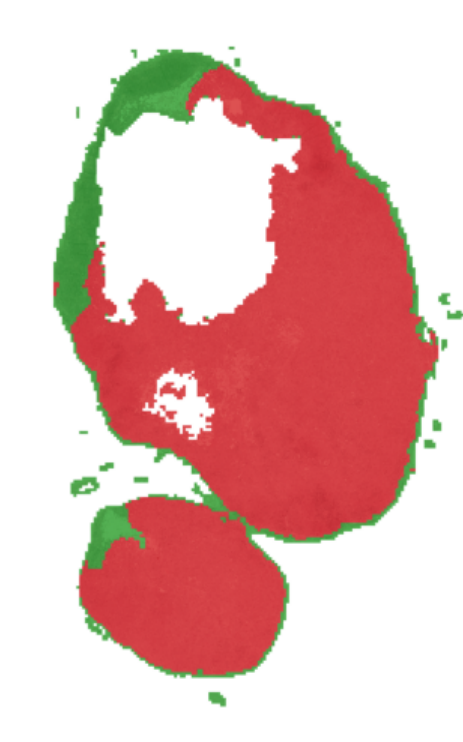}
    			& 
    			\includegraphics[trim={1cm 1cm 0.5cm 6.5cm},clip,width=0.19\textwidth]
    			{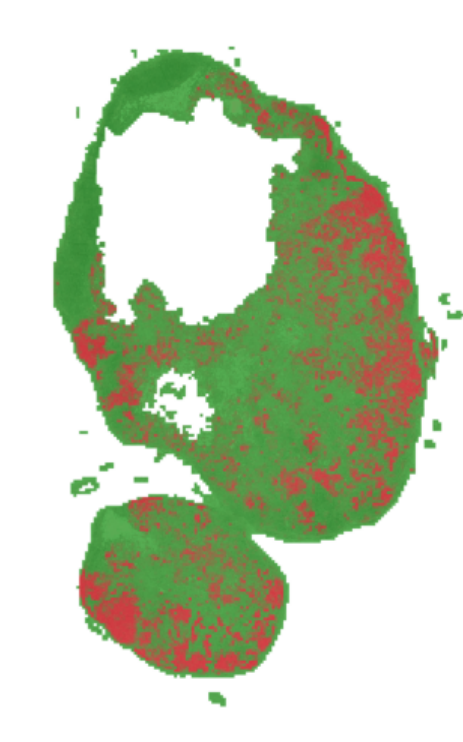}
    			& 
    			\includegraphics[trim={1cm 1cm 0.5cm 6.5cm},clip,width=0.19\textwidth]
    			{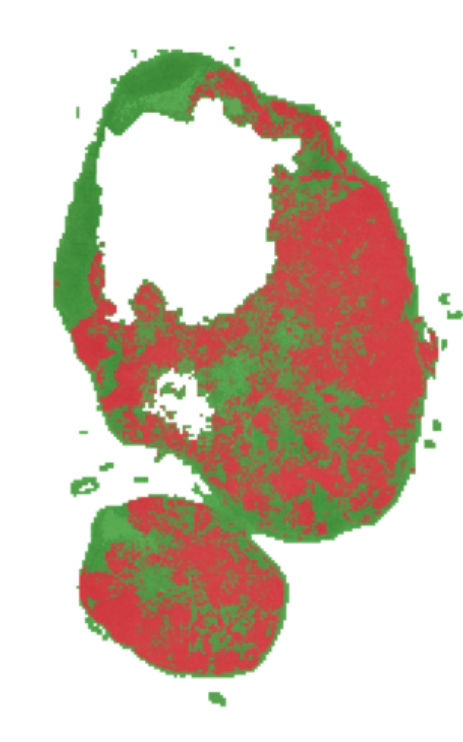}
    			& 
    			\includegraphics[trim={1cm 1cm 0.5cm 6.5cm},clip,width=0.19\textwidth]
    			{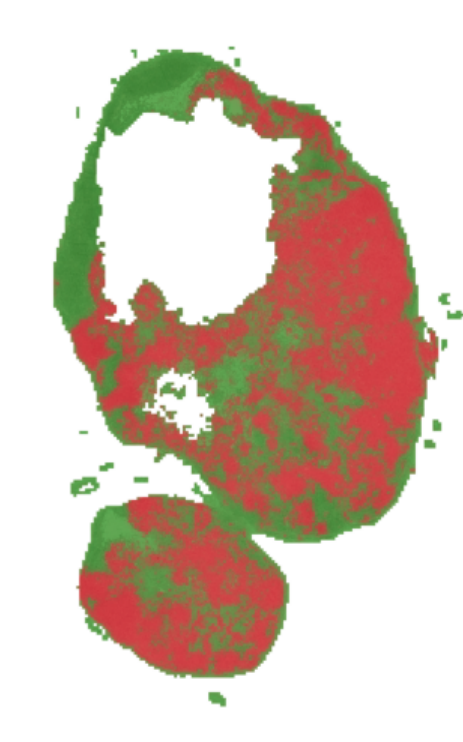}
    			& 
    			\includegraphics[trim={1cm 1cm 0.5cm 6.5cm},clip,width=0.19\textwidth]
    			{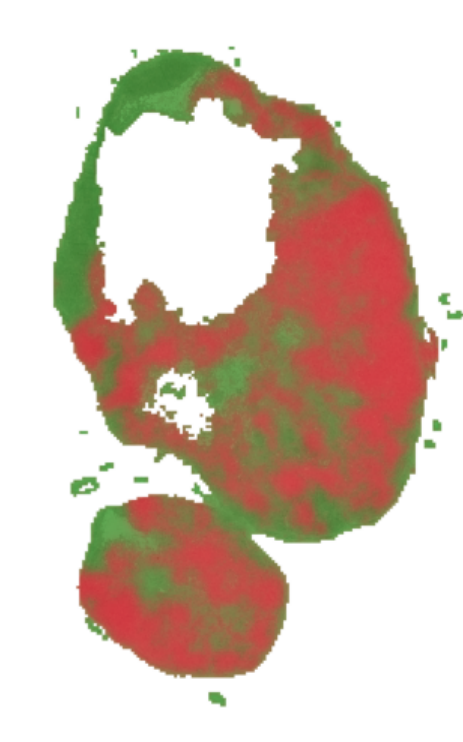} \\
    			\midrule
    			\includegraphics[trim={0.5cm 0.5cm 0.5cm 0.5cm},clip,width=0.19\textwidth]
    			{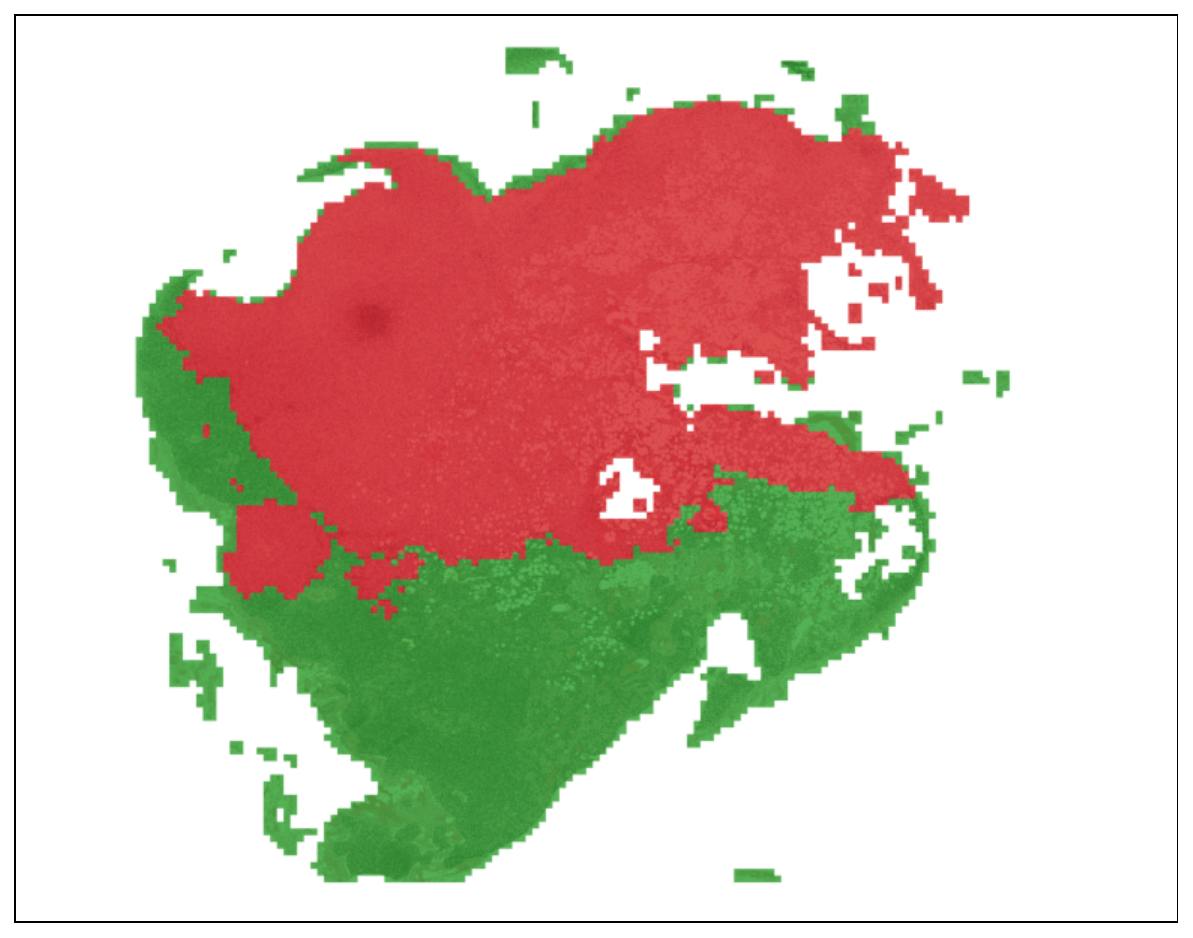}
    			& 
    			\includegraphics[trim={0.5cm 0.5cm 0.5cm 0.5cm},clip,width=0.19\textwidth]
    			{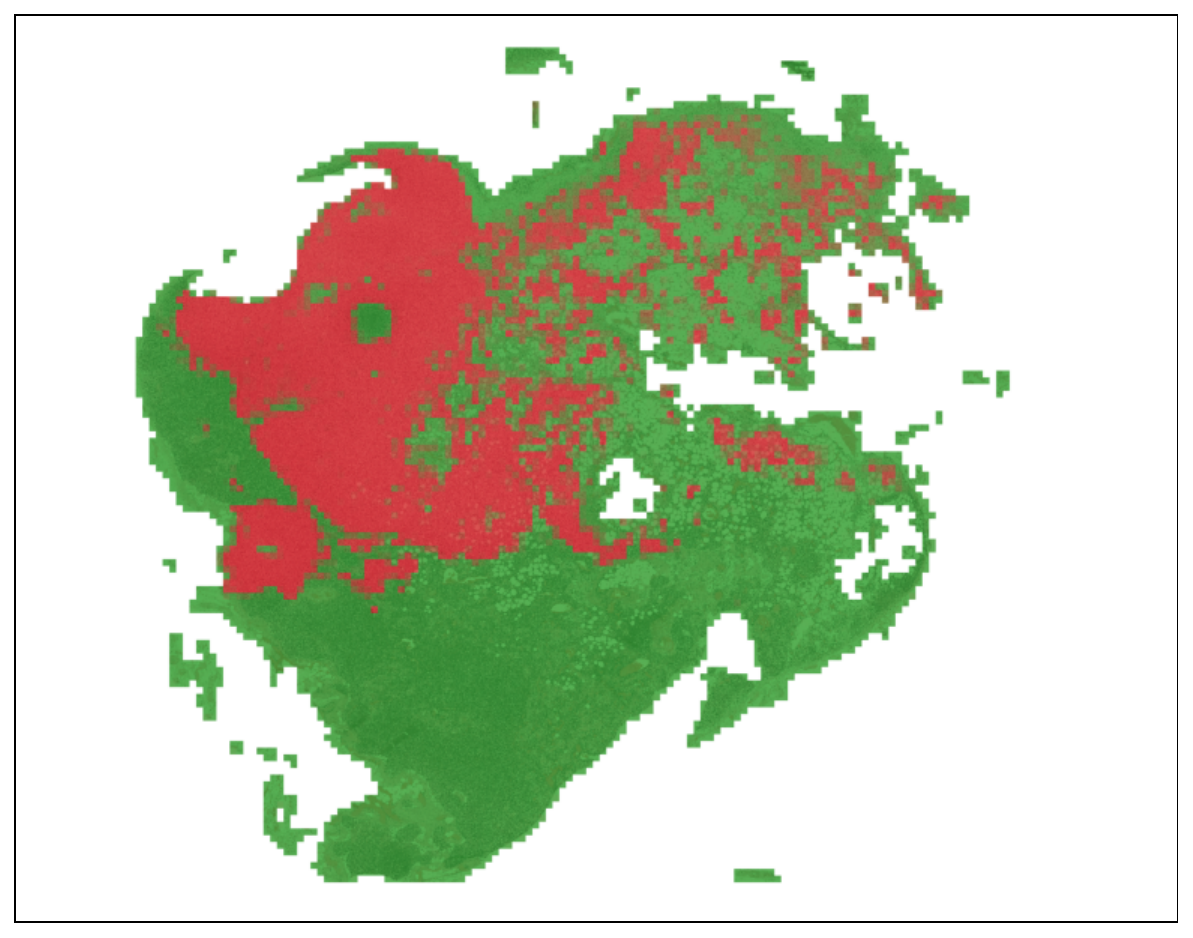}
    			& 
    			\includegraphics[trim={0.5cm 0.5cm 0.5cm 0.5cm},clip,width=0.19\textwidth]
    			{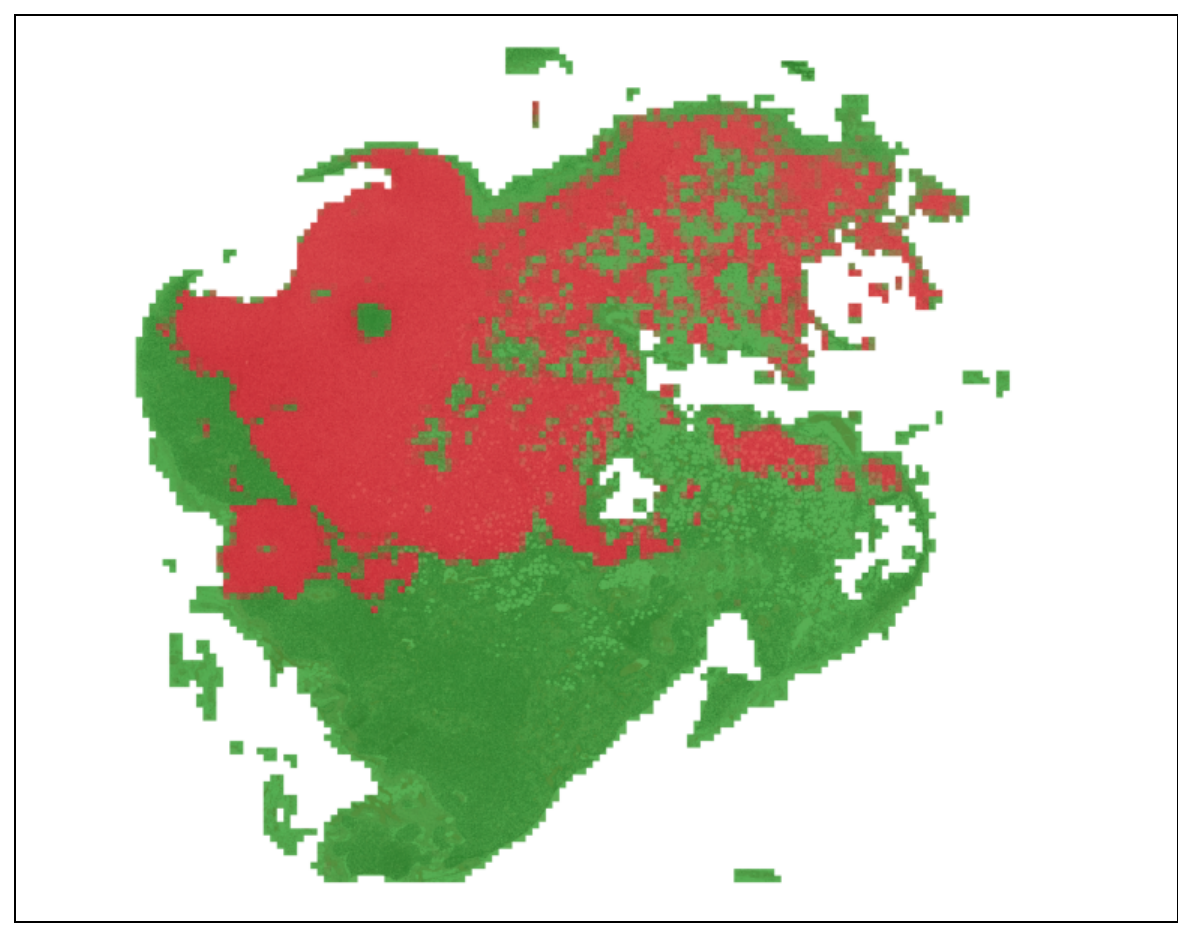}
    			& 
    			\includegraphics[trim={0.5cm 0.5cm 0.5cm 0.5cm},clip,width=0.19\textwidth]
    			{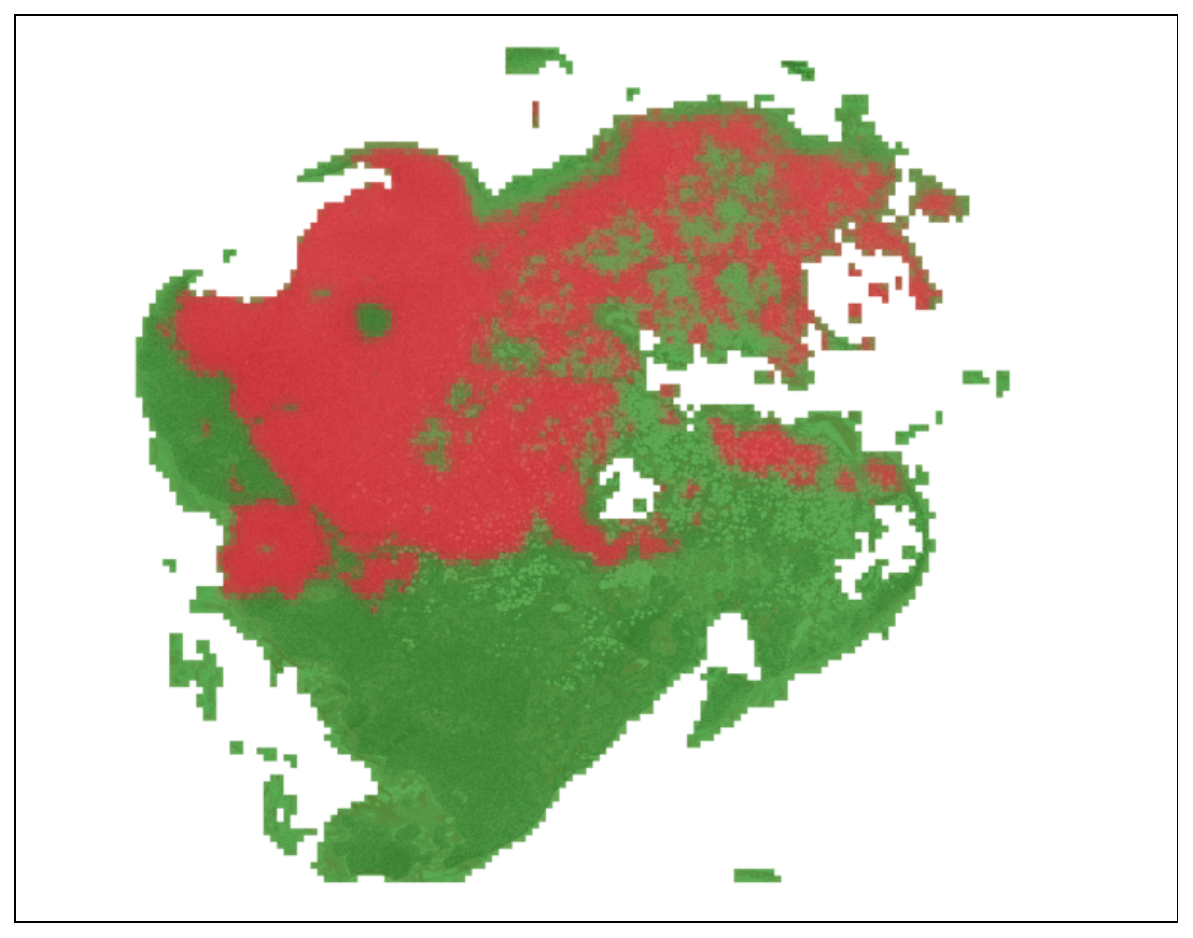}
    			& 
    			\includegraphics[trim={0.5cm 0.5cm 0.5cm 0.5cm},clip,width=0.19\textwidth]
    			{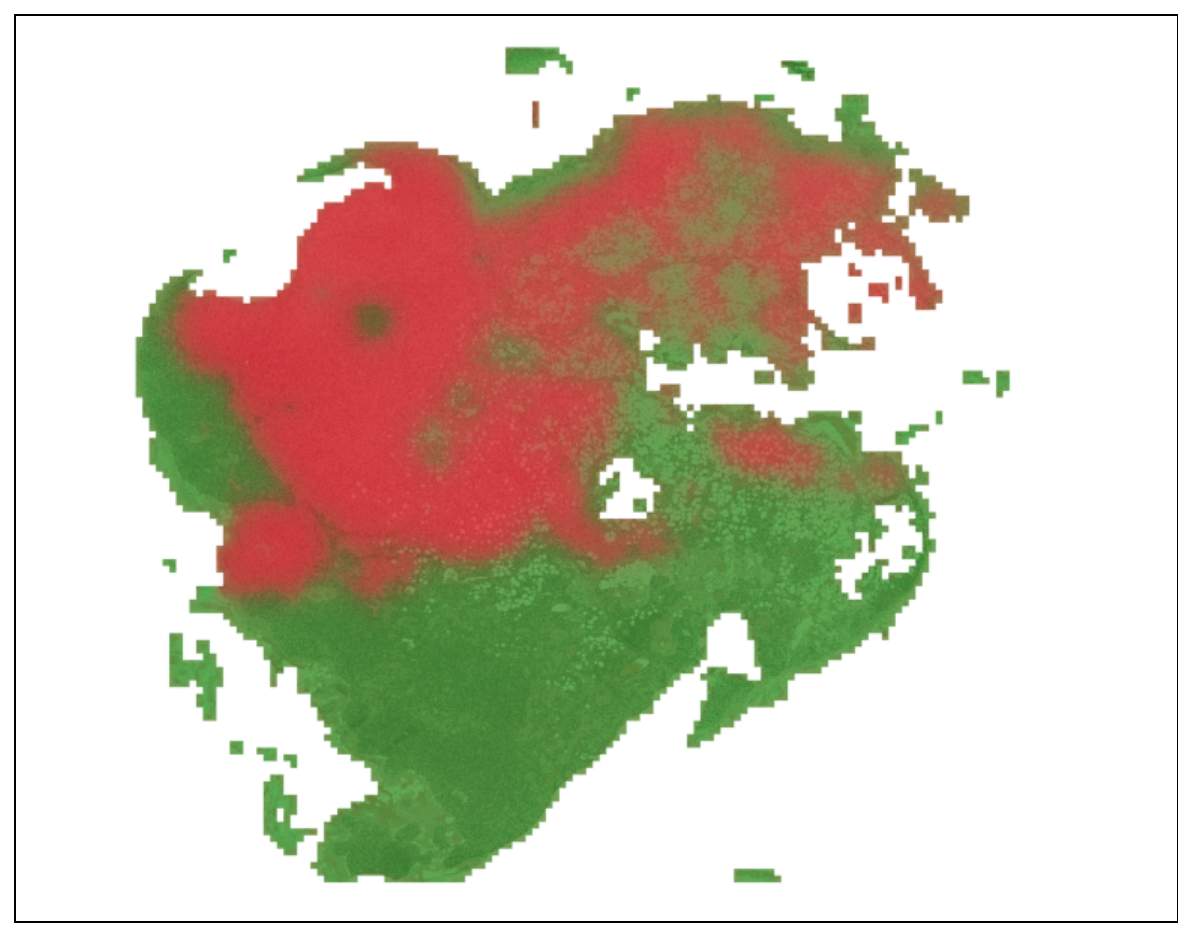} \\
    			\midrule
    			\includegraphics[trim={0.5cm 0.5cm 0.5cm 0.5cm},clip,width=0.19\textwidth]
    			{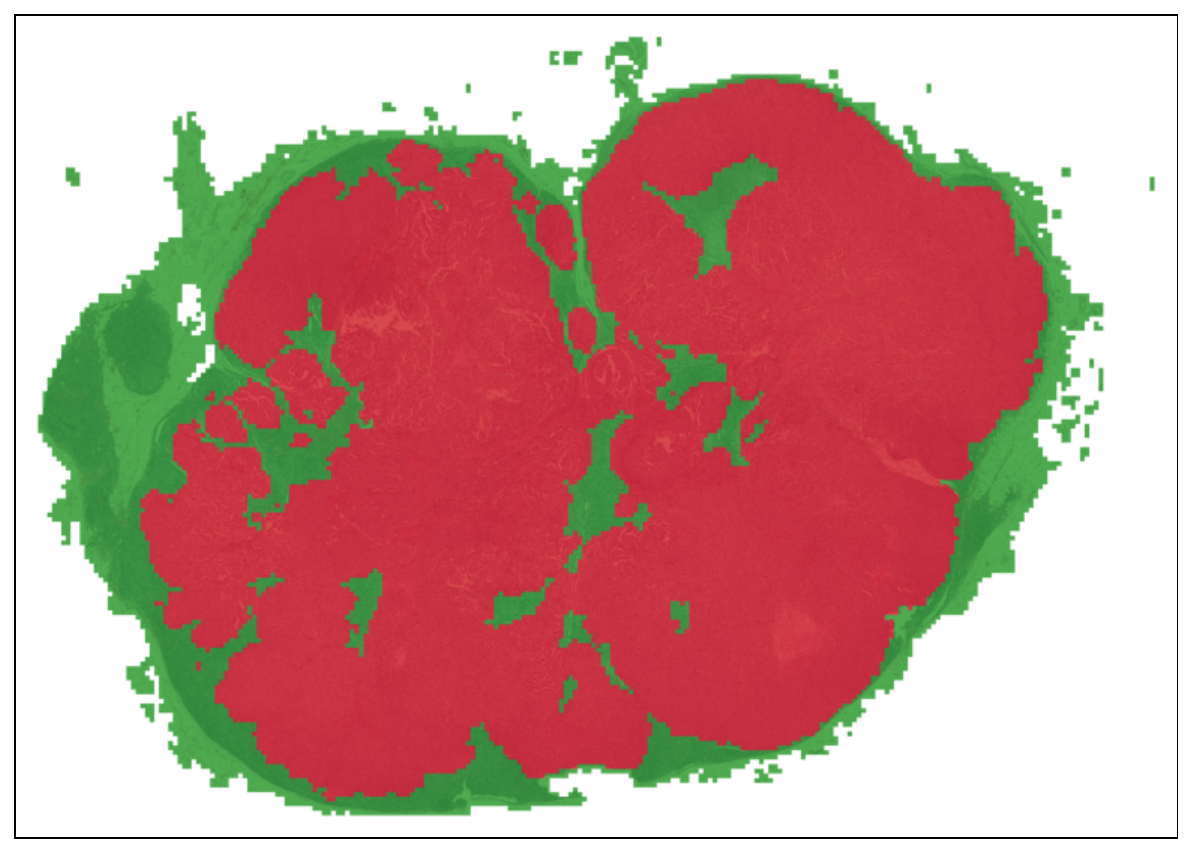}
    			& 
    			\includegraphics[trim={0.5cm 0.5cm 0.5cm 0.5cm},clip,width=0.19\textwidth]
    			{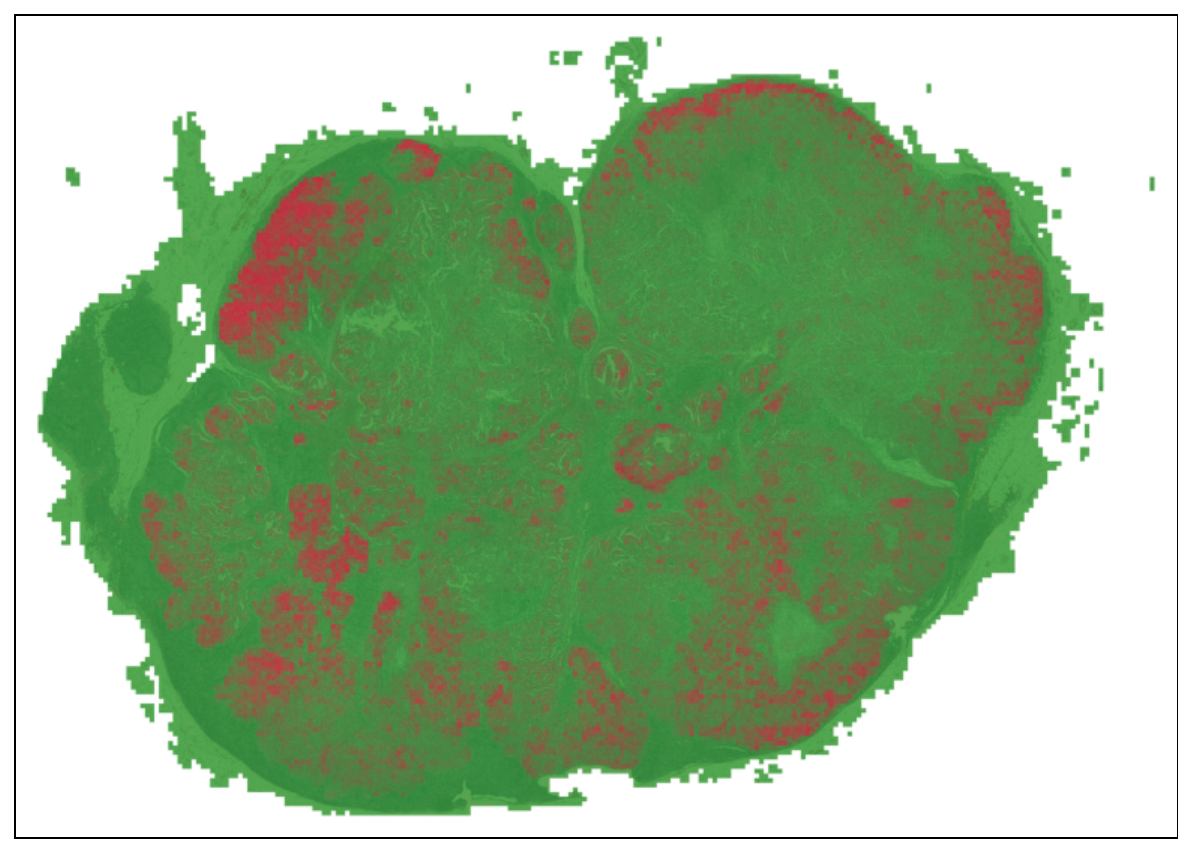}
    			& 
    			\includegraphics[trim={0.5cm 0.5cm 0.5cm 0.5cm},clip,width=0.19\textwidth]
    			{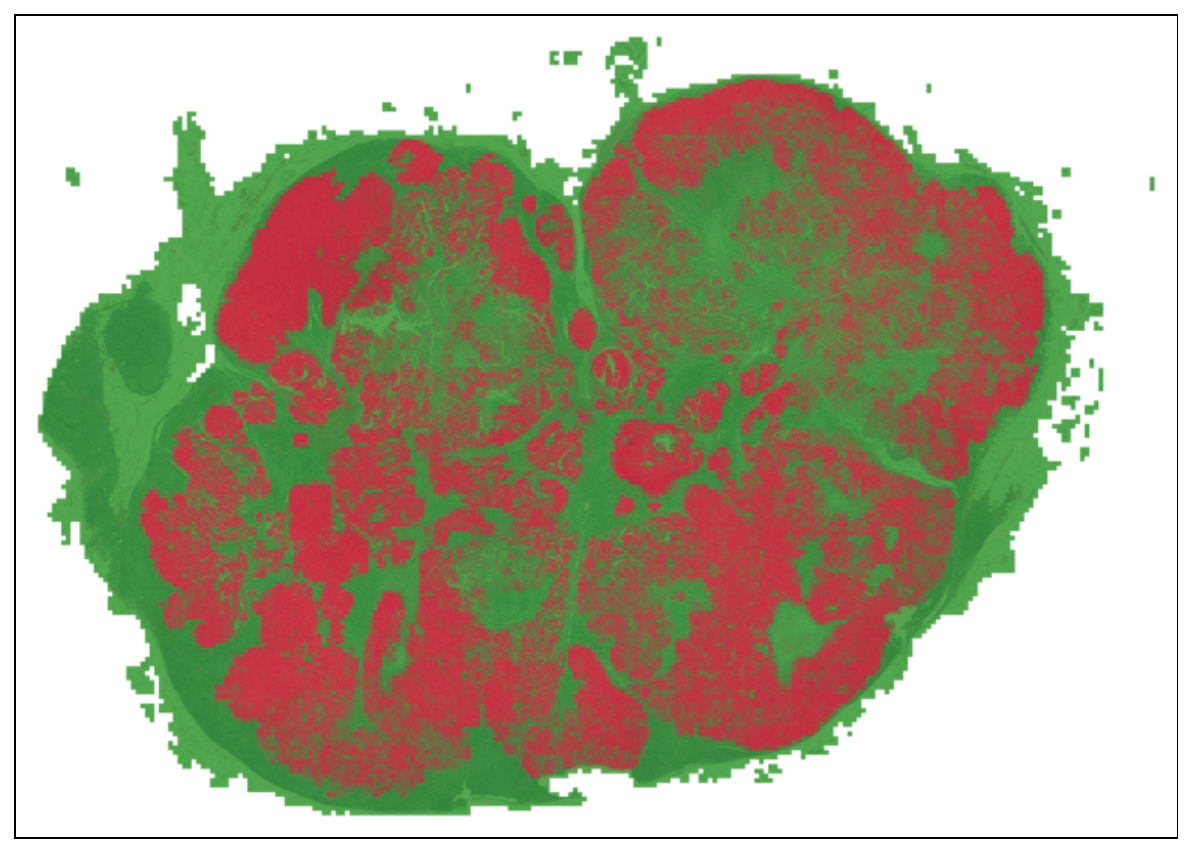}
    			& 
    			\includegraphics[trim={0.5cm 0.5cm 0.5cm 0.5cm},clip,width=0.19\textwidth]
    			{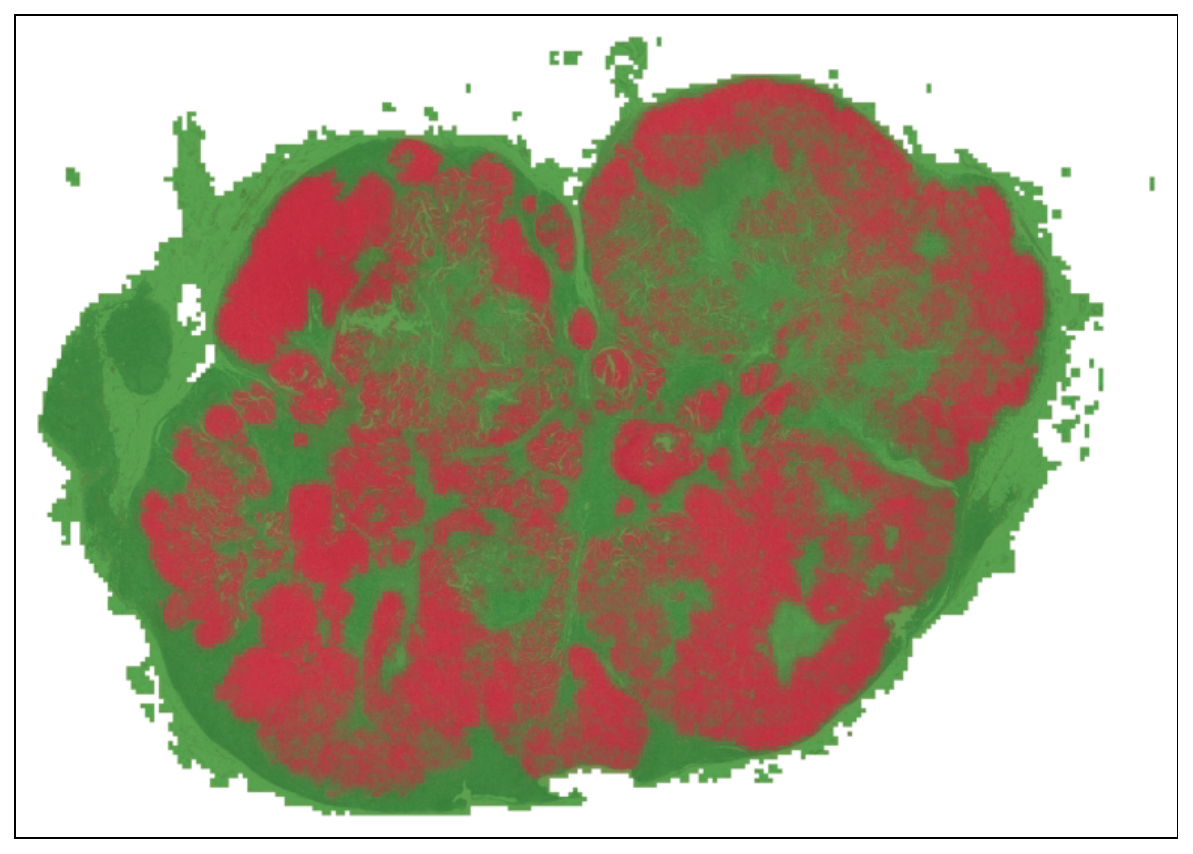}
    			& 
    			\includegraphics[trim={0.5cm 0.5cm 0.5cm 0.5cm},clip,width=0.19\textwidth]
    			{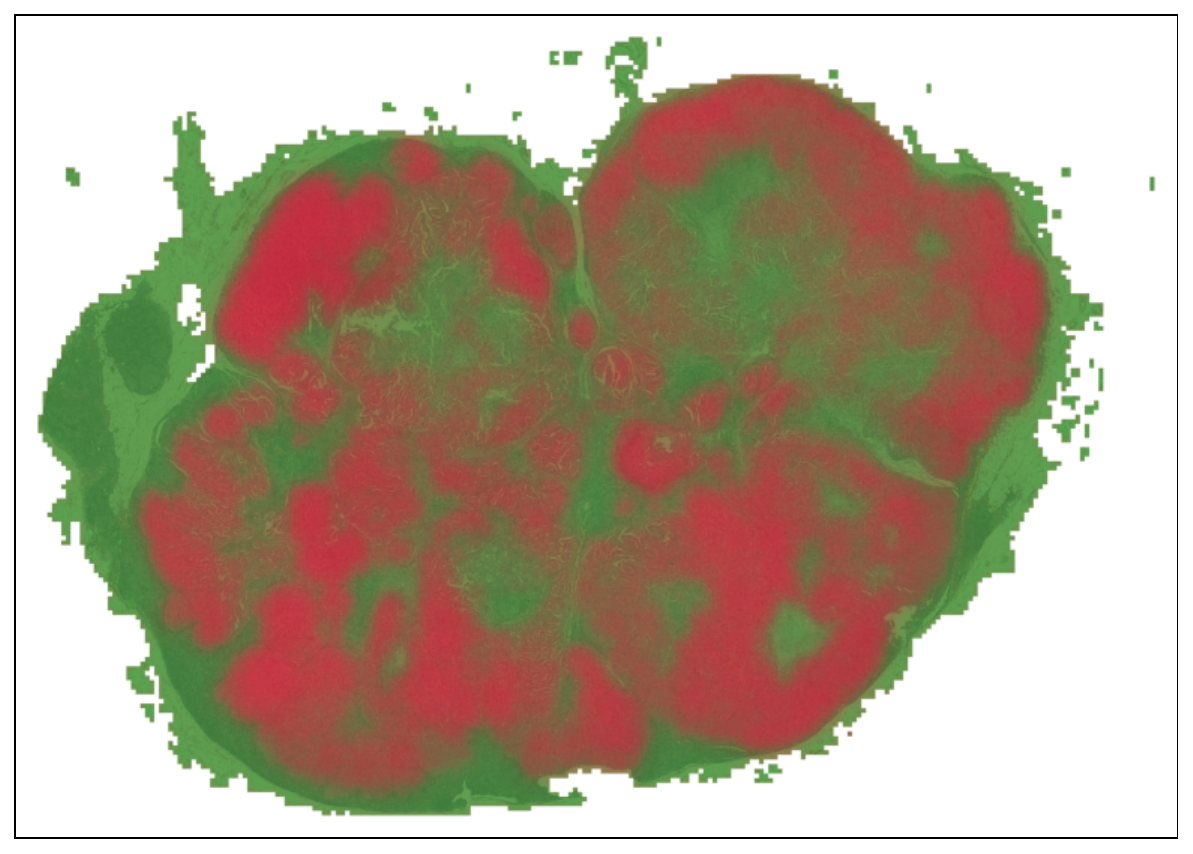} \\
    			\bottomrule			
    		\end{tabular}
    	\end{adjustbox}
    	\captionof{figure}{Ground truth and \smoothattpool\ attention maps of three different WSIs from CAMELYON16. As expected theoretically, a larger $\alpha$ produces smoother attention maps. }
        \label{fig:attmaps-camelyon-alpha-appendix}
    \end{center}
\end{figure}

% \subsection{Attention histograms}
% \label{subsection:appendix-att_histograms}

\begin{figure}

    \begin{center}
        \centering
        \begin{adjustbox}{width=\textwidth}
        \begin{tabular}{ccccc}
            \includegraphics[trim={0cm 0cm 0cm 0cm},clip,width=0.18\textwidth]
            {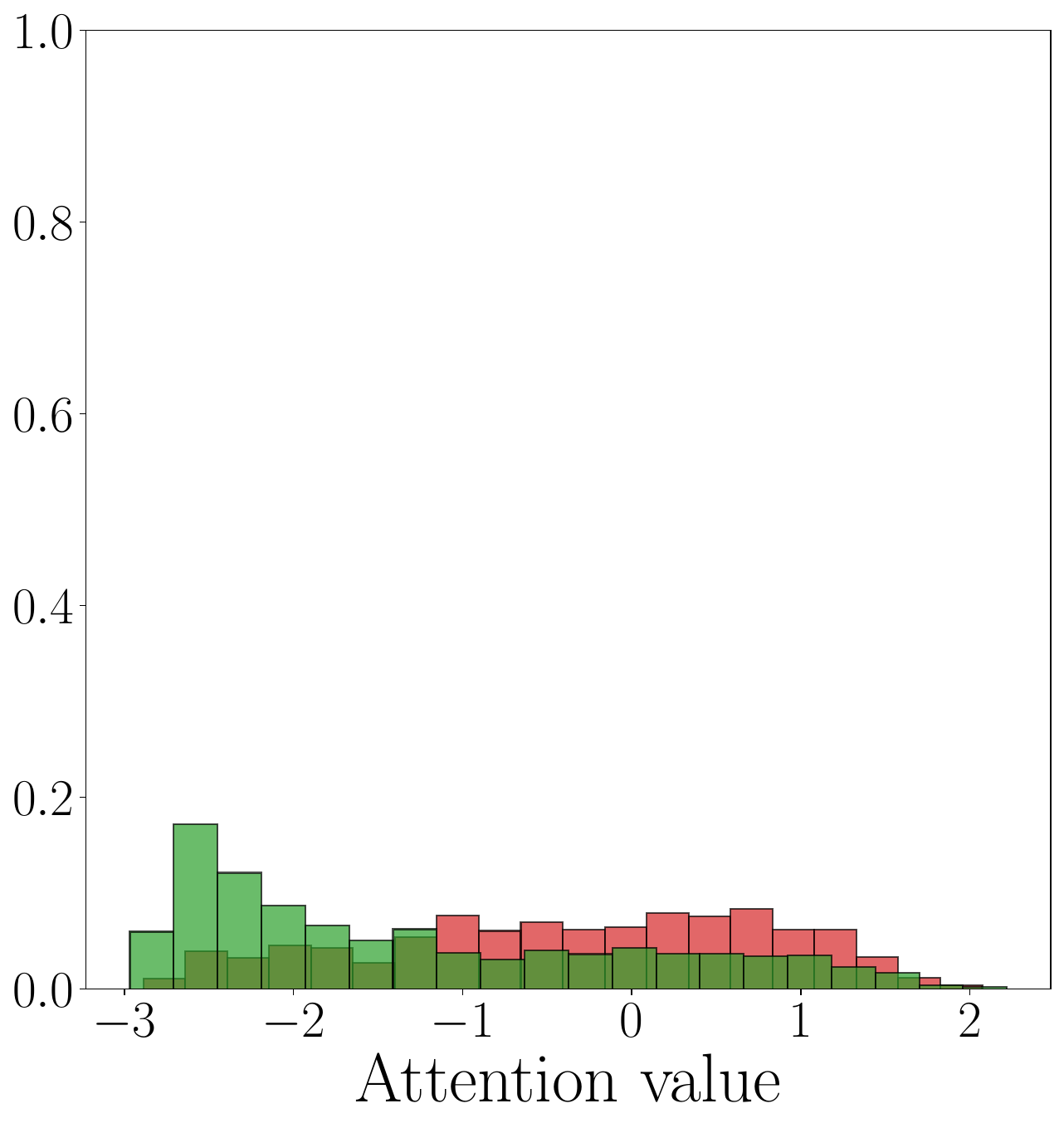}
            & 
            \includegraphics[trim={0cm 0cm 0cm 0cm},clip,width=0.18\textwidth]
            {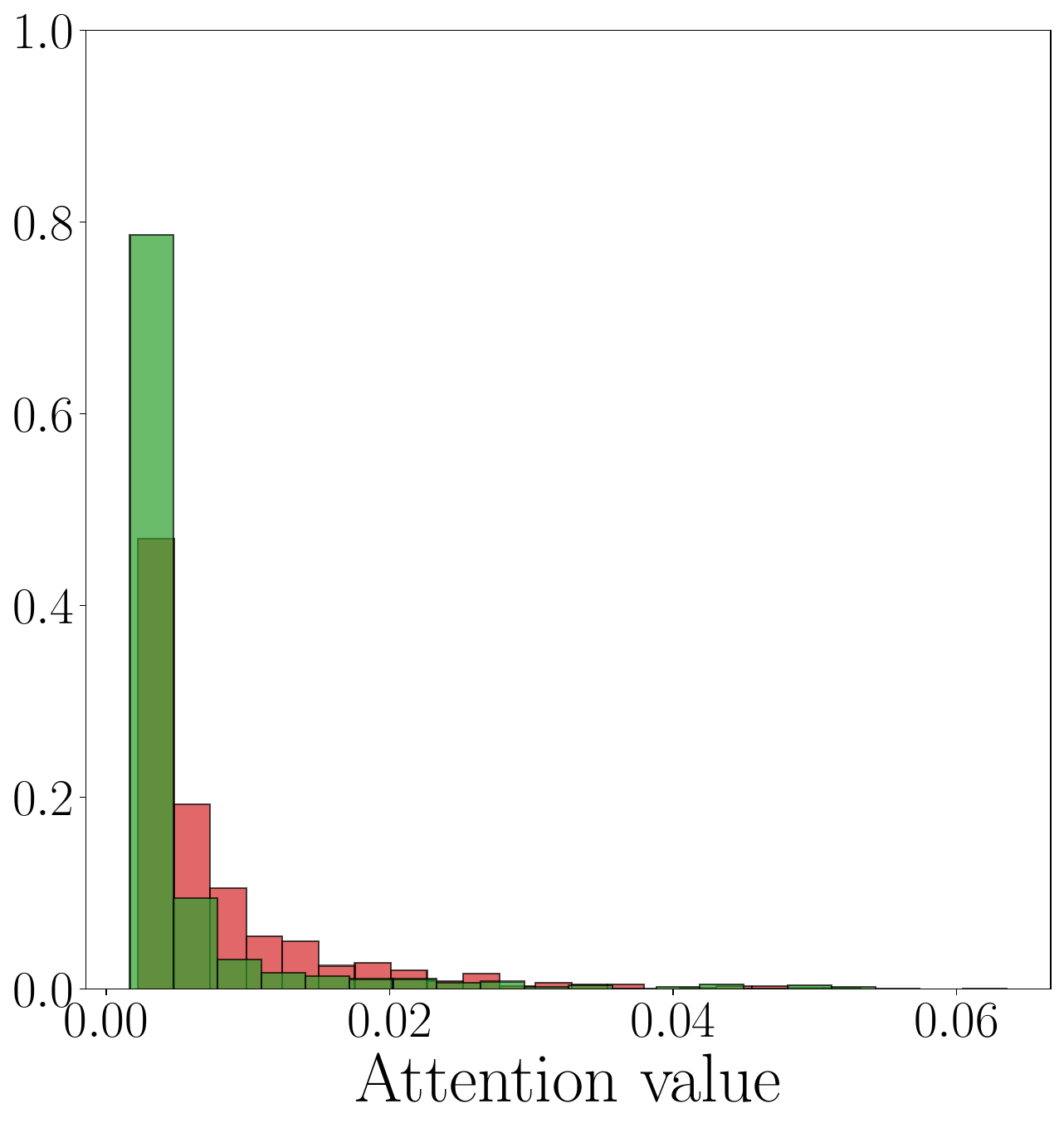}
            & 
            \includegraphics[trim={0cm 0cm 0cm 0cm},clip,width=0.18\textwidth]{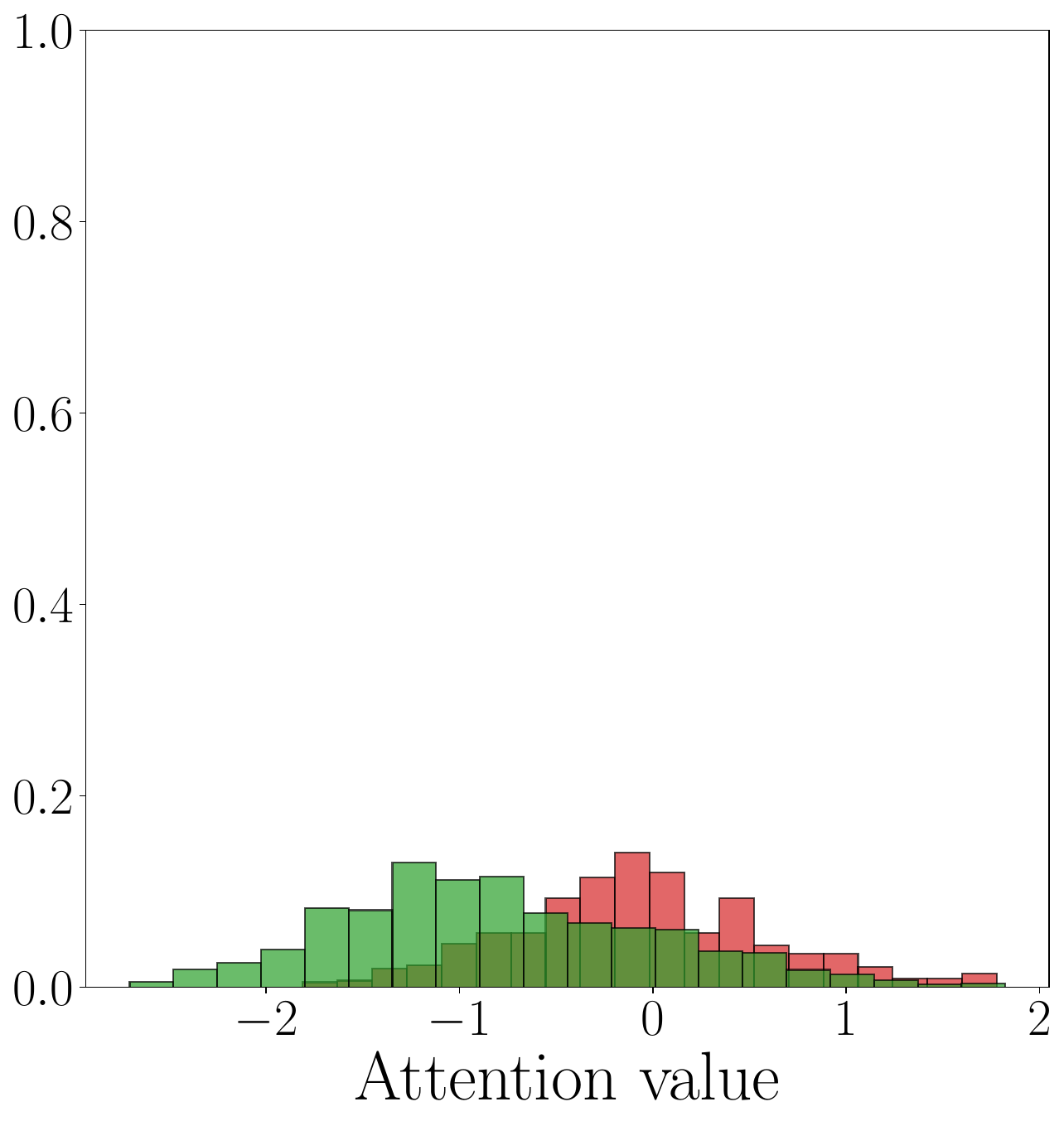}
            & 
            \includegraphics[trim={0cm 0cm 0cm 0cm},clip,width=0.18\textwidth]
            {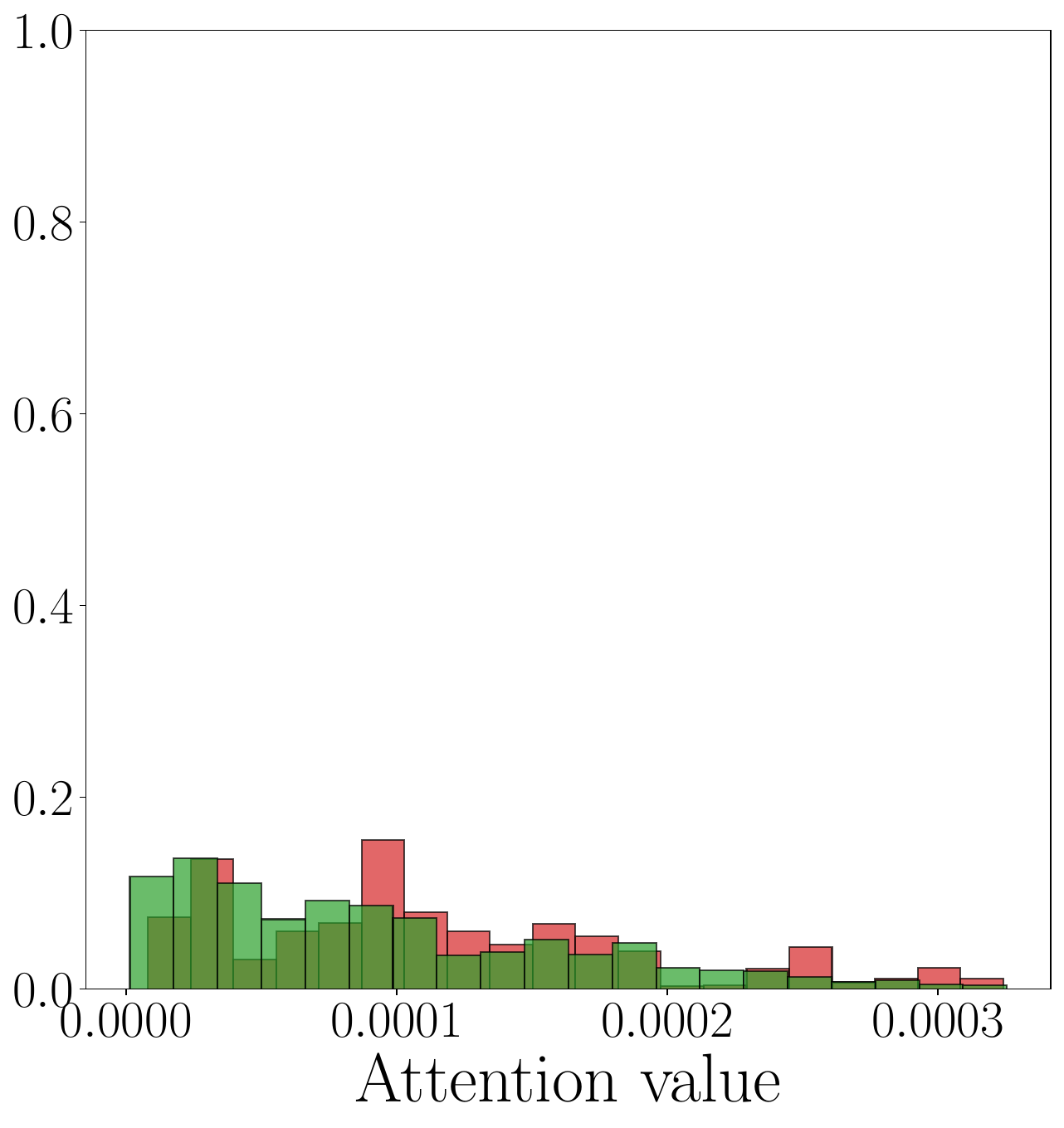}
            &
            \includegraphics[trim={0cm 0cm 0cm 0cm},clip,width=0.18\textwidth]
            {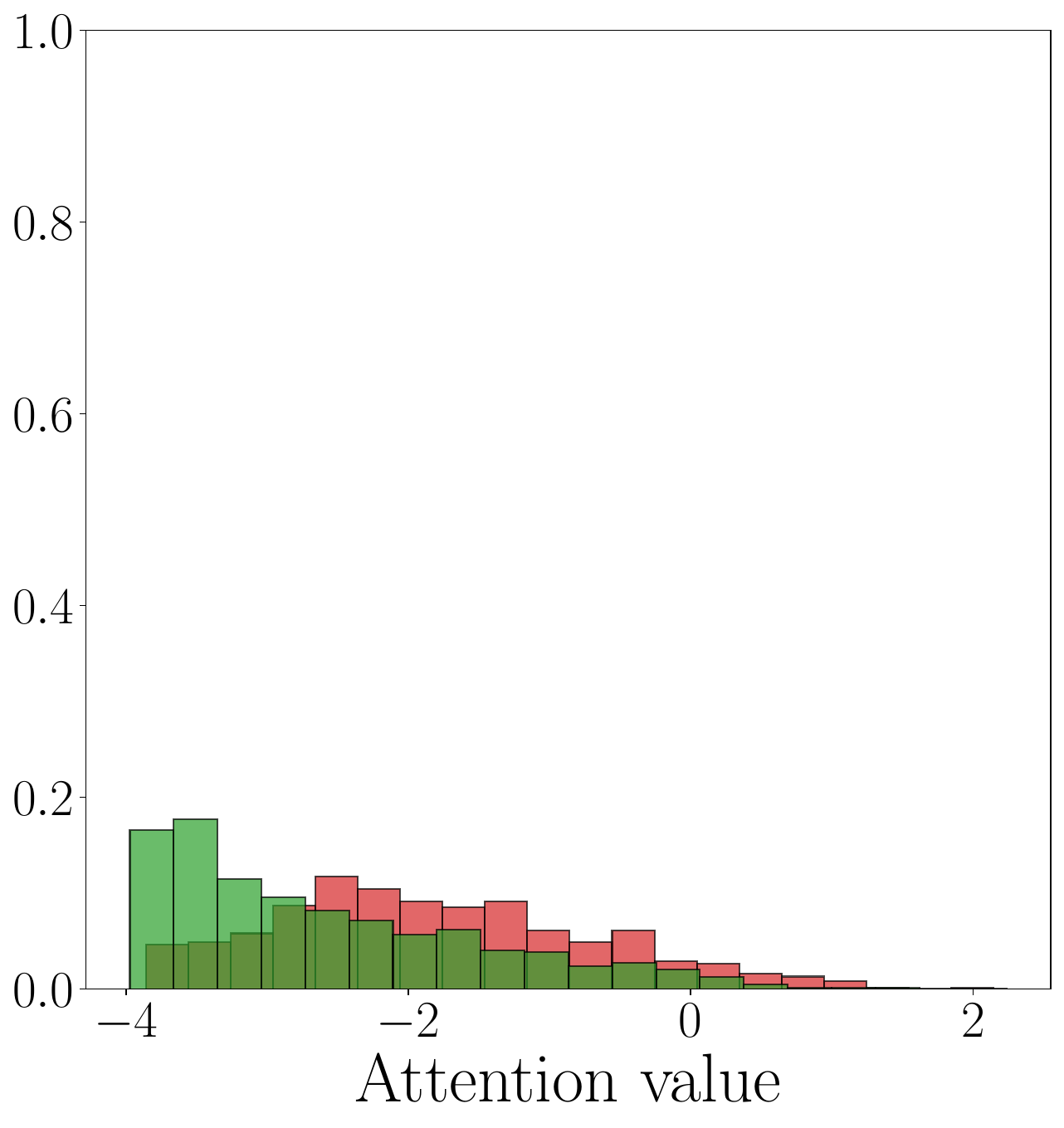}
            \\
             \smoothtransformerattpool &  \transmil &  \setmil &  \gtp &  \camil \\
            \includegraphics[trim={0cm 0cm 0cm 0cm},clip,width=0.18\textwidth]
            {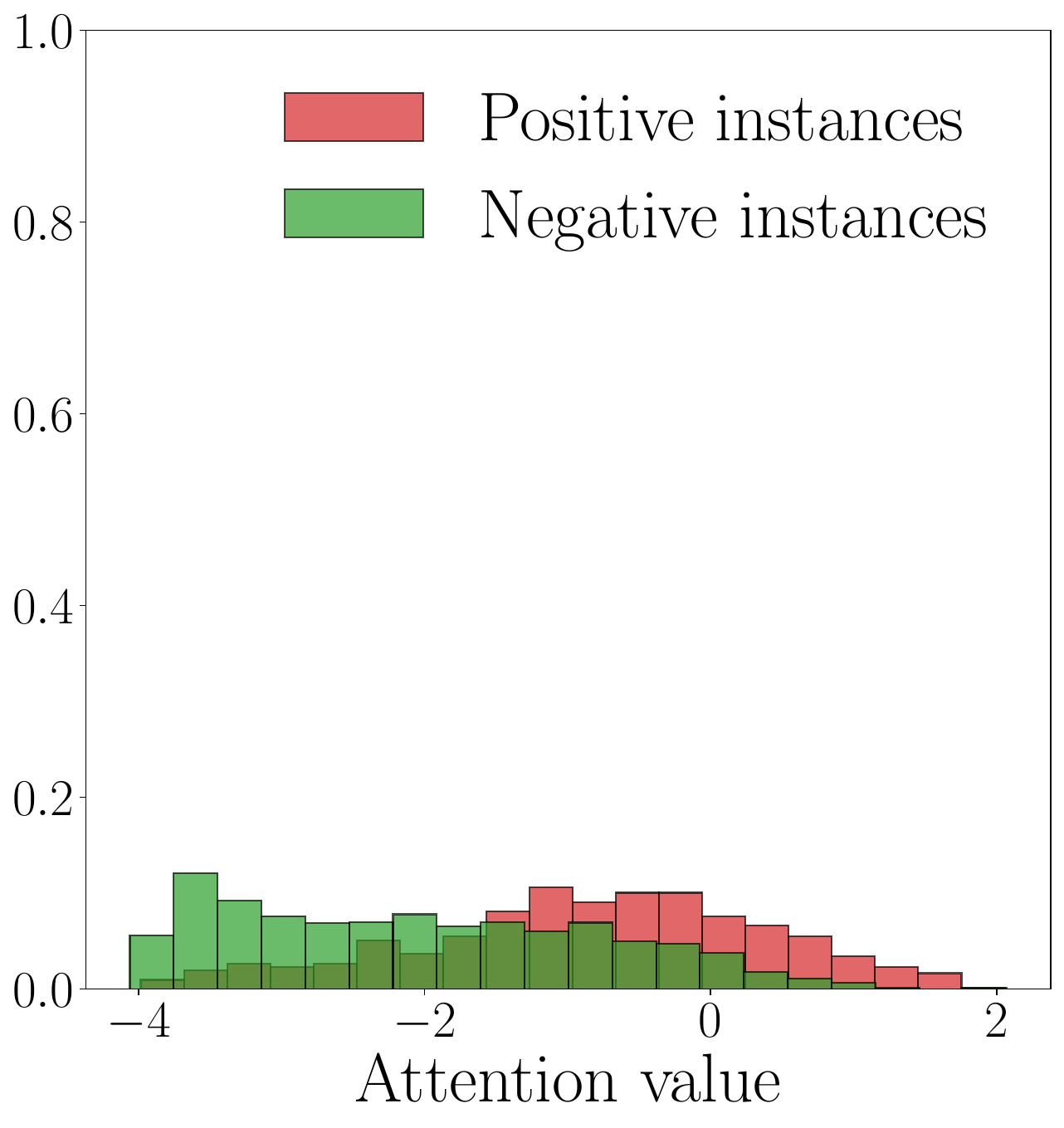}
            & 
            \includegraphics[trim={0cm 0cm 0cm 0cm},clip,width=0.18\textwidth]
            {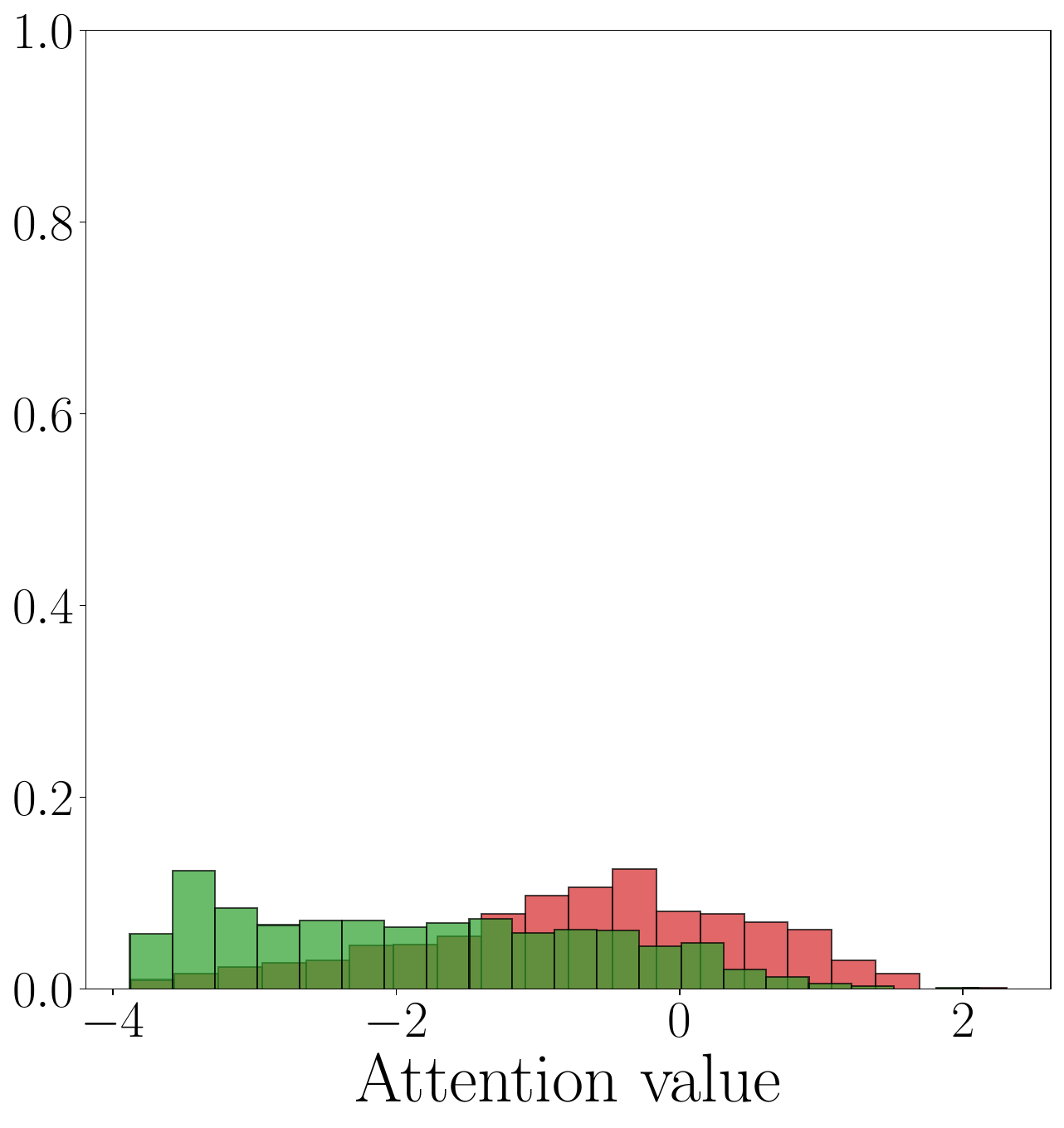}
            & 
            \includegraphics[trim={0cm 0cm 0cm 0cm},clip,width=0.18\textwidth]
            {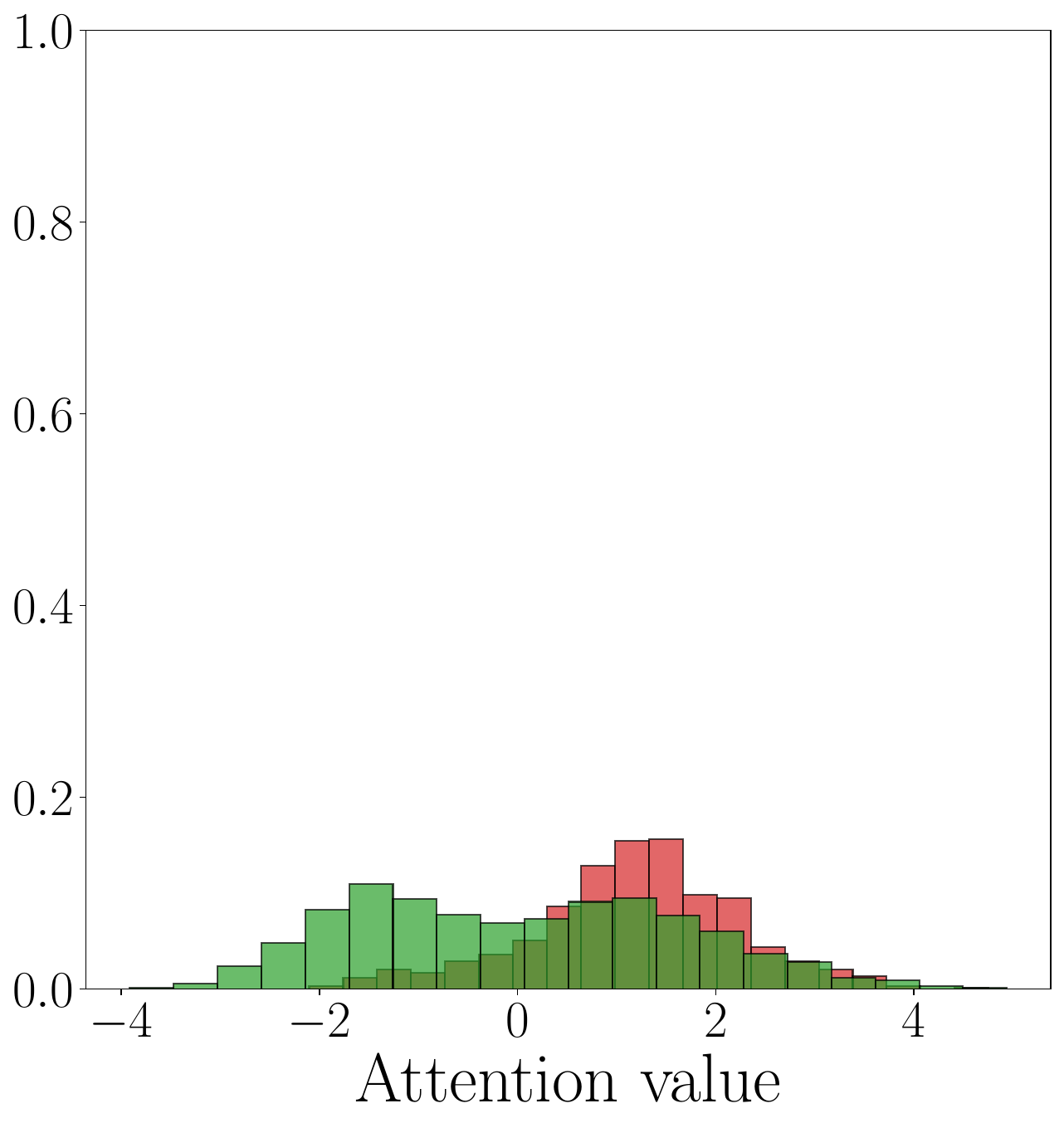}
            & 
            \includegraphics[trim={0cm 0cm 0cm 0cm},clip,width=0.18\textwidth]
            {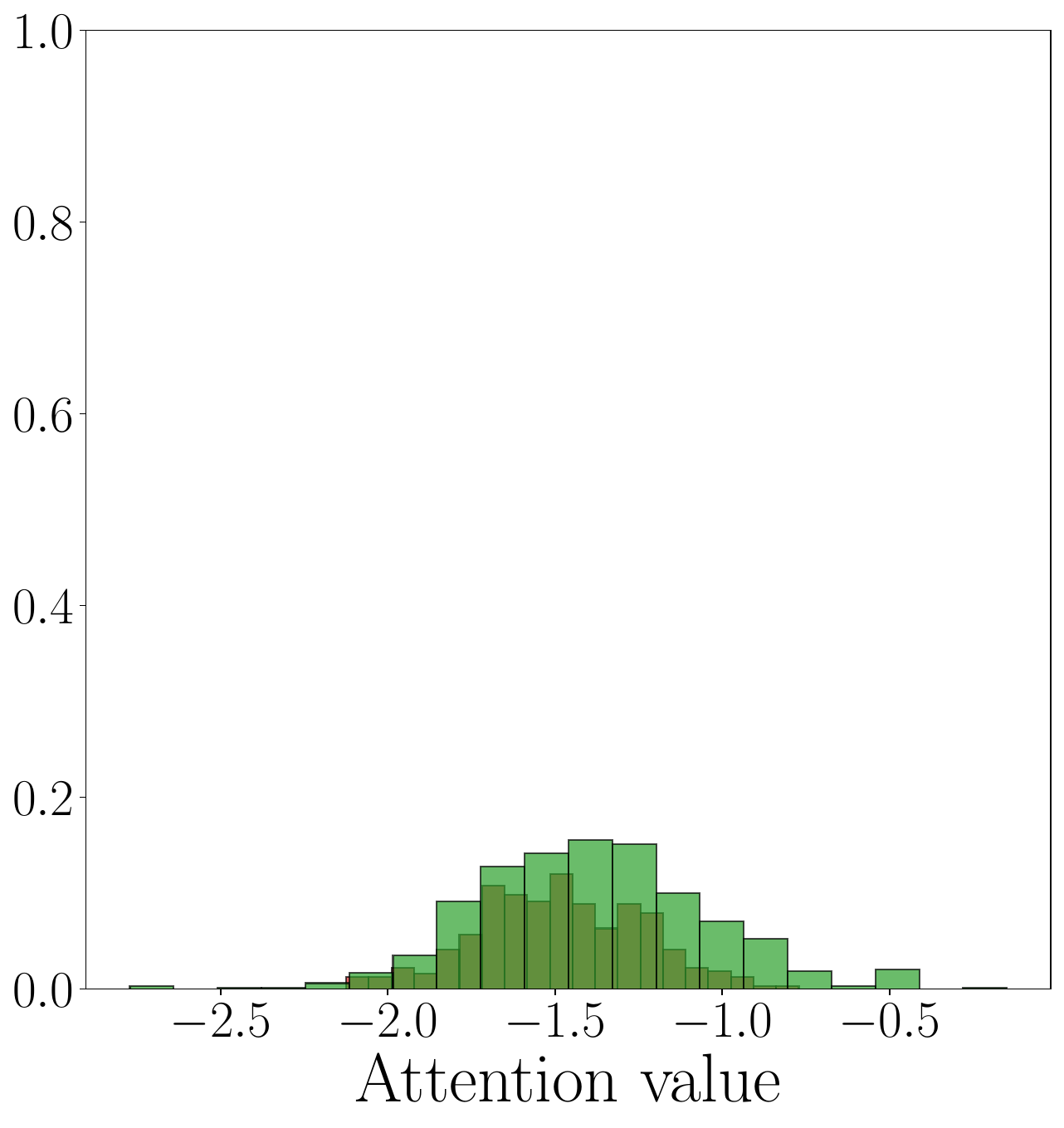}
            &
            \includegraphics[trim={0cm 0cm 0cm 0cm},clip,width=0.18\textwidth]{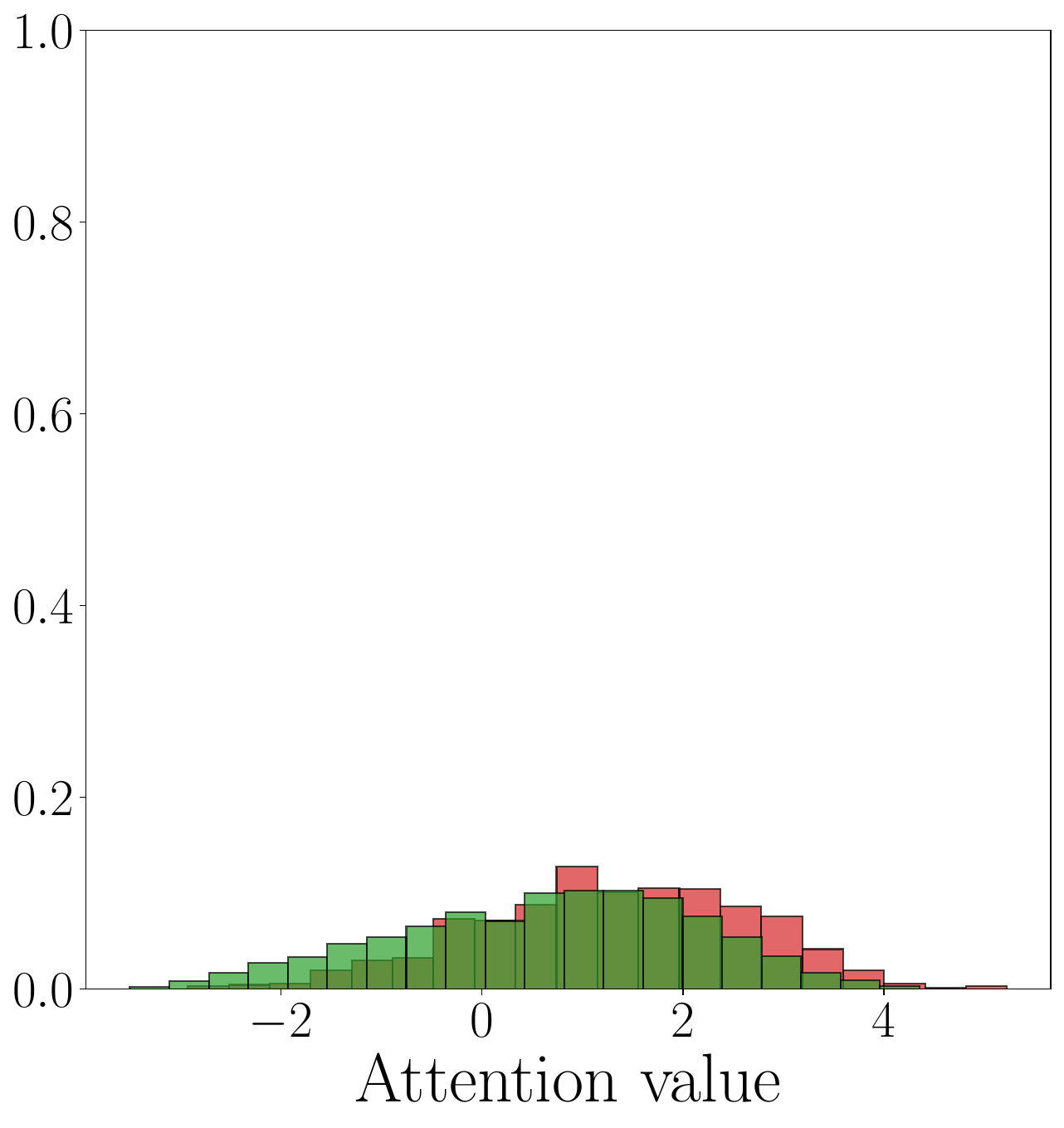}
            \\
             \smoothattpool &  \abmil &  \clam &  \dsmil &  \dftdmil
        \end{tabular}
        \end{adjustbox}
        \captionof{figure}{RSNA attention histograms.}
        \label{fig:attval_histograms-rsna-appendix}
    \end{center}
\end{figure}

% \begin{figure}
    \begin{center}
        \centering
        \begin{adjustbox}{width=\textwidth}
        \begin{tabular}{ccccc}
            \includegraphics[trim={0cm 0cm 0cm 0cm},clip,width=0.19\textwidth]
            {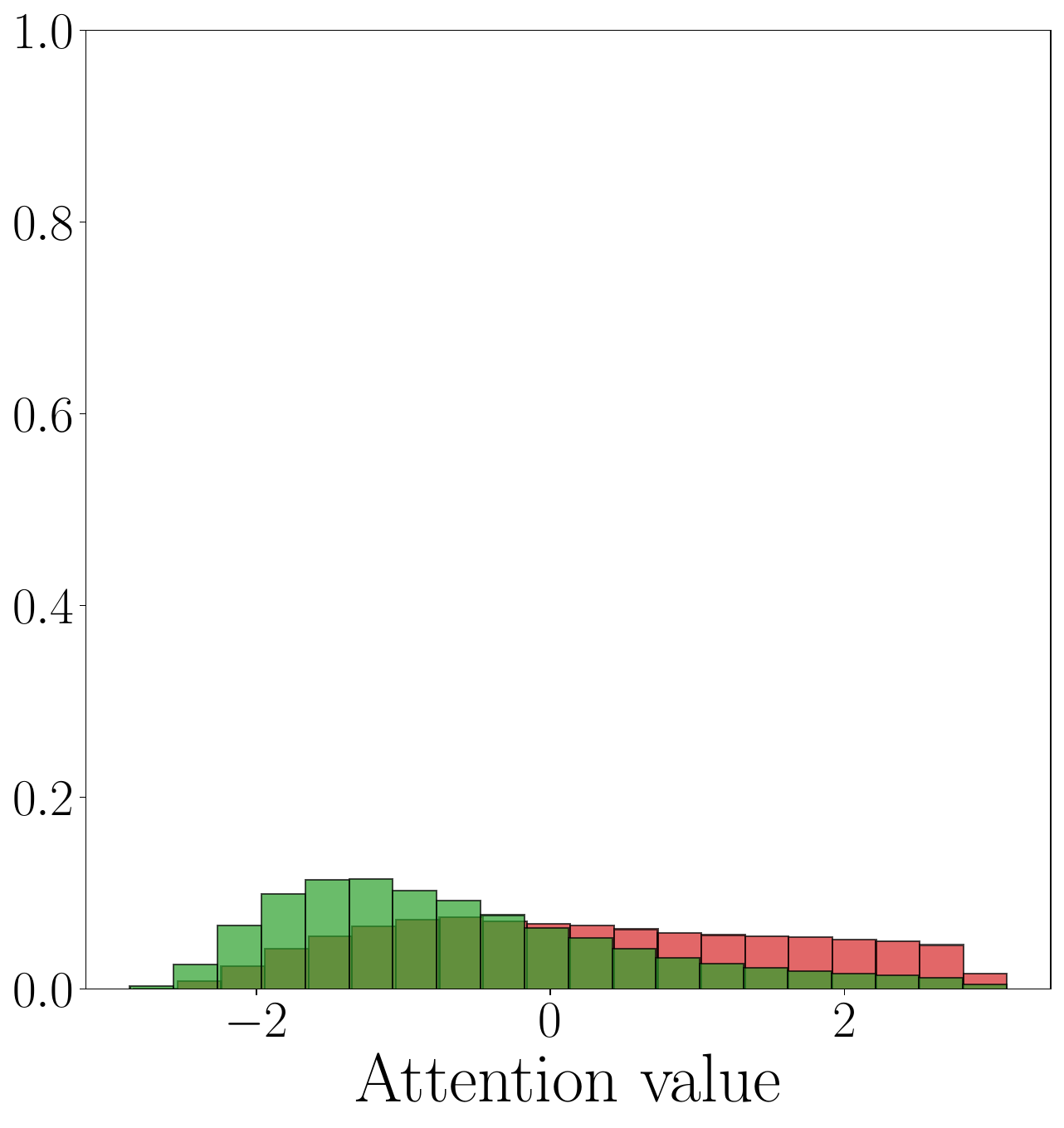}
            & 
            \includegraphics[trim={0cm 0cm 0cm 0cm},clip,width=0.19\textwidth]
            {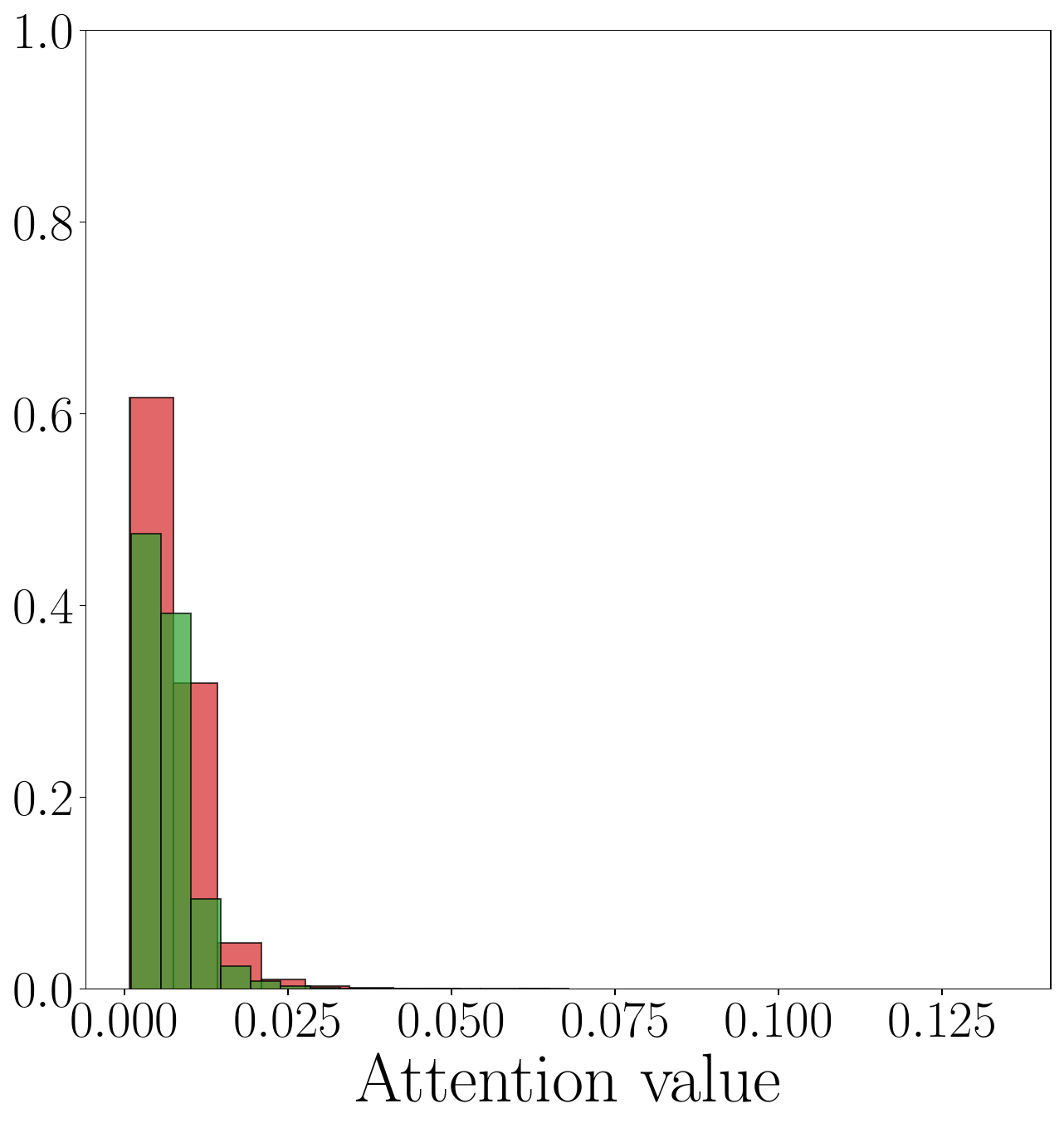}
            & 
            \includegraphics[trim={0cm 0cm 0cm 0cm},clip,width=0.19\textwidth]{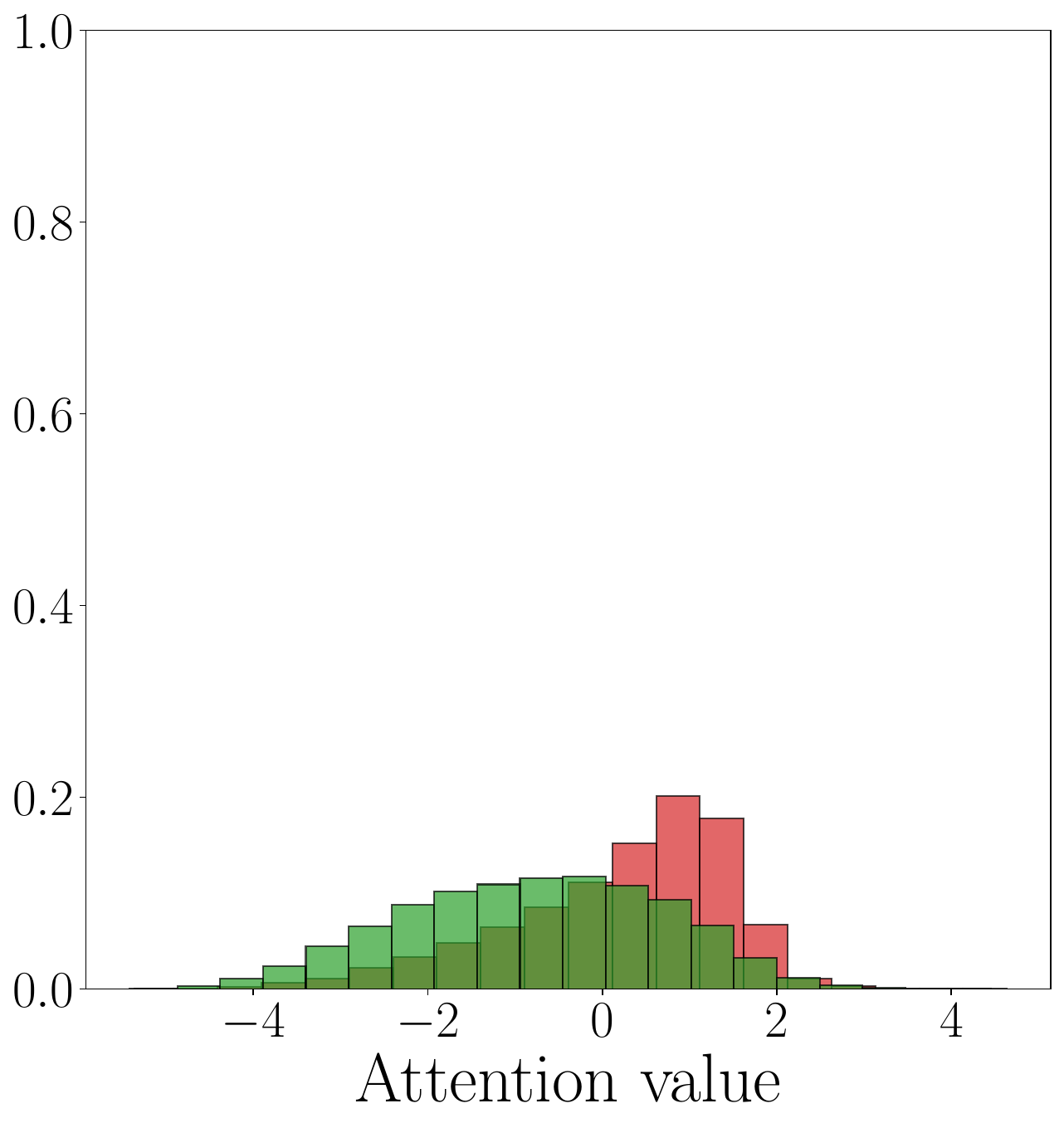}
            & 
            \includegraphics[trim={0cm 0cm 0cm 0cm},clip,width=0.19\textwidth]
            {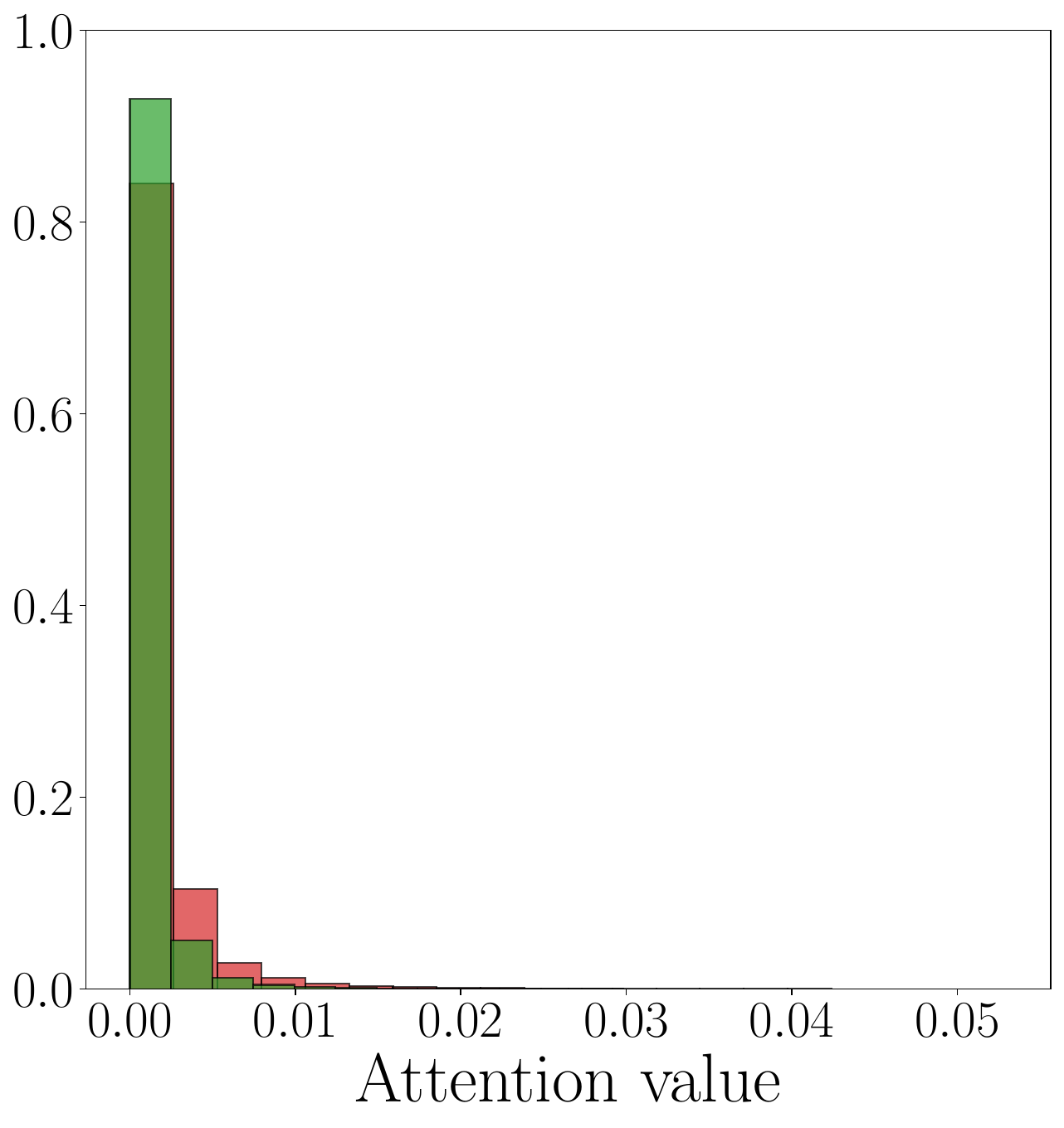}
            &
            \includegraphics[trim={0cm 0cm 0cm 0cm},clip,width=0.19\textwidth]
            {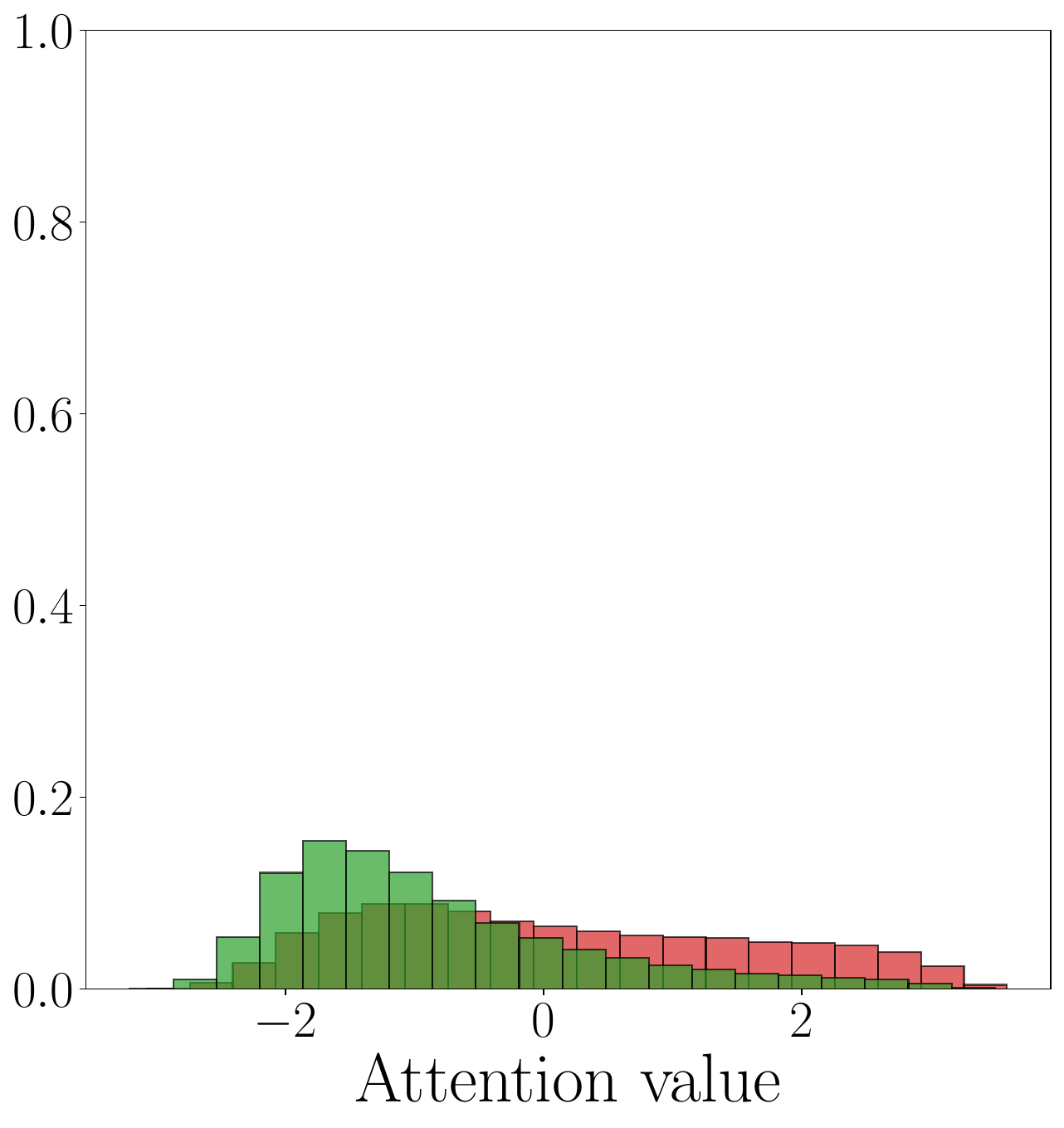}
            \\
             \smoothtransformerattpool &  \transmil &  \setmil &  \gtp &  \camil \\
            \includegraphics[trim={0cm 0cm 0cm 0cm},clip,width=0.19\textwidth]
            {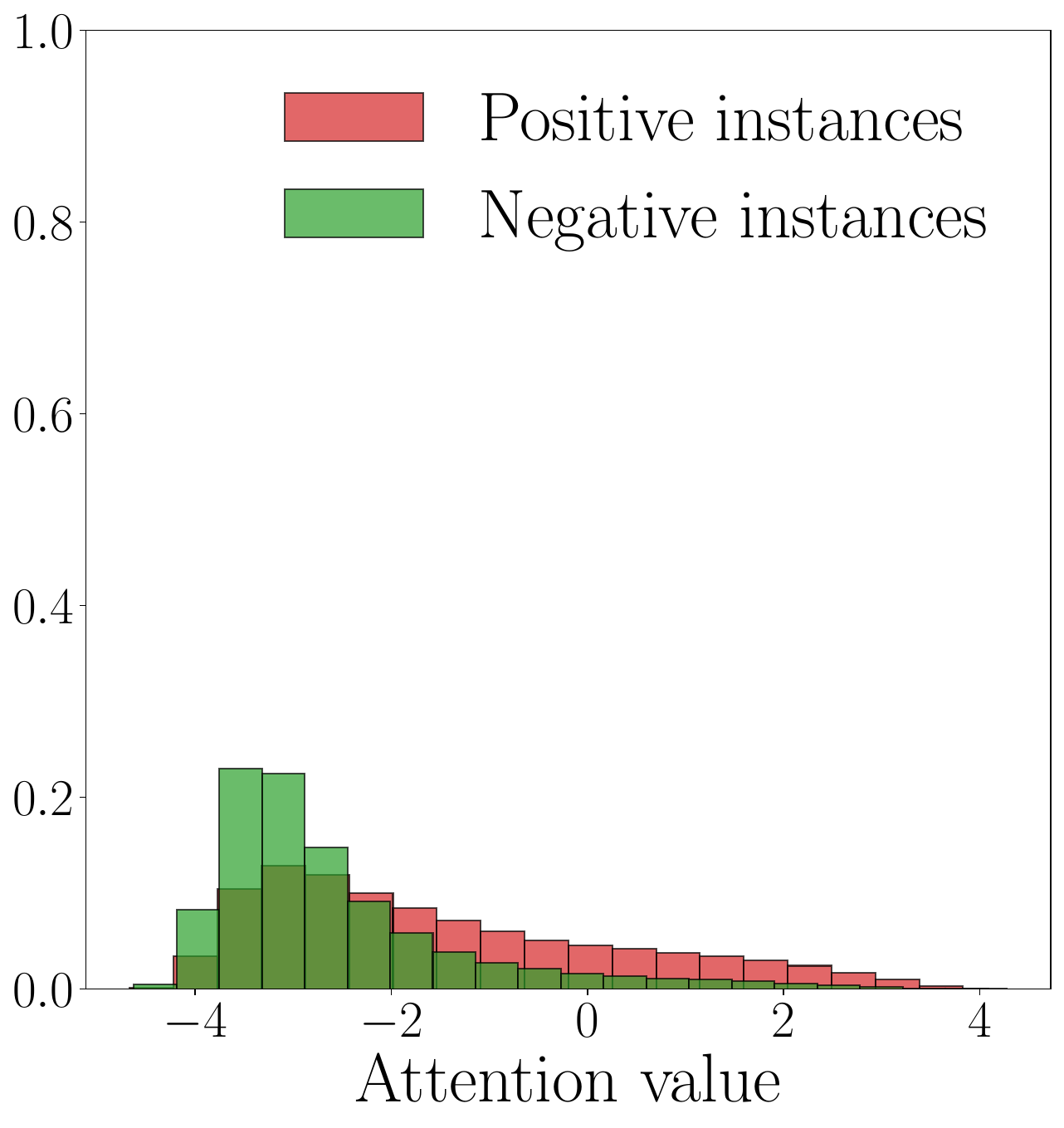}
            & 
            \includegraphics[trim={0cm 0cm 0cm 0cm},clip,width=0.19\textwidth]
            {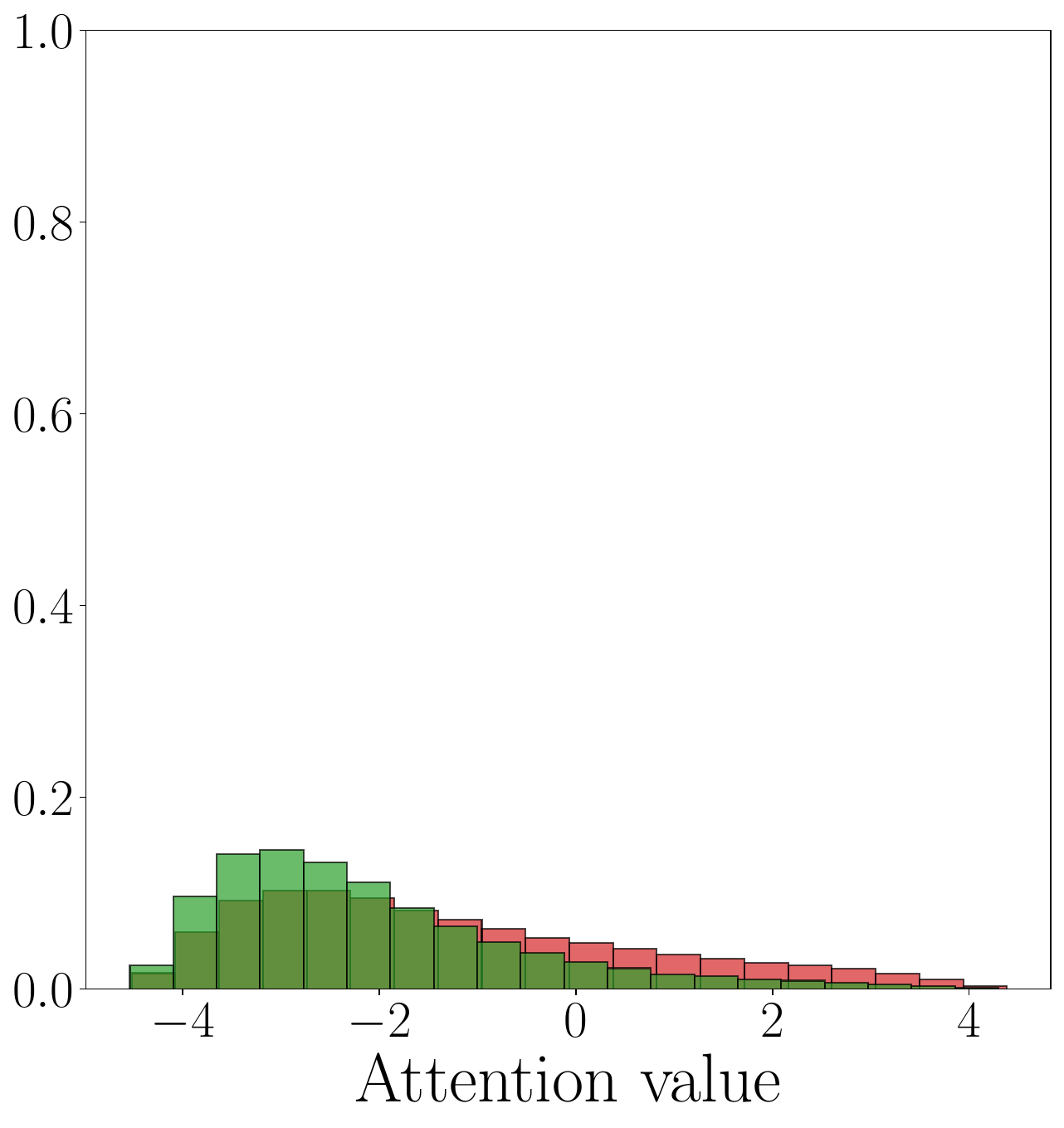}
            & 
            \includegraphics[trim={0cm 0cm 0cm 0cm},clip,width=0.19\textwidth]
            {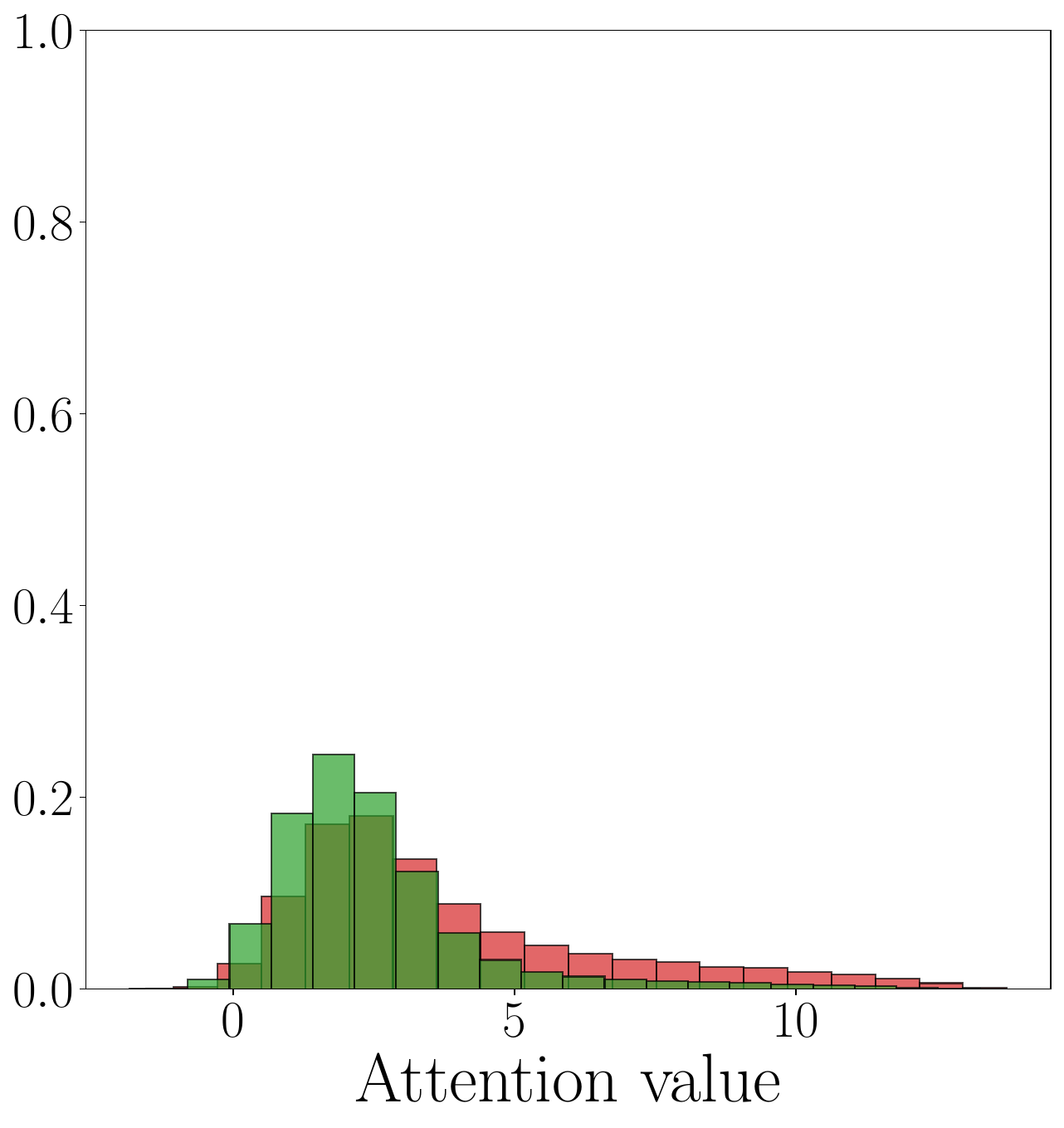}
            & 
            \includegraphics[trim={0cm 0cm 0cm 0cm},clip,width=0.19\textwidth]
            {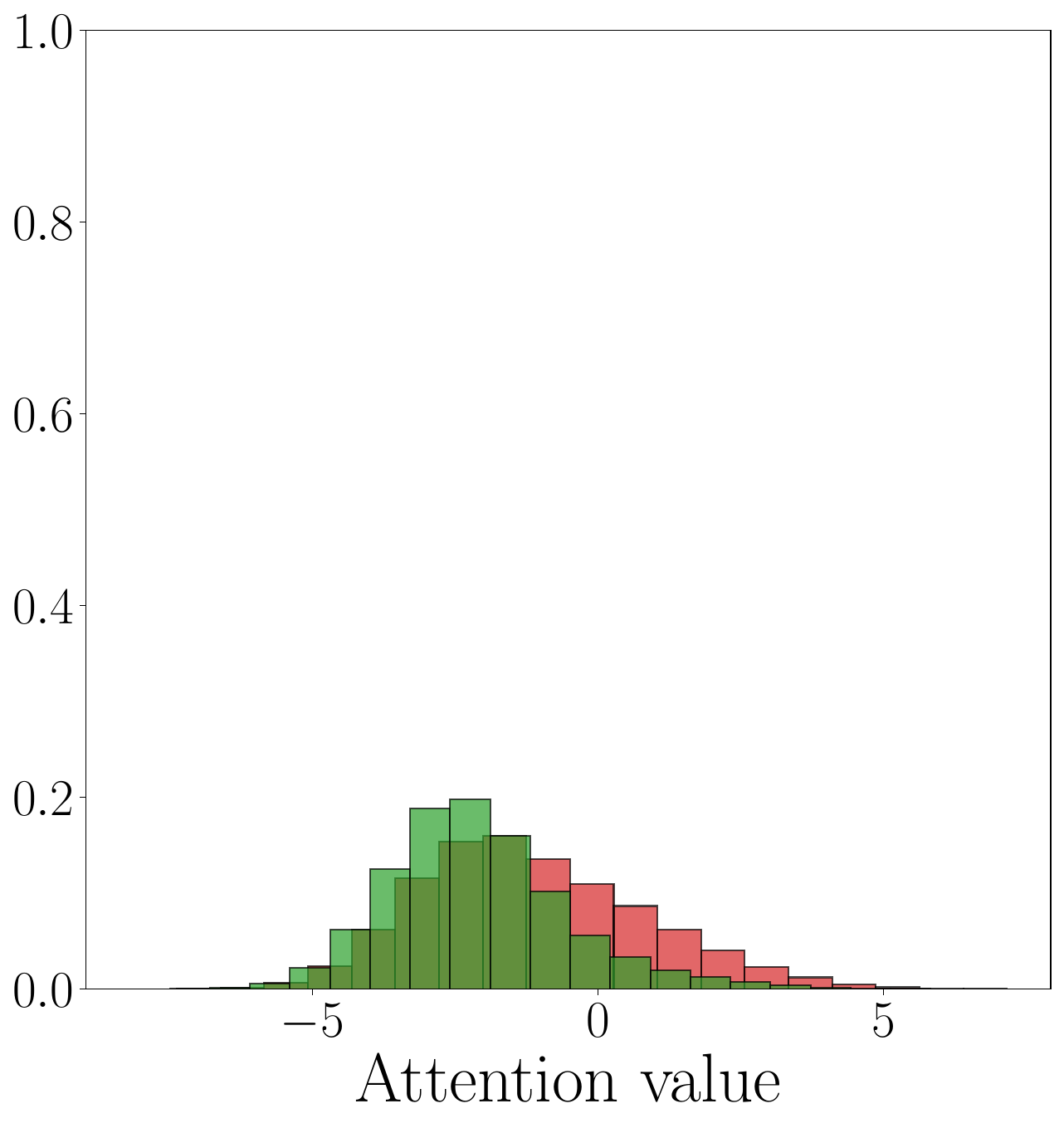}
            &
            \includegraphics[trim={0cm 0cm 0cm 0cm},clip,width=0.19\textwidth]{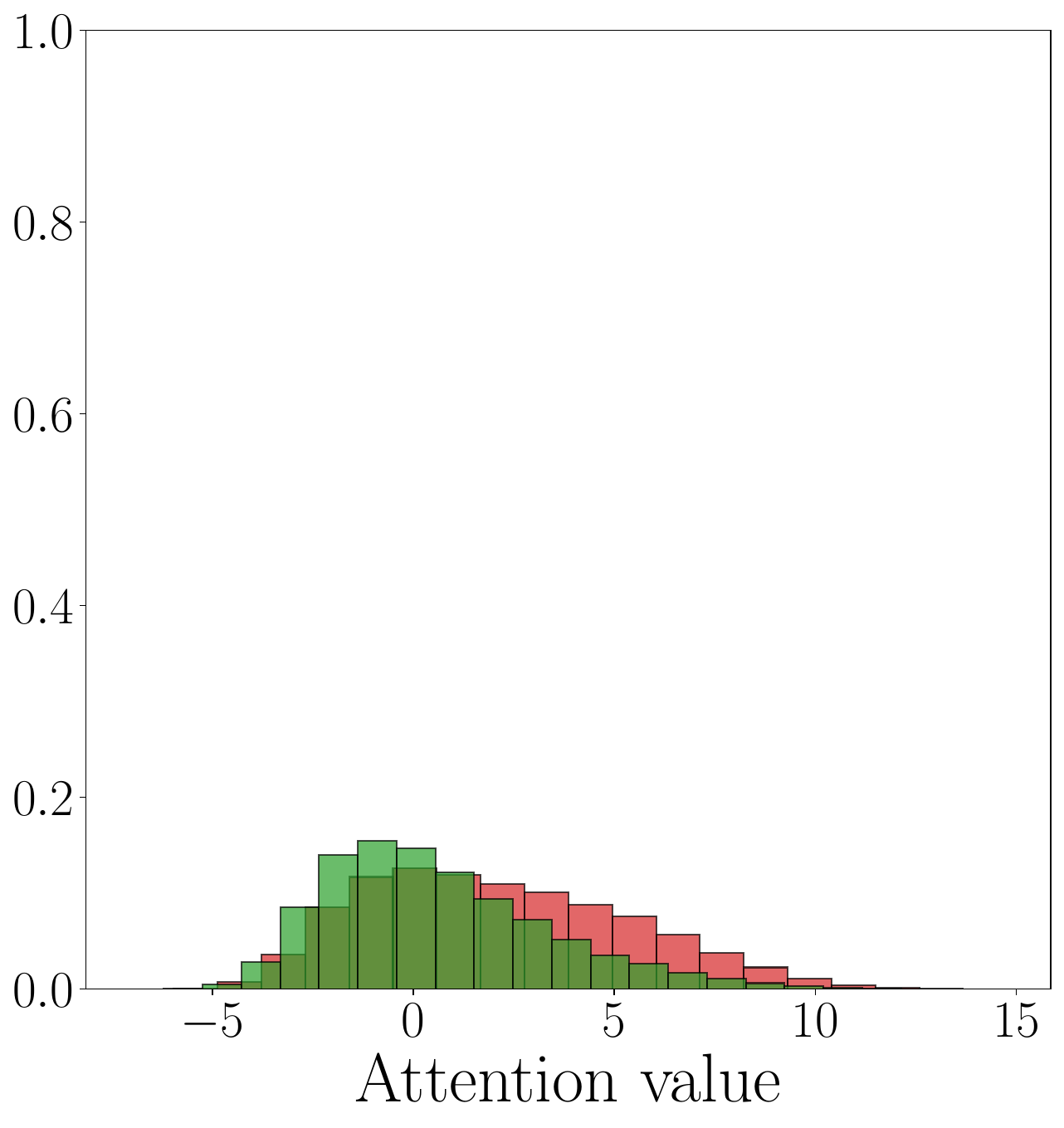}
            \\
             \smoothattpool &  \abmil &  \clam &  \dsmil &  \dftdmil
        \end{tabular}
        \end{adjustbox}
        \captionof{figure}{PANDA attention histograms.}
        %histograms. \textcolor{orange}{P: I do not fully understand these plots. What's the y axis? (relative frequency). Does this plot include different images? That could be a bit confusing and unclear (different images may have different ``separating thresholds''?), it is perhaps better to show just one or a few images of each dataset. Well, probably is better to show different images (as is now), because they all use the same threshold (calculated with a validation set, right?). Discuss. In any case, I do not understand ``relative frequency''. It does not seem to add up to 1 or 100. }
        \label{fig:attval_histograms-panda-appendix}
    \end{center}
% \end{figure}

\end{document}